\documentclass{article}

\PassOptionsToPackage{numbers, compress}{natbib}

    \usepackage[preprint]{neurips_2024}

\usepackage[utf8]{inputenc} % allow utf-8 input
\usepackage[T1]{fontenc}    % use 8-bit T1 fonts
\usepackage{hyperref}       % hyperlinks
\usepackage{url}            % simple URL typesetting
\usepackage{booktabs}       % professional-quality tables
\usepackage{amsfonts}       % blackboard math symbols
\usepackage{nicefrac}       % compact symbols for 1/2, etc.
\usepackage{microtype}      % microtypography
\usepackage{xcolor}         % colors
\usepackage{amsmath}

\usepackage{colortbl}
\usepackage{graphicx}
\usepackage{algorithm}
\usepackage{ulem}
\usepackage{algorithmic}
\usepackage{multirow}
\usepackage{wrapfig}
\usepackage{array}
\usepackage{caption}
\usepackage{subcaption}

\newcolumntype{C}[1]{>{\centering\arraybackslash}p{#1}}
\newcolumntype{L}{>{\hspace*{-\tabcolsep}}l}
\newcolumntype{R}{c<{\hspace*{-\tabcolsep}}}
\definecolor{highlightcolor}{rgb}{0.0, 0.5, 0.0}

\title{Out-of-Distribution Detection using Neural \\ Activation Prior}

\author{%
  Weilin Wan, Weizhong Zhang, Quan Zhou, Fan Yi, Cheng Jin\thanks{Corresponding author.}\\ 
  School of Computer Science, Fudan University \\
  \texttt{wlwan23@m.fudan.edu.cn, weizhongzhang@fudan.edu.cn}\\
  \texttt{qzhou21@m.fudan.edu.cn, fyi21@m.fudan.edu.cn, jc@fudan.edu.cn}\\
}

\begin{document}

\maketitle

\begin{abstract}
  Out-of-distribution detection (OOD) is a crucial technique for deploying machine learning models in the real world to handle the unseen scenarios. 
In this paper, we first propose a simple yet effective \textbf{N}eural \textbf{A}ctivation \textbf{P}rior (NAP) for OOD detection.
Our neural activation prior is based on a key observation that,  for a channel before the global pooling layer of a fully trained neural network, the probability of a few  neurons being activated with a large response by an in-distribution (ID) sample is significantly higher than that by an OOD sample. An intuitive explanation is that for a model fully trained on ID dataset, each channel   would play a role in detecting a certain pattern in  the ID dataset, and a few neurons can be activated with a large response when the pattern is detected in an input sample.
Then, a new scoring function based on this prior is proposed  to highlight the role of these strongly activated neurons in OOD detection.
Our approach is plug-and-play and does not lead to any performance degradation on ID data classification and requires no extra training or statistics from training or external datasets.
Notice that previous methods primarily rely on  post-global-pooling features of the neural networks, while the within-channel distribution information we leverage would be discarded by the global pooling operator. Consequently, our method is orthogonal to existing approaches and can be effectively combined with them in various applications. Experimental results show that our method achieves the state-of-the-art performance on CIFAR benchmark and ImageNet dataset, which demonstrates the power of the proposed prior. 
Finally, we extend our method to Transformers and the experimental findings indicate that NAP can also significantly enhance the performance of OOD detection on Transformers, thereby demonstrating the broad applicability of this prior knowledge.
\end{abstract}
\section{Introduction}
\label{sec:intro}

Deep learning has developed rapidly in the last decade and become a crucial technique in various fields. 
However, neural networks would frequently make erroneous judgments in inference when encounter  the data  that differs greatly from their training data, which is known as out-of-distribution (OOD) data. This challenge is growing more prevalent and is particularly vital in safety-critical areas such as autonomous driving~\cite{filos2020can, janai2020computer} and medical diagnosis~\cite{pooch2020can}, which urges the development of effective OOD detection methods.

In practice, OOD data exhibits   large diversity and is difficult to identify~\cite{yang2021generalized_survey}.
Existing studies typically formulate OOD detection as a one-class classification task, utilizing prior knowledge. They~\cite{hendrycks2016msp, liu2020energy, yu2023block}  propose various priors, based on which  they further design scoring functions to distinguish OOD samples from ID samples.
For example, Hendrycks et al.~\cite{hendrycks2016msp}  observed that OOD samples always exhibit lower maximum softmax probabilities, and accordingly proposed using the maximal softmax probability output by a neural network as an OOD indicator. Liu et al.~\cite{liu2020energy} found that OOD samples usually have lower logits values, and based on this, an energy function was proposed for OOD detection.
Drawing from these precedents, it's clear that existing methods largely rely on the introduction of certain priors.  While some promising results highlight the effectiveness of these heuristics, a gap remains in meeting the practical requirements of real-world applications. Importantly, we note that the focus of these priors is narrowly concentrated on features and weights following global pooling, while the characteristics before the pooling layer are consistently ignored. Thus, we believe that finding and incorporating priors that can complement these existing focuses is essential, which will constitute the main contribution of our work.

\begin{figure}[!t]
  \centering
  \begin{subfigure}{0.63\linewidth}
    \includegraphics[width=0.99\linewidth]{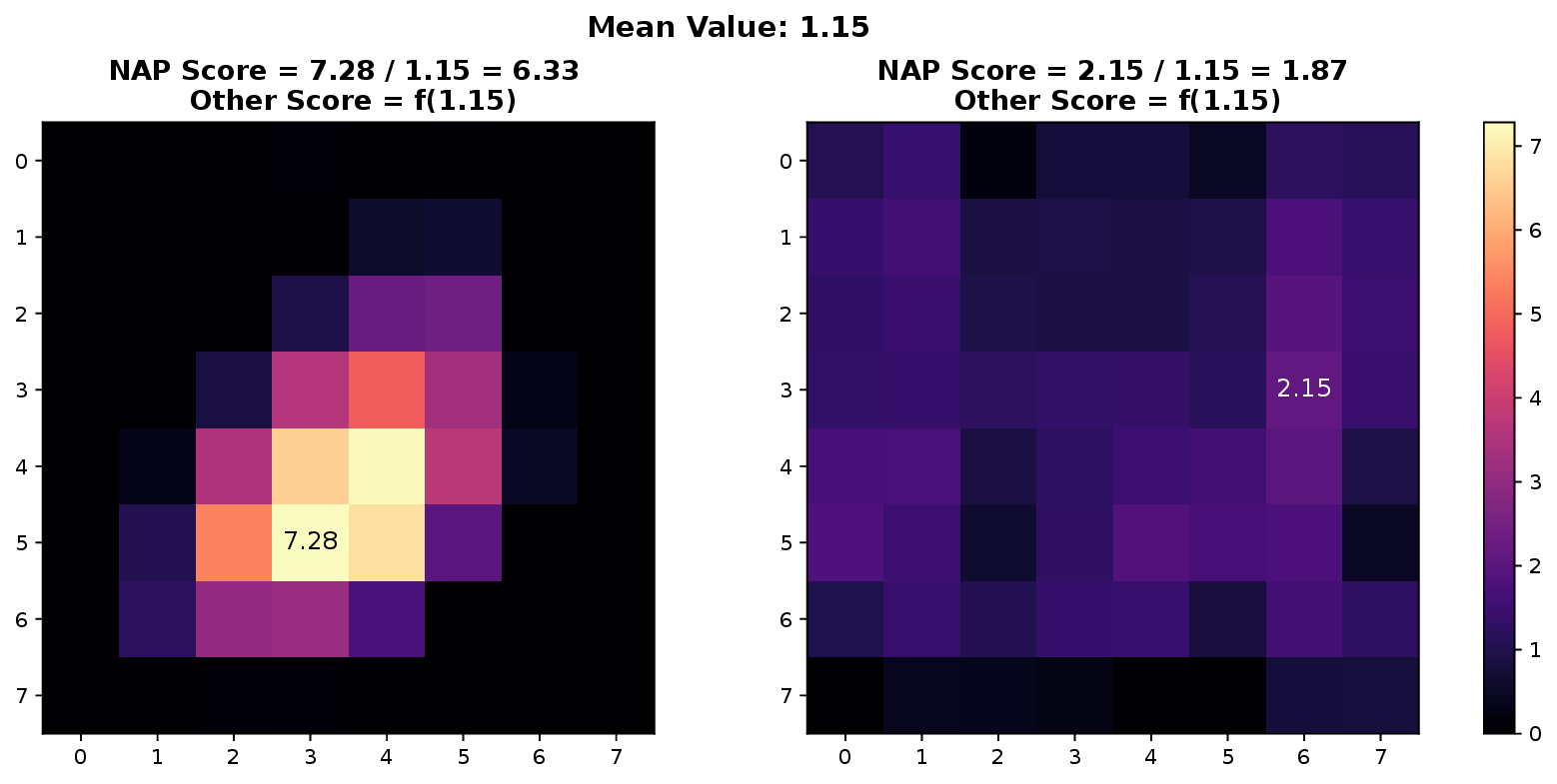}
    \caption{Activation Maps of ID and OOD samples.}
    \label{fig:activation_map}
  \end{subfigure}
  \hfill
  \begin{subfigure}{0.36\linewidth}
    \includegraphics[width=0.99\linewidth]{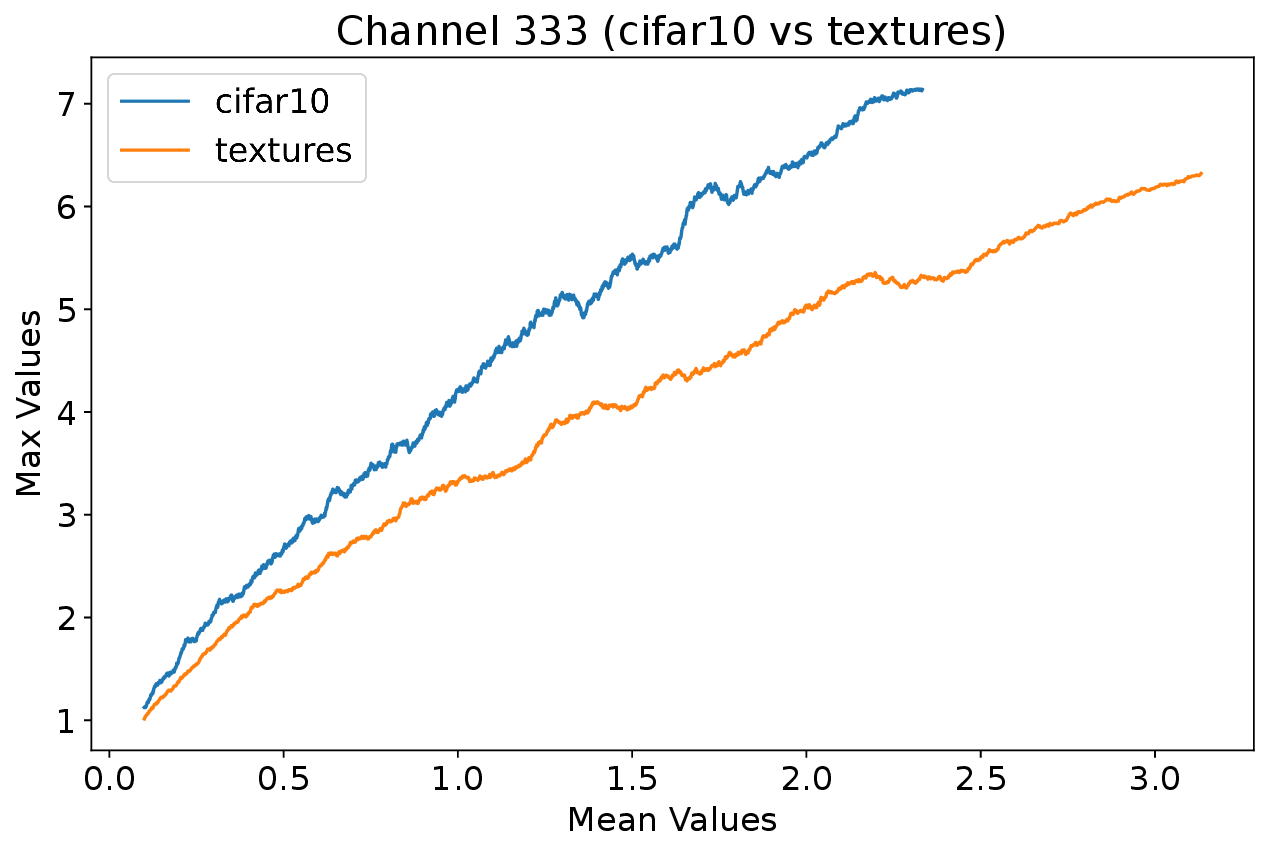}
    \caption{Activation distribution difference between ID data and OOD data.}
    \label{fig:activation_dis}
  \end{subfigure}
  \caption{\textbf{Global pooling leads to the neglect of distributional characteristics  of activation values within channels, thereby making it difficult to distinguish between ID and OOD samples.} (a) illustrates the activation value maps in the penultimate layer for in-distribution (ID) samples (left) and out-of-distribution (OOD) samples (right) within the same channel. In this layer, different channels typically focus on different semantic features. When a specific feature appears in an image, such as the central region in the left image, that location will have very high activation values. Although OOD samples lack these specific features, the model's unfamiliarity with OOD data leads to unpredictable activations, possibly causing weak noise activations (right image). This is detailed in \cite{sun2021react}. Existing methods often rely on pooled activation values for OOD detection. Thus, in (a), both ID and OOD samples have a mean channel activation of 1.15, making them indistinguishable by existing methods. However, the NAP score proposed in this paper can effectively distinguish between them. More examples like (a) are provided in the Appendix~\ref{appendix:vis}.
  (b) shows the activation distribution differences between ID and OOD data within the channel. The horizontal axis represents the average activation value, while the vertical axis represents the maximum activation value. Interestingly, at the same average activation level, ID data (CIFAR-10) shows significantly higher maximum activation values than OOD data (textures). This pattern is \textbf{not unique} to the \(333^\text{rd}\) channel but is observable in most channels, corroborating the phenomenon in (a).}
  \label{fig:one}
\end{figure}

In this study, we propose a novel prior, called \textbf{N}eural \textbf{A}ctivation \textbf{P}rior (NAP), for OOD detection. NAP characterizes our key observation that for channels before the global pooling layer in a fully trained neural network (illustrated to Figure~\ref{fig:pos}), a few neurons have a   significantly high probability to show 
a larger response when activated by in-distribution (ID) samples compared to OOD samples. An intuitive explanation for NAP is that channels in a model fully trained on an ID dataset play a role in detecting certain patterns in the input samples from the ID dataset. 
When such patterns are detected, a few neurons can be activated~\cite{hoefler2021sparsity}, resulting in larger responses. These large responses usually occur when the input sample is ID data, but when the input is OOD data, such responses are rarely observed. This is because the pattern that the neuron focuses on is unlikely to be present in the OOD data.
To verify our proposed prior, we employed DenseNet architecture~\cite{huang2017densely} on CIFAR-10~\cite{krizhevsky2009cifar} and Texture~\cite{cimpoi2014texture} datasets, analyzing mean and maximum within-channel activations before global pooling at the penultimate layer. Figure~\ref{fig:one} clearly demonstrates that ID samples exhibit significantly higher maximal activation values than OOD samples at equal average activation levels. Consequently, a series of methods, such as those in~\cite{hendrycks2016msp,sun2022knn,liu2020energy,liang2018odin_confi,lee2018maha_dist,sun2022dice,djurisic2022extremely,ahn2023line,sun2021react}, based on pooled activation values are unable to effectively distinguish these OOD samples, given the non-discriminative nature of their average activation values.

% \begin{figure}[!t]
%   \centering
%   \includegraphics[width=0.6\linewidth]{fig/channel_vis/channel_333.pdf}
%    \caption{\textbf{Activation distribution difference between ID data and OOD data inside channel.} This illustrates the difference in within-channel activation distribution between CIFAR-10~\cite{krizhevsky2009cifar} (ID data) and the Texture~\cite{cimpoi2014texture} (OOD data) datasets. The horizontal axis represents the average activation value of the channel, while the vertical axis shows the maximal activation value of the channel. Interestingly, at the same mean activation levels, the ID data (CIFAR-10) show significantly higher maximum activation values within the channel compared to the OOD data (Texture). It's important to note that this pattern is \textbf{not unique} to the \(333^\text{rd}\) channel but is observed across most channels.}
%    \label{fig:one}
%    % \vspace{-5pt}
% \end{figure}

It is worth noting that our proposed prior is orthogonal to that used in current OOD detection methods. In the OOD deteciton field, as shown in Figure~\ref{fig:pos}, existing methods~\cite{hendrycks2016msp,sun2022knn,liu2020energy,liang2018odin_confi,lee2018maha_dist,sun2022dice,djurisic2022extremely,ahn2023line,sun2021react} mainly focus on the outputs and weights of the neural network after global pooling, and use them to design scoring functions for OOD detection. In contrast, the prior we proposed is focus within the channels of the penultimate layer before global pooling. Since the information carried in our prior can be easily lost during the global pooling process, this demonstrates that our NAP  is essentially complementary to the priors used in current OOD detection studies. To this end, we would like to emphasize that the contribution of this paper would lie in how much improvement we can achieve by integrating our method with  existing approaches, rather than a direct comparison with existing methods in the field.

\begin{figure}[!t]
  \centering
  \begin{subfigure}{0.32\linewidth}
    \includegraphics[width=0.99\linewidth]{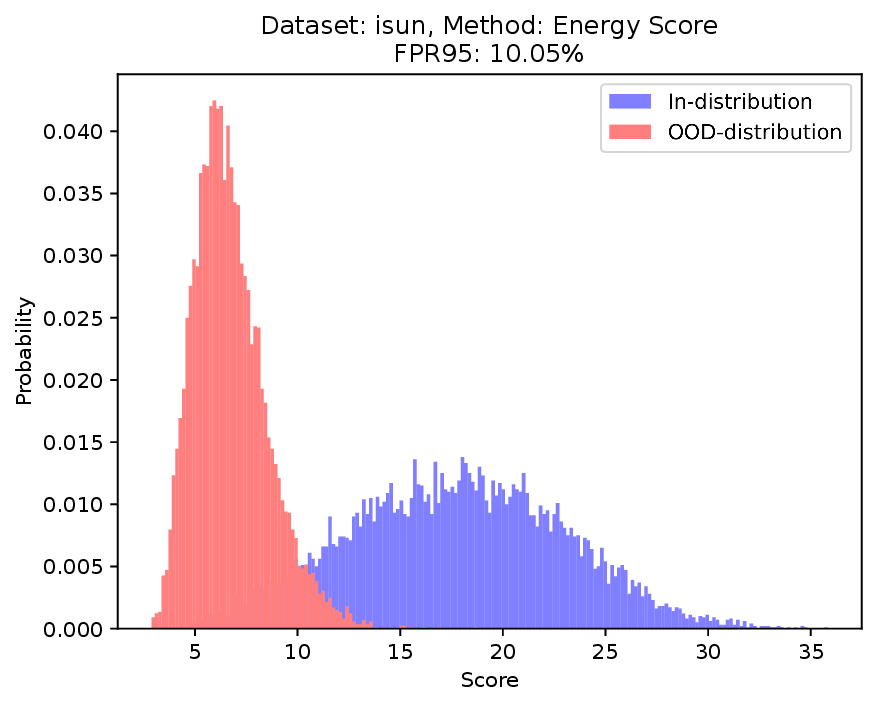}
    \caption{Energy Score}
    \label{fig:hist-a}
  \end{subfigure}
  \hfill
  \begin{subfigure}{0.32\linewidth}
    \includegraphics[width=0.99\linewidth]{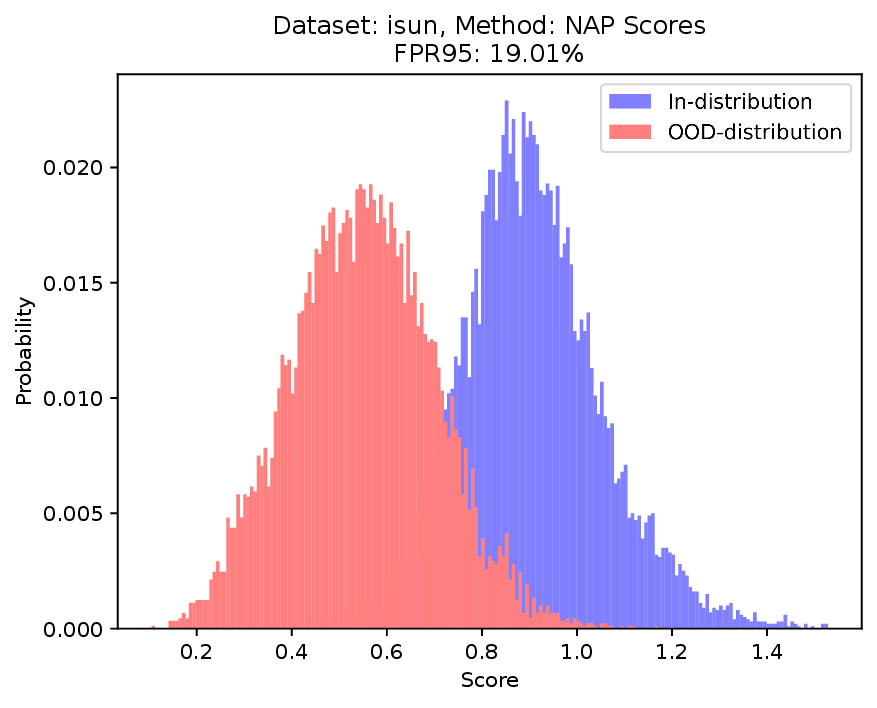}
    \caption{NAP Score}
    \label{fig:hist-b}
  \end{subfigure}
  \hfill
  \begin{subfigure}{0.32\linewidth}
    \includegraphics[width=0.99\linewidth]{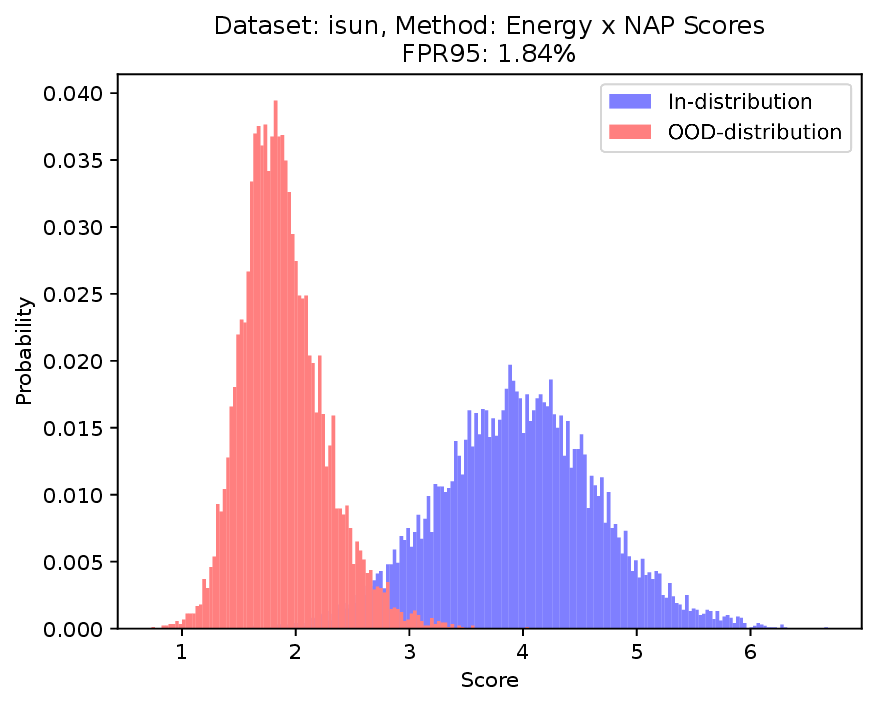}
    \caption{Energy $\times$ NAP Score}
    \label{fig:hist-c}
  \end{subfigure}
  \caption{\textbf{Score distribution visualization using DenseNet on CIFAR-10 (ID) and iSun (OOD)}
  The integration of (a) Energy Score and (b) NAP Score through multiplication yields the (c) Energy $\times$ NAP Score, demonstrating superior differentiation between ID and OOD datasets. The effectiveness of this approach is attributed to the orthogonal nature of the proposed NAP relative to conventional OOD detection methods exemplified by the Energy Score. This illustrates that a simple multiplicative combination with NAP enhances detection capability. Importantly, the objective is not merely to surpass the performance of the Energy Score itself, but to underscore the synergistic potential of NAP as a complementary enhancement to the Energy Score and similar method.}
  \label{fig:score-hists}
  % \vspace{-15pt}
\end{figure}

Furthermore, in this paper, we propose a simple yet effective scoring function for OOD detection based on our prior NAP. To be precise, our scoring function is based on the ratio of the maximal and averaged activation values within a channel.
The rationale behind the scoring function can be understood from two primary perspectives: conceptual inspiration from the Signal-to-Noise (SNR) and empirical validation.
On the conceptual front, inspired by the concept of SNR, we can consider the maximal activation value as signal strength, while the averaged activation value represents noise strength. Therefore, the ratio of the maximum to the average value can be used to measure the quality of information contained in the channel. On the empirical front, as shown in Figure~\ref{fig:one}, the ratio of the maximal and the  averaged values for ID samples is significantly higher than that of OOD data. Regarding practical deployment, as previously mentioned, our scoring function complements and, when multiplied with existing metrics, improves OOD detection, as depicted in Figure~\ref{fig:score-hists}. Also, it is noteworthy that this scoring function is a plug-and-play method, requiring no additional training, extra data, or reliance on pre-calculated statistical data from the training set, which makes it broadly applicable.

Experimental results show that our method achieves state-of-the-art performance on CIFAR-10~\cite{krizhevsky2009cifar}, CIFAR-100~\cite{krizhevsky2009cifar} and ImageNet datasets. Specifically, our method significantly reduces the false positive rate by 48.23\% (from 15.05 to 7.79) on the CIFAR-10 dataset~\cite{krizhevsky2009cifar}, a reduction of 37.89\% (from 41.40 to 25.71) on the CIFAR-100 dataset~\cite{krizhevsky2009cifar}, and a reduction of 16.26\% (from 35.66 to 29.86) on the ImageNet dataset. The large drop in FPR95 highlights the effectiveness of our approach in different environments.
The above experimental results demonstrate the power of our proposed prior. We believe that these findings will provide inspiration for other researchers and thus promote progress in the field of OOD detection.

Additionally, while convolutional neural networks (CNNs) are predominantly employed in OOD detection tasks, the advent of Transformer architectures~\cite{vaswani2017attention} and their variants has showcased substantial efficacy across a diverse array of applications. Motivated by this, we extend our method to ensure compatibility with Transformer models. The empirical evidence obtained from our experiments affirms the robustness of our approach, demonstrating its adaptability to various architectural paradigms.

In summary, our contributions are as follows:
\begin{itemize}
\item We introduce the Neural Activation Prior (NAP), a novel contribution to OOD detection. Uniquely, NAP is orthogonal to priors utilized in existing methods, offering a distinct and complementary perspective that paves the way for advanced OOD detection research.
\item Based on the proposed prior, we develop a simple yet effective OOD detection scoring function. 
% It is a plug-and-play approach that does not impair the model's inherent capabilities. 
It can be readily integrated with many existing OOD detection techniques, enhancing their ability to balance OOD detection with ID accuracy.
\item We demonstrate the state-of-the-art performance of our approach through extensive experiments across various datasets, including a reduction in FPR95 by up to 48.23\%. These results underscore the method's operational efficiency, simplicity of deployment, and overall efficacy. 
\item We extend the method to accommodate Transformer architectures. Experimental results are encouraging, validating the method's efficacy across various architectural designs.
\end{itemize}
\section{Related work}

\subsection{OOD detection}
The OOD detection community has explored a variety of techniques to underscore the distinctions between ID and OOD samples. These methods encompass classification-based~\cite{huang2021mos_confi,liang2018odin_confi,bendale2016towards_confi,devries2018learning_confi,hendrycks2016baseline_confi, sastry2020gram, tack2020csi, du2022vos}, density-based~\cite{zong2018deep_densi,abati2019latent_densi,nalisnick2018deep_densi,zisselman2020deep_densi,jiang2021revisiting_densi,pidhorskyi2018generative_densi,kirichenko2020normalizing_densi,sabokrou2018adversarially_densi}, and distance-based approaches~\cite{sun2022out_dist,lee2018maha_dist,ming2022cider_dist,chen2020boundary_dist,zaeemzadeh2021out_dist,van2020uncertainty_dist,techapanurak2020hyperparameter_dist,lu2023uncertainty_dist, sun2022knn}, with classification-based techniques generally outperforming the other types~\cite{yang2021generalized_survey}. In classification-based methods, the basic work of OOD detection starts with a simple and effective baseline: using the Maximum Softmax Probability (MSP)~\cite{hendrycks2016baseline_confi} to measure the probability that a certain sample is an ID sample.
On this basis, early approaches~\cite{liang2018odin_confi, hsu2020generalized, liu2020energy} focused on developing enhanced OOD indicators derived from neural network outputs. In addition, some researchers have proposed strategies involving OOD sample generation~\cite{lee2018training,du2022vos} and gradient-based~\cite{liang2018odin_confi} techniques. Among these, certain post-hoc methods~\cite{hendrycks2016msp, liang2018odin_confi,liu2020energy, sun2021react, sun2022dice, djurisic2022extremely, yu2023block,du2022vos} are notable for their simplicity and because they do not necessitate changes in the training process or objectives. This feature is particularly valuable for implementing OOD detection in real production environments, where the additional cost and complexity associated with retraining would be unacceptable.

The MSP method, initially presented by Hendrycks et al.~\cite{hendrycks2016msp}, was a formative step in post hoc OOD detection, using a neural network’s softmax output as a heuristic for distinguishing ID from OOD samples. Its straightforward application facilitated early adoption in OOD studies.
Despite MSP's influence, its limitations prompted further innovation, giving rise to the Energy method. This method, proposed by Liu et al.~\cite{liu2020energy}, refines the approach by assigning an energy score to network outputs, showing quantitative improvements over MSP with theoretical and empirical support.
Advancements in post hoc OOD detection have led to diverse methodological branches stemming from MSP~\cite{hendrycks2016baseline_confi} and Energy~\cite{liu2020energy} paradigms. LINe~\cite{ahn2023line} innovates by reducing neuron-induced noise through the calculation of Shapley values. Yu et al.~\cite{yu2023block} distinguish ID from OOD data by identifying neural network blocks with optimal differentiation based on the norms of their features. DICE~\cite{sun2022dice} improves discrimination by pruning weights in the fully connected layer according to the contribution units make during classification.
On the other end of the spectrum, entirely computation-free post hoc methods such as ReAct~\cite{sun2021react} and ASH~\cite{djurisic2022extremely} have shown promise. ReAct~\cite{sun2021react} investigates activations prior to the fully connected layer, applying rectification to suppress extreme activations that OOD data tend to trigger, thereby achieving refined detection outcomes. Similarly, ASH~\cite{djurisic2022extremely} prunes the activations inputted to the fully connected layer, but it achieves even more enhanced results compared to DICE~\cite{sun2022dice} by its selective pruning strategy. In this paper, our comparison mainly focuses on post hoc methods, since our method also belongs to this category.
\section{Neural activation prior}
\label{sec:nap}
Our NAP is based on the following observation for OOD detection:
for a channel located before the global pooling layer in a fully trained neural network, the likelihood that a small number of its neurons activated with a stronger response to an ID sample is significantly higher compared to an OOD sample. For the behavior in other layers of the neural network, refer to the discussion around Figure~\ref{fig:max-mean} in Section~\ref{sec:does}. 

\begin{figure}[!h]
% \vspace{-9pt}
  \centering
  \begin{subfigure}{0.49\linewidth}
    \includegraphics[width=\linewidth]{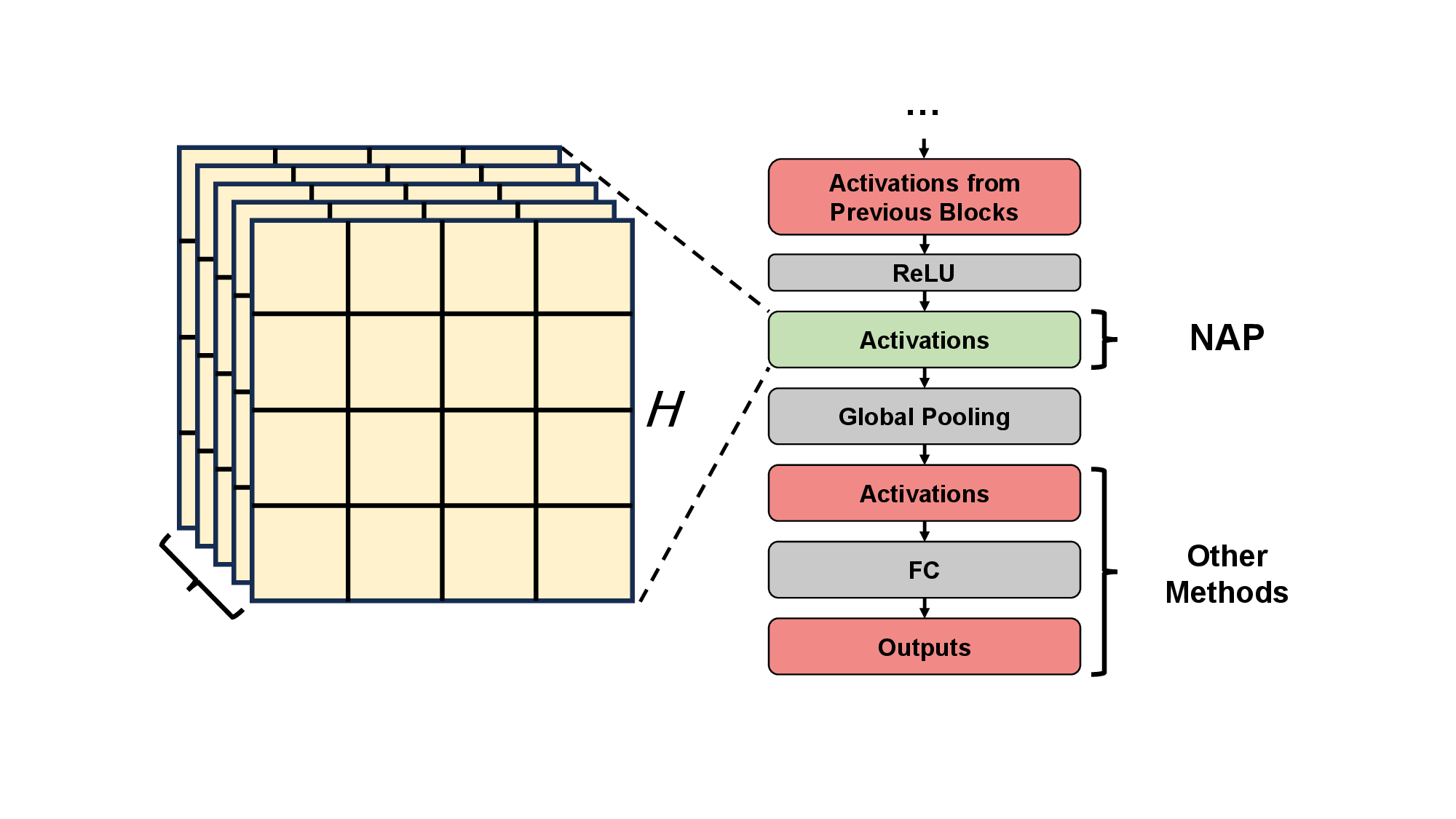}
    \caption{NAP in CNN}
    \label{fig:pos}
  \end{subfigure}
  \hfill
  \begin{subfigure}{0.49\linewidth}
    \includegraphics[width=\linewidth]{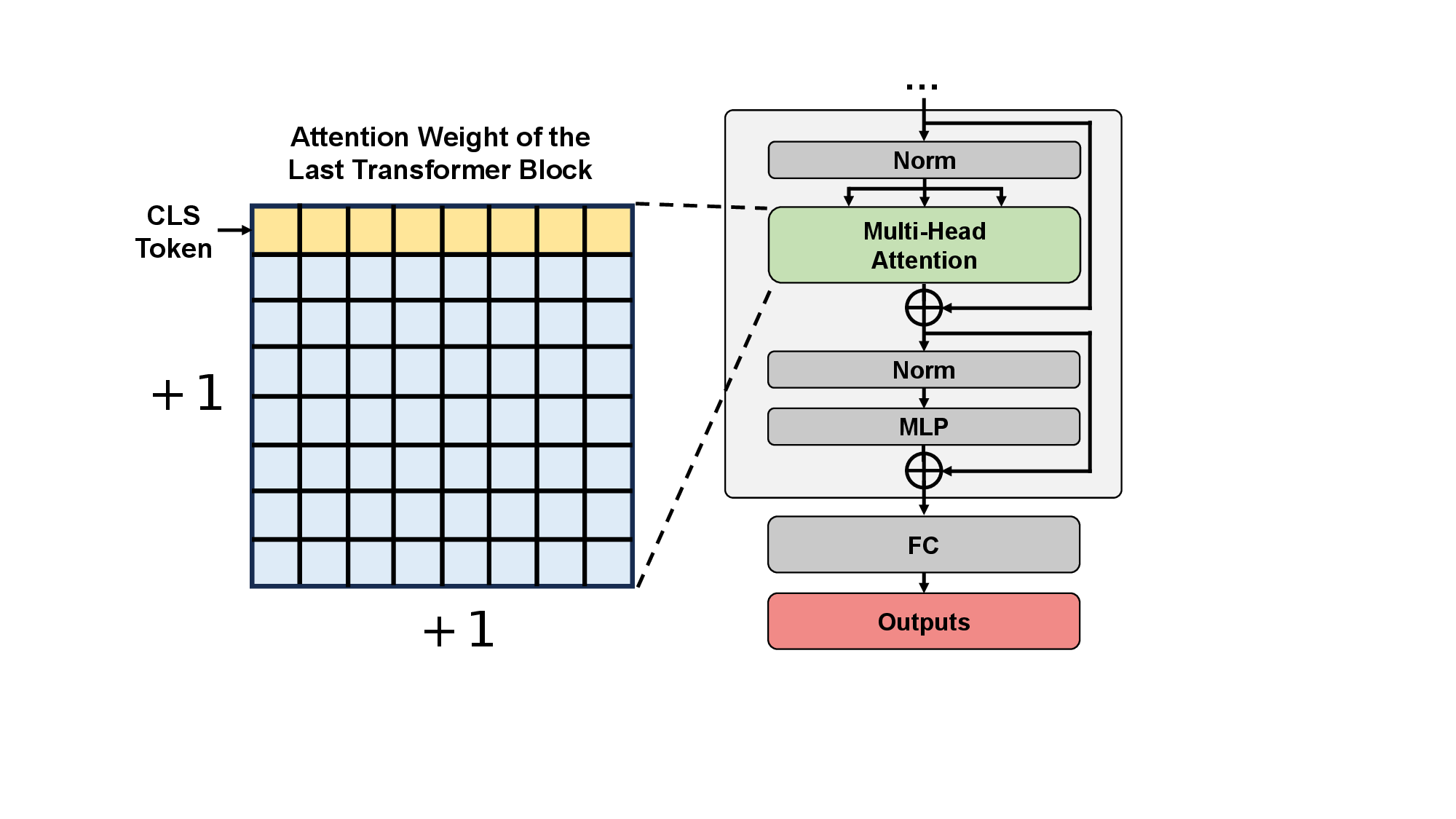}
    \caption{NAP in Transformer}
    \label{fig:vit}
  \end{subfigure}
  \caption{\textbf{Illustration identifying the focus zone of Neural Activation Prior (NAP) in classification neural networks.} The figure highlights the specific position of NAP in the network: (a) For the CNN-based model, the activation value in the green part is the location of the NAP proposed in this article, which is before global pooling. (b) For the Transformer model, NAP is highlighted within the cls token's attention weights in the last Transformer block, illustrating a targeted approach in contrast to most existing methods that focus on regions after global pooling.}
  \label{fig:pos-both}
  % \vspace{-9pt}
\end{figure}

To formally describe this observation, we first define the concept of neural activation. 
Consider a trained classification neural network \( f \), assuming it receives \( D \)-dimensional input \( x \) and outputs \( K \)-dimensional logits. That is \( f: \mathbb{R}^D \rightarrow \mathbb{R}^K \). We concentrate on the activation tensor \( \mathbf{A} \) , located at the penultimate layer just before the global pooling operation, as illustrated in Figure~\ref{fig:pos}. Let the dimensions of \( \mathbf{A} \) be \( C \times H \times W \), where \( C \) is the number of channels, and \( H \) and \( W \) are the spatial dimensions.

Let \( \mathbf{A}_j \) represent the activation tensor of the \( j \)-th channel. We define two key statistical indicators of \( \mathbf{A}_j \) as follows:
\begin{itemize}
\item \textbf{Maximum activation value:}
\begin{equation}
\text{Max}(\mathbf{A}_j) = \max_{k,l} \mathbf{A}_{jkl},
\end{equation}
where \( \text{Max}(\mathbf{A}_j) \) is the maximum value among all elements in the activation vector \( \mathbf{A}_j \). Here, \( k \) and \( l \) index the spatial dimensions (height and width) of the activation map, respectively. Notice that $\mathbf{A}_{jkl}$ is processed by ReLU (as shown in Figure~\ref{fig:pos}), hence $\mathbf{A}_{jkl}\geq0$.
% comment: relu
\item \textbf{Mean activation value:}
\begin{equation}
\text{Mean}(\mathbf{A}_j) = \frac{1}{h \times w} \sum_{k=1}^{h} \sum_{l=1}^{w} \mathbf{A}_{jkl},
\end{equation}
where \( \text{Mean}(\mathbf{A}_j) \) is the average of all activation values in the \( j \)-th channel.

\end{itemize}

Inspired by the concept of signal-to-noise ratio, we interpret \( \text{Max}(\mathbf{A}_j) \) as the signal strength and \( \text{Mean}(\mathbf{A}_j) \) as the noise strength. The ratio of these values can be viewed as an indicator of the quality of the activation vector \( \mathbf{A}_j \). Note that we calculate this metric separately for each channel in the network, since different channels are usually used to detect different patterns. In the context of OOD detection, this quality measure can be used to assess whether the neural network recognizes the input sample --- in other words, this measure can be used to judge whether a sample is within the training set distribution. As shown in Figure~\ref{fig:one}, this phenomenon is termed as the neural activation prior.

It is worth noting that our proposed prior is orthogonal to existing OOD detection methods. As illustrated in Figure~\ref{fig:pos}, existing methods mainly focus on the network output and weight of the penultimate layer after global pooling, and leverage them to design various scoring functions for OOD detection. In contrast, the prior we propose focuses on the channels of the penultimate layer before global pooling. Since the distribution information within these channels is inevitably lost during the global pooling process, our proposed prior is complementary to existing work. 

\noindent \textbf{Extension to Transformer backbones.} We observe that the classify (cls) token in the last block of Transformers can be effectively utilized as an analogue to the pooled activations used in CNNs for our method. Consequently, as illustrated in Figure~\ref{fig:vit}, we calculate NAP score by employing the attention vectors associated with the \textbf{cls} token from the final Transformer block. This approach mirrors our methodological framework in CNNs, facilitating a coherent extension across both architectures. The specifics of the scoring function for NAP, applicable to both CNNs and Transformers, will be further elucidated in Section~\ref{sec:scoring}.
\section{OOD Detection with NAP}
\subsection{Basics}
First, we will provide a brief overview of typical settings for OOD detection in image classification networks. Typically, classification networks use ID data, that is, known training data sets, to train a classification model. Once training is completed, the model's parameters are fixed, enabling it to effectively classify the categories within the training data set.
In the testing phase, to identify OOD samples, researchers usually introduce a scoring function into the model. During the inference, samples mixed with OOD data are fed into this trained model. The model not only classifies each sample, but also uses a scoring function to generate a score for each one. This score is used to predict whether a sample belongs to an ID class in the training set, or an unknown OOD class.

\subsection{Design of scoring function}
\label{sec:scoring}
Based on our prior proposed previously in Section~\ref{sec:nap}, we propose a SNR-like scoring function. In our formulation, the mean activation value is interpreted as the noise intensity, while the maximal activation value is regarded as the signal strength. This conceptual framework leads to the following scoring function:
\begin{equation}
   S_{\textit{NAP}}(x;f)=\frac{1}{C}\sum_{j=1}^{C}\left(\frac{\text{Max}(\mathbf{A}_j)}{\text{Mean}(\mathbf{A}_j) + \epsilon}\right)^2 , 
\end{equation}
% \[ S_{\textit{NAP}}(x;f)=\frac{1}{C}\sum_{j=1}^{C}\left(\frac{\text{Max}(\mathbf{A}_j)}{\text{Mean}(\mathbf{A}_j) + \epsilon}\right)^2 , \]
where \( C \) represents the number of channels before global pooling. Note that a small constant $\epsilon>0$ is added to ensure the numerical stability of the computation.

\noindent\textbf{Scoring function for Transformers.} 
Consistent with the NAP score used in CNNs, we calculate the mean and maximum values of the attention that the cls token has towards all other tokens. 
The attention vector, denoted as $A$, has a dimensionality of 
\((l+1)\), where \(l\) represents the sequence length.
To maintain consistency with the NAP score calculation method used in CNN networks, we would typically divide the maximum value by the mean. However, we note that the mean value of the attention vector is always \(1/(l+1)\), rendering the denominator superfluous. Therefore, for simplicity, we design the NAP score function for Transformers as 
$S_{NAP}^{Former} = \text{Max}(A)$.

\noindent\textbf{OOD detection.}
The usage of scoring function $S_{\textit{NAP}}(x;f)$ in this paper is similar to that of the energy score $-E(x;f)$. The energy scoring function \(E(x; f)\) converts the logits output of the classification network \(f\) into a scalar \(E(x; f) = - \log \sum_{i=1}^{K} e^{f_i(x)}\), where \(f_i(x)\) is the logits output of category \(i\). In OOD detection, the score employed for OOD detection is the negative energy score, \(-E(x; f)\). Therefore, ID data is given a higher score, while OOD data is assigned a lower score.

We can combine these two scoring functions. In this paper, we adopt the weighted geometric mean method to combine them:
% $$S_{\textit{NAP-E}}(x;f,w)=-E(x;f)^w\cdot S_{\textit{NAP}}(x;f)^{1-w}. $$
\begin{equation}
    S_{\textit{NAP-E}}(x;f,w)=-E(x;f)^w\cdot S_{\textit{NAP}}(x;f)^{1-w}. 
\end{equation}
And when using $S_{\textit{NAP}}(x;f)$ to enhance other energy score based OOD methods, we simply replace the function $f$ in the above formula with the specific function of the corresponding method, such as \(f^{\textit{DICE}}\), \(f^{\textit{ReAct}}\), \(f^{\textit{ASH}}\), etc.

\noindent\textbf{How to find a optimal parameter \(w\)?}
\label{sec:how_to_find_w}
When combining NAP with different OOD detection methods, the optimal weight parameter \(w\) varies. To obtain the optimal parameter, we utilized a set of data transformation techniques (such as Gaussian noise, glass blur, motion blur, etc., more details in the Appendix~\ref{appendix:w}) to generate a corrupted dataset based on the ID dataset, serving as pseudo OOD data. For the choice of transformation types, we referred to \cite{hendrycks2018corp}. Utilizing this set of OOD data, we employed a binary search method to find the optimal \(w\). Through experimentation with various datasets and methods, we found that this search approach quickly identifies the optimal \(w\), which generalizes well to real OOD datasets. Refer to Appendix~\ref{appendix:w} for more details about this process.

\noindent\textbf{Discussion.}
\begin{itemize}
    \item \textbf{Plug-and-play simplicity:} The scoring function we proposed is a  plug-and-play approach that can be easily integrated into existing neural network architectures. It requires no additional training or external data and retains the model's inherent classification capabilities. These properties make it practical and suitable for a variety of applications.

    \item \textbf{Orthogonal to existing approaches:} Based on our proposed priors, the scoring function we design is orthogonal to existing methods. As shown in Figure~\ref{fig:one}, the value ranges of the within-channel activation mean values of ID samples and OOD samples overlap. This puts existing methods into trouble when distinguishing between ID and OOD samples with close means values. Based on the prior we proposed, this kind of dilemma can be solved naturally, which illustrates the power of our proposed priors to provide new perspectives for identifying OOD data.
\end{itemize}
\section{Experiments}
\label{sec:exp}
In this section, we conduct experiments on various real-world datasets. In our experiments, we use \textbf{NAP-[initial]} to denote the combination of NAP with another method, where \textbf{[initial]} represents the initial letter of the method's name (e.g., NAP-A for the combination with ASH). Specifically, we combine NAP with a series of common OOD detection methods, including ASH, DICE, Energy, KNN, MSP, and ReAct, denoted as NAP-A, NAP-D, NAP-E, NAP-K, NAP-M, and NAP-R, respectively. CIFAR-10, CIFAR-100, and ImageNet are used as ID datasets. For each combination of NAP with other methods, we use the approach described in Section~\ref{sec:how_to_find_w} to determine the optimal combination parameter \( w \). Detailed optimal \( w \) values for different experimental setups can be found in Appendix~\ref{appendix:w}. It's noted that the value of $\epsilon$ in our scoring functions is consistently set to $1.0$ for numerical stability. All experiments were conducted on an NVIDIA GeForce RTX 3090 GPU.

\subsection{Evaluation on CIFAR benchmarks}
\noindent \textbf{Implementation details.} In our experiments, consistent with recent studies~\cite{sun2021react, sun2022dice, djurisic2022extremely}, we utilized 10,000 test images from both CIFAR-10~\cite{krizhevsky2009cifar} and CIFAR-100~\cite{krizhevsky2009cifar} as ID data. To gauge the performance of the model, six widely-used OOD datasets were employed as benchmarks. These datasets include SVHN~\cite{netzer2011svhn}, Textures~\cite{cimpoi2014texture}, iSUN~\cite{xu2015isun}, LSUN-Crop~\cite{yu2015lsun}, LSUN-Resize~\cite{yu2015lsun}, and Places365~\cite{zhou2017places}. As for pre-trained model, we employed DenseNet~\cite{huang2017densely}, and we follow the training setting of DenseNet introduced in~\cite{sun2022dice}. 

\begin{table}[!b]
\centering
\caption{\textbf{Comparison with competitive post-hoc OOD detection method on CIFAR benchmarks.} All values are percentages and are averaged over 6 OOD test datasets.  Note: A. = Area Under the ROC Curve; F. = False Positive Rate at 95\% True Positive Rate. Methods include ASH~\cite{djurisic2022extremely}, DICE~\cite{sun2022dice}, Energy~\cite{liu2020energy}, KNN~\cite{sun2022knn}, MSP~\cite{hendrycks2016msp}, and ReAct~\cite{sun2021react}.\\}
\label{tab:cifar-avg}
\scriptsize
\resizebox{\textwidth}{!}{
\begin{tabular}{@{}cc|c
c| c
c| c
c| c
c| c
c| c
c @{}}
\toprule
\textbf{Method}                      &    & ASH & \textbf{NAP-A} & DICE  & \textbf{NAP-D} & Energy & \textbf{NAP-E} & KNN   & \textbf{NAP-K} & MSP   & \textbf{NAP-M} & ReAct & \textbf{NAP-R} \\ \midrule
                                     & F.~$\downarrow$ & 15.05 & \textbf{11.14}    & 20.83 & \textbf{11.66} & 26.55  & \textbf{9.02}  & 16.12 & \textbf{7.79}  & 48.69 & \textbf{19.09} & 26.45 & \textbf{9.18}  \\
\multirow{-2}{*}{\textbf{CIFAR-10}}  & A.~$\uparrow$ & 96.91 & \textbf{97.48}    & 95.24 & \textbf{97.47} & 94.67  & \textbf{98.15} & 96.79 & \textbf{98.38} & 92.52 & \textbf{95.11} & 94.67 & \textbf{98.02} \\ \midrule
                                     & F.~$\downarrow$ & 41.40 & \textbf{35.40}    & 49.72 & \textbf{32.34}      & 68.45  & \textbf{32.61} & 44.91 & \textbf{33.63} & 80.13 & \textbf{48.20}      & 62.27 & \textbf{25.71} \\
\multirow{-2}{*}{\textbf{CIFAR-100}} & A.~$\uparrow$ & 90.02 & \textbf{91.21}    & 87.23 & \textbf{92.23}      & 81.19  & \textbf{92.84} & 86.58 & \textbf{91.54} & 74.36 & \textbf{88.45}      & 84.47 & \textbf{93.18} \\ \bottomrule
\end{tabular}
}
\end{table}

\noindent \textbf{Experimental results.} Table \ref{tab:cifar-avg} presents the comparison of NAP combined with other post hoc OOD detection methods on the CIFAR-10 and CIFAR-100 benchmarks. As shown in the table, our approach significantly enhances the performance of all methods on both CIFAR-10 and CIFAR-100 datasets. Notably, the maximum reductions in FPR95 on CIFAR-10 and CIFAR-100 are 66.03\% (NAP-E) and 58.71\% (NAP-R), respectively. Note that the table showcases the average performance across six OOD datasets; for complete performance details, please refer to Appendix~\ref{appendix:cifar}.

\begin{table}[!t]
\scriptsize
\centering
\caption{\textbf{OOD detection results on ImageNet-1k~\cite{deng2009imagenet}.} All values are percentages. Methods include ASH~\cite{djurisic2022extremely}, DICE~\cite{sun2022dice}, Energy~\cite{liu2020energy}, KNN~\cite{sun2022knn}, MSP~\cite{hendrycks2016msp}, and ReAct~\cite{sun2021react}.\\ }
\label{tab:imagenet}
\resizebox{\textwidth}{!}{
\begin{tabular}{LcccccccccR}
\toprule
\multicolumn{1}{c}{\multirow{3}{*}{\textbf{Method}}} & \multicolumn{8}{c}{\textbf{OOD Datasets}}                                                                                                                                                                                                                                                             & \multicolumn{2}{c}{\multirow{2}{*}{\textbf{Average}}} \\ \cmidrule(lr){2-9}
\multicolumn{1}{c}{}                                 & \multicolumn{2}{c}{\textbf{iNaturalist}~\cite{van2018inaturalist}} & \multicolumn{2}{c}{\textbf{SUN}~\cite{xiao2010sun}} & \multicolumn{2}{c}{\textbf{Places}~\cite{zhou2017places}} & \multicolumn{2}{c}{\textbf{Textures}~\cite{cimpoi2014texture}} & \multicolumn{2}{c}{} \\ \cmidrule(l){10-11} 
\multicolumn{1}{c}{}                                 & FPR95~$\downarrow$          & AUROC~$\uparrow$                              & FPR95~$\downarrow$                              & AUROC~$\uparrow$                               & FPR95~$\downarrow$                              & AUROC~$\uparrow$                               & FPR95~$\downarrow$                              & AUROC~$\uparrow$                               & FPR95~$\downarrow$          & AUROC~$\uparrow$              \\ \midrule
ASH~\cite{djurisic2022extremely}                    & 31.46          & 94.28                              & 38.45                              & 91.61                              & 51.80                              & 87.56                              & 20.92                              & 95.07                              & 35.66          & 92.13                        \\
\rowcolor[HTML]{C0C0C0}
\textbf{NAP-A}                                       & \textbf{26.26}          & \textbf{95.10}                              & \textbf{32.89}                     & \textbf{92.77}                     & \textbf{48.69}                     & \textbf{87.92}                     & \textbf{11.60}                              & \textbf{97.32}                              & \textbf{29.86} & \textbf{93.28}                     \\ 
DICE~\cite{sun2022dice}                             & 43.09          & 90.83                              & 38.69                              & 90.46                              & 53.11                              & 85.81                              & 32.80                              & 91.30                              & 41.92          & 89.60                              \\
\rowcolor[HTML]{C0C0C0}
\textbf{NAP-D}                                       & \textbf{27.48}          &  \textbf{94.13}                             &  \textbf{36.14}                    &  \textbf{90.66}                    &  \textbf{51.84}                    &  \textbf{85.03}                    &   \textbf{9.02}                            &  \textbf{97.92}                             & \textbf{31.12} &  \textbf{91.94}                    \\ 
Energy~\cite{liu2020energy}                          & 59.50          & 88.91                              & 62.65                              & 84.50                              & 69.37                              & 81.19                              & 58.05                              & 85.03                              & 62.39          & 84.91                              \\
\rowcolor[HTML]{C0C0C0}
\textbf{NAP-E}                                       & \textbf{29.90}          & \textbf{94.47}                              & \textbf{39.69}                              & \textbf{90.46}                              & \textbf{55.17}                              & \textbf{85.15}                              & \textbf{11.74}                              & \textbf{97.28}                              & \textbf{34.12}          & \textbf{91.84}    \\ 
KNN~\cite{sun2022knn}                          & 85.91          & 72.67                              & 90.49                              & 65.39                              & 93.18                              & 60.08                              & 14.08                              & 96.98                              & 70.92          & 73.78                              \\
\rowcolor[HTML]{C0C0C0}
\textbf{NAP-K}                                       & \textbf{38.23}          & \textbf{89.80}                              & \textbf{56.55}                     & \textbf{80.01}                     & \textbf{70.89}                     & \textbf{71.35}                     & \textbf{7.02}                              & \textbf{98.39}                              & \textbf{43.17} & \textbf{84.89}                     \\ 
MSP~\cite{hendrycks2016msp}                          & 64.29          & 85.32                              & 77.02                              & 77.10                              & 79.23                              & 76.27                              & 73.51                              & 77.30                              & 73.51          & 79.00           \\
\rowcolor[HTML]{C0C0C0}
\textbf{NAP-M}                                       & \textbf{35.47}          & \textbf{92.53}                              & \textbf{51.19}                     & \textbf{86.51}                     & \textbf{63.77}                     & \textbf{80.61}                     & \textbf{15.14}                              &  \textbf{97.09}                             & \textbf{41.39} & \textbf{89.19}                     \\ 
ReAct~\cite{sun2021react}                           & 42.40          & 91.53                              & 47.69                              & 88.16                              & 51.56                              & 86.64                              & 38.42                              & 91.53                              & 45.02          & 89.47                              \\
\rowcolor[HTML]{C0C0C0}
\textbf{NAP-R}                                       & \textbf{24.58} & \textbf{95.55}                     & \textbf{38.47}                               & \textbf{91.12}                             & \textbf{53.32}                              & \textbf{86.24}                              & \textbf{9.57}                      & \textbf{97.60}                     & \textbf{31.49}          & \textbf{92.63}                     \\  \bottomrule
\end{tabular}
}
\end{table}

\subsection{Evaluation on ImageNet}
\label{sec:imagenet}
\noindent \textbf{Implementation details.} In real-world applications, models are confronted with high-resolution images spanning a diverse range of scenes and features. Evaluations on large-scale datasets can provide insights into the performance of models in practical deployments. Thus, in line with recent research~\cite{sun2021react, sun2022dice, djurisic2022extremely}, we conduct experiments with NAP on the expansive ImageNet-1k~\cite{deng2009imagenet} dataset in this study. Four dataset subsets, with all overlapping categories with ImageNet-1k eliminated, were employed as OOD benchmarks. These OOD datasets comprise Textures~\cite{cimpoi2014texture}, Places365~\cite{zhou2017places}, iNaturalist~\cite{van2018inaturalist}, and SUN~\cite{xiao2010sun}. We used MobileNetV2~\cite{sandler2018mobilenetv2} architecture, which pre-trained on ImageNet-1k~\cite{deng2009imagenet}. The architecture and parameters remain unchanged during the OOD detection stage.
% Parallel to the experiments conducted on the CIFAR~\cite{krizhevsky2009cifar} benchmarks, we experimented with three distinct implementations. For parameter settings, $w$ was set to $0.6$ for \textbf{NAP-E}, $0.8$ for \textbf{NAP-A}, and $0.8$ for \textbf{NAP-R}. 

\noindent \textbf{Experimental results.} Table \ref{tab:imagenet} shows the comparison results of NAP with other post-hoc OOD detection methods on the ImageNet-1k~\cite{deng2009imagenet} benchmark. As shown in the table, our approach significantly enhances the performance of all methods on the ImageNet-1k dataset~\cite{deng2009imagenet}. We have observed that our method brings significant performance improvements on the iNaturalist and Textures datasets, which is particularly notable. Intuitively, most of the pictures in the Textures~\cite{cimpoi2014texture} dataset are of a texture nature, so there is a small probability of obtaining a large response value. And while samples in iNaturalist have a relatively simple background, and the animals and plants in the foreground have certain semantic differences from the samples in ImageNet-1k, therefor, a sample in iNaturalist~\cite{van2018inaturalist} is not likely to trigger large response values. The reason why existing methods do not perform well on this dataset may be due to their focus on activation values after global pooling. We speculate that although texture pictures cannot cause large activation values, they can excite small-amplitude noise on the entire feature map. Since large response values tend to only appear in a small area on the feature map, after global pooling, ID samples and texture class samples are likely to obtain similar average activation values. This may compromise the separability of the two types of samples, causing existing work to perform poorly on such samples.

\subsection{Evaluation on Transformer}
Following the experimental setup described in Section~\ref{sec:imagenet}, we conduct experiments on the Vision Transformer~\cite{dosovitskiy2020image} (ViT-B/16) using ImageNet-1k as the ID dataset. Energy~\cite{liu2020energy} and MSP~\cite{hendrycks2016msp} were selected as baseline methodologies for this analysis. The study further explores the enhancement of these baseline methods through the integration of NAP, resulting in two variants: NAP-E and NAP-M. Comparative results detailed in Table~\ref{tab:vit} demonstrate that the NAP method substantially boosts the performance on Transformer architectures beyond the baselines, affirming the utility and versatility of NAP within such contexts.
\begin{table}[!h]
\scriptsize
\centering
\caption{\textbf{OOD detection results on ViT-B/16 using ImageNet-1k as ID data.} All values are percentages. \\ }

\label{tab:vit}
\resizebox{\textwidth}{!}{
\begin{tabular}{LcccccccccR}
\toprule
\multicolumn{1}{c}{\multirow{3}{*}{\textbf{Method}}} & \multicolumn{8}{c}{\textbf{OOD Datasets}}                                                                                                                                                                                                                                                             & \multicolumn{2}{c}{\multirow{2}{*}{\textbf{Average}}}                   \\ \cmidrule(lr){2-9}
\multicolumn{1}{c}{}                                 & \multicolumn{2}{c}{\textbf{iNaturalist}~\cite{van2018inaturalist}}                                & \multicolumn{2}{c}{\textbf{SUN}~\cite{xiao2010sun}}                                        & \multicolumn{2}{c}{\textbf{Places}~\cite{zhou2017places}}                                     & \multicolumn{2}{c}{\textbf{Textures}~\cite{cimpoi2014texture}}                                   & \multicolumn{2}{c}{}                                                    \\ \cmidrule(l){10-11} 
\multicolumn{1}{c}{}                                 & FPR95~$\downarrow$          & AUROC~$\uparrow$                               & FPR95~$\downarrow$                               & AUROC~$\uparrow$                               & FPR95~$\downarrow$                              & AUROC~$\uparrow$                               & FPR95~$\downarrow$                              & AUROC~$\uparrow$                               & FPR95~$\downarrow$          & AUROC~$\uparrow$                               \\ \midrule
Energy~\cite{liu2020energy}                           & \multicolumn{1}{c}{64.08}          & 79.24                              & 72.77                              & 70.25                              & 74.30                              & 68.44                              & 58.46                              & 79.30                              & \multicolumn{1}{c}{67.40}          & 74.31                              \\
\rowcolor[HTML]{C0C0C0}
\textbf{NAP-E}                                & \textbf{60.97}  & \textbf{80.77}                     & \textbf{64.05}                               & \textbf{77.34}                              & \textbf{69.34}                               & \textbf{73.30}                              & \textbf{45.04}                       & \textbf{86.93}                     & \textbf{59.85} & \textbf{79.58}                     \\  
MSP~\cite{hendrycks2016msp}                            & \multicolumn{1}{c}{51.47}          & 88.16                              & 66.53                              & 80.93                              & 68.65                              & 80.38                              & 60.21                              & 82.99                              & \multicolumn{1}{c}{61.72}          & 83.12                              \\
\rowcolor[HTML]{C0C0C0}
\textbf{NAP-M}            &      \textbf{47.09}    &      \textbf{88.23}        & \textbf{59.45}                              &      \textbf{82.78}    & \textbf{63.38}             & \textbf{80.48}                              &      \textbf{47.70}    &      \textbf{87.93}        &      \textbf{54.40}                              &      \textbf{84.85}    \\ \bottomrule
\end{tabular}
}
\end{table}

\noindent\textbf{Discussion.} We compare only with MSP~\cite{hendrycks2016msp} and Energy~\cite{liu2020energy} because methods such as ASH~\cite{djurisic2022extremely} and ReAct~\cite{sun2021react} are specifically designed for CNNs and fail completely when applied to ViT. For fairness, we do not include these methods in our comparison.

\subsection{Does NAP work across all network layers?}

\label{sec:does}
To assess the efficacy of the NAP across various layers within the DenseNet architecture, we conduct a detailed examination of activation distributions utilizing the CIFAR-10~\cite{krizhevsky2009cifar} and Places365 datasets~\cite{zhou2017places}. Analysis focused on four pivotal points within the network: post-first convolutional layer, pre-pooling in the first and second transition blocks, and immediately preceding the final global pooling layer.
Selected instances from the empirical findings are illustrated in Selected examples from our tests are shown in Figure~\ref{fig:max-mean}, with more detailed visuals in the Appendix~\ref{appendix:layer}. From these figures, it's clear that the deeper the layer, the easier it is to tell apart ID and OOD samples based on the maximum and average activation values. 
In the network's early layers, neurons pick up basic features common to both ID and OOD samples, which explains why the lines cross in Figures~\ref{fig:t1} and~\ref{fig:t2}. Moving to deeper layers, neurons start capturing more complex, meaningful features. As seen in Figures~\ref{fig:t3} and~\ref{fig:t4}, ID samples with specific meanings trigger larger responses. This indicates that NAP works best closer to the end of the network, especially before the global pooling layer, highlighting its value in making models more reliable for OOD detection.

\begin{figure}[h]
  \centering
  \begin{subfigure}{0.24\linewidth}
    \includegraphics[width=0.99\linewidth]{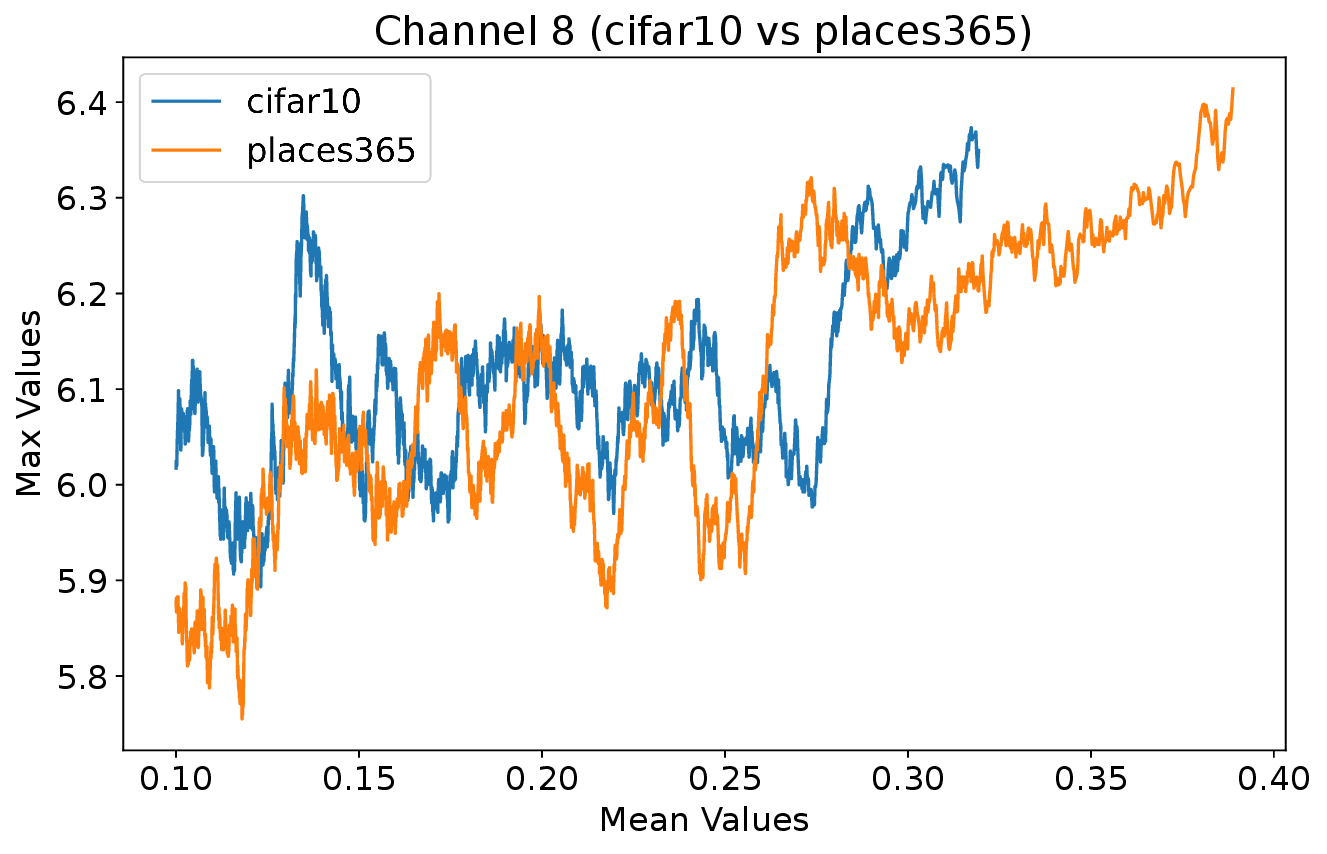}
    \caption{After conv layer 1}
    \label{fig:t1}
  \end{subfigure}
  \hfill
  \begin{subfigure}{0.24\linewidth}
    \includegraphics[width=0.99\linewidth]{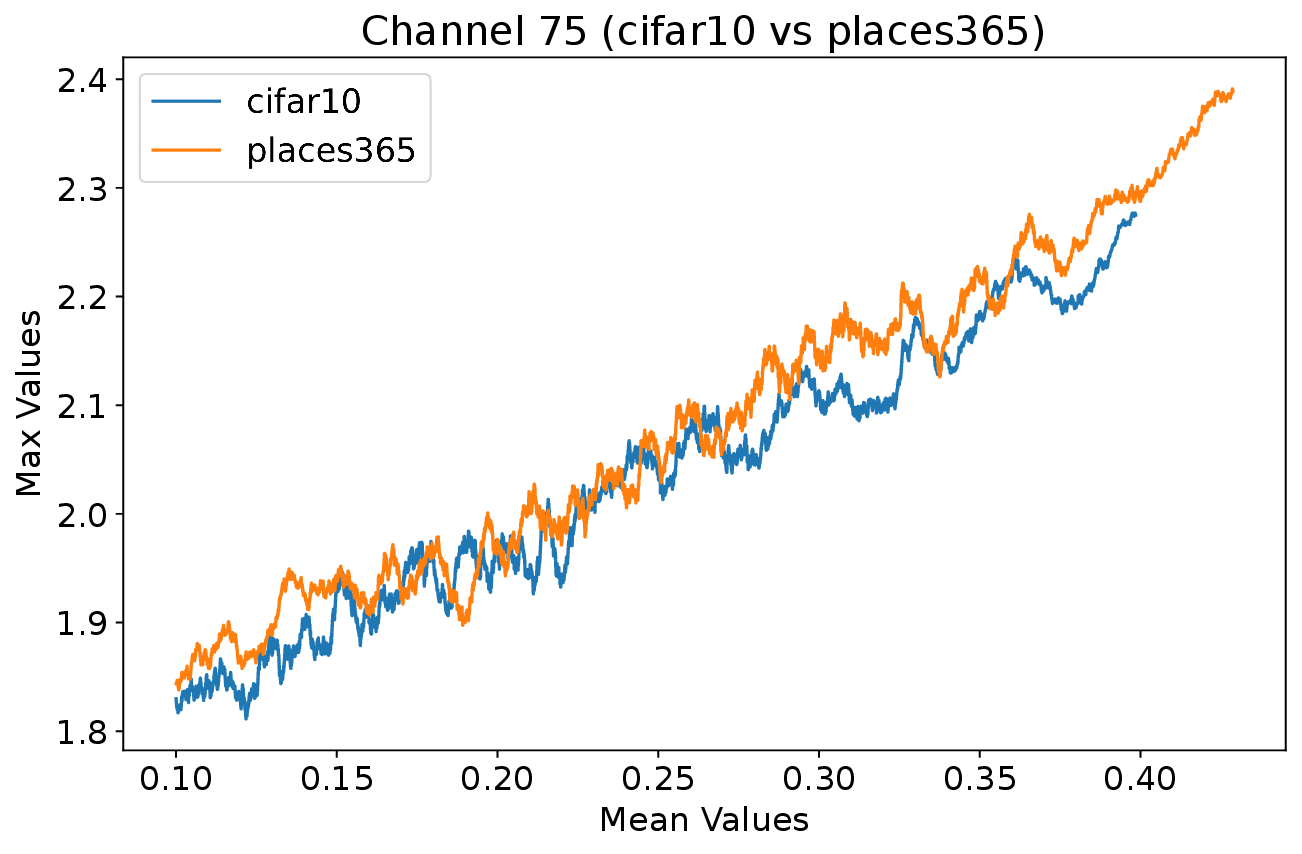}
    \caption{In Trans Block 1}
    \label{fig:t2}
  \end{subfigure}
  \hfill
  \begin{subfigure}{0.24\linewidth}
    \includegraphics[width=0.99\linewidth]{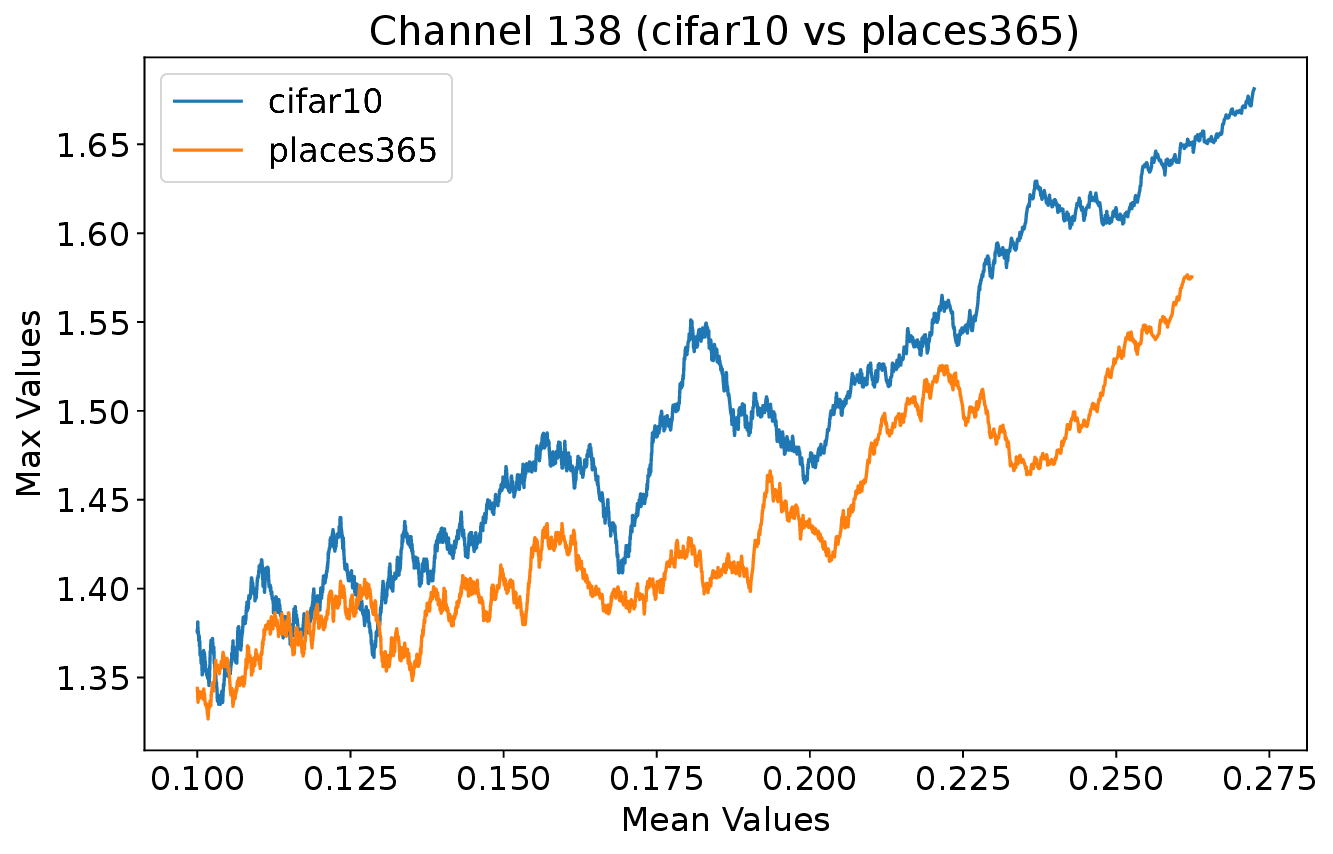}
    \caption{In Trans Block 2}
    \label{fig:t3}
  \end{subfigure}
  \hfill
  \begin{subfigure}{0.24\linewidth}
    \includegraphics[width=0.99\linewidth]{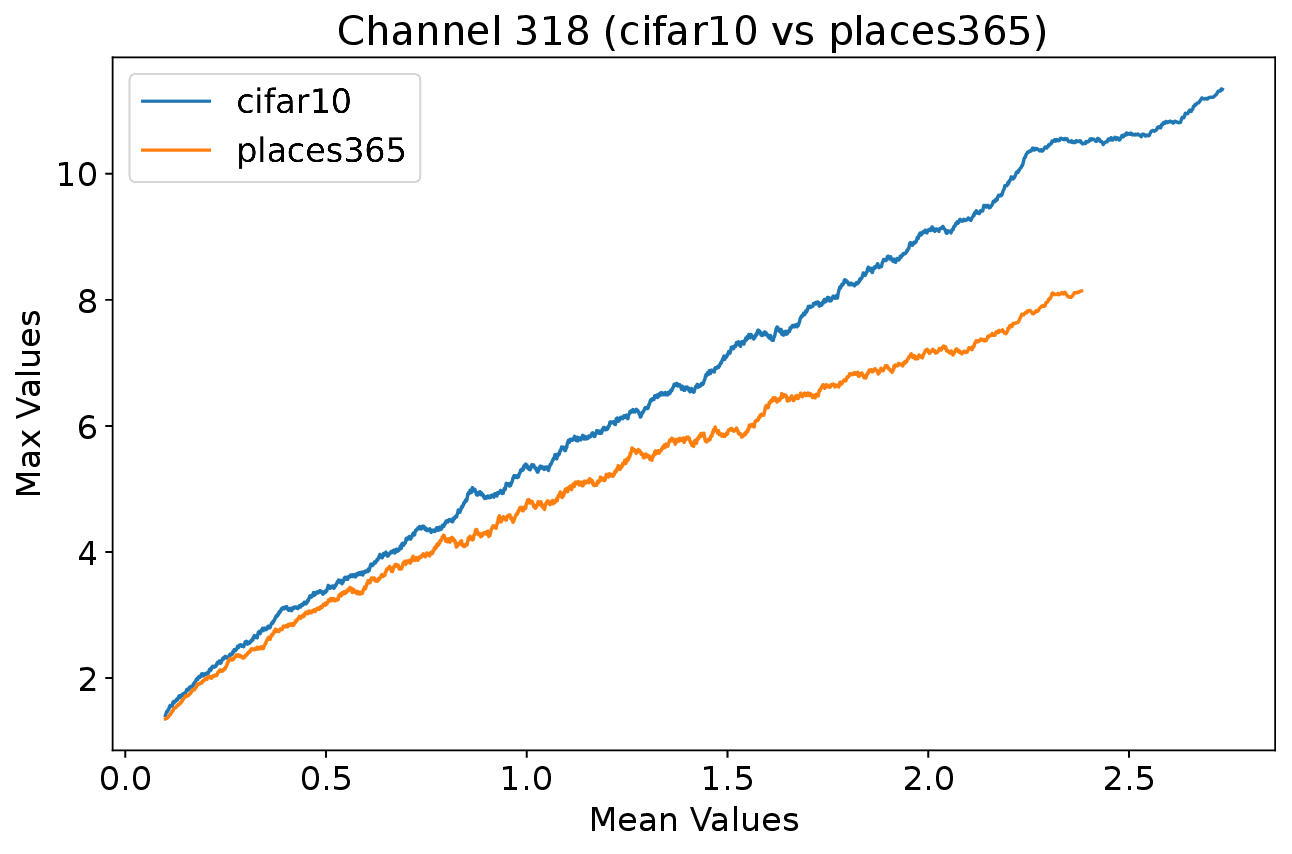}
    \caption{Before Pooling}
    \label{fig:t4}
  \end{subfigure}
  \caption{\textbf{Activation distribution at different positions within the DenseNet architecture~\cite{huang2017densely} applied to CIFAR-10~\cite{krizhevsky2009cifar} and Places365 datasets~\cite{zhou2017places}.} For this analysis, four specific locations within the network were chosen: (a) after the first convolution layer, (b) just before the pooling operation in the first transition block, (c) just before the pooling operation in the second transition block, and (d) right before the final global pooling layer. It is observed that with increasing depth, the separability between ID and OOD samples becomes more pronounced in the two-dimensional space defined by the maximum and average activation values.}
  \label{fig:max-mean}
  % \vspace{-15pt}
\end{figure}
\section{Conclusion}
This paper proposes a novel Neural Activation Prior (NAP) for OOD detection in machine learning models. Our proposed prior is grounded on the observation that in a fully trained neural network, ID samples typically induce stronger activation responses in some neurons of a channel compared to OOD samples. This discovery led to our novel scoring function based on within-channel distribution.
Its main advantage lies in its simplicity and easy integration. It necessitates neither additional training nor external data and does not compromise the classification performance on ID data. Experimental results on various datasets and architectures show that our method achieves state-of-the-art performance in OOD detection. This not only verifies the effectiveness of neural activation priors, but also demonstrates the potential of rethinking the way neural network features are utilized in OOD scenarios.
{
	\small
	\bibliographystyle{abbrv}
	\bibliography{main}
}

\newpage
\appendix
\section*{Appendix}
In this appendix, we provide comprehensive additional materials to supplement the main text. The contents include:

\begin{itemize}
    \item \textbf{Broader impacts (Section~\ref{appendix:impacts}):} A discussion on the broader implications of our research.
    \item \textbf{Pseudo code for NAP (Section~\ref{appendix:pseudo}):} The algorithmic representation of the Neural Activation Prior (NAP) methodology.
      \item \textbf{Why is the penultimate layer more effective for NAP? (Section~\ref{appendix:layer}):} An exploration of the reasons behind the superior effectiveness of the penultimate layer in the context of NAP.
      \item \textbf{Evaluating multi-layer integration with NAP for OOD detection (Section~\ref{appendix:integration}):} Investigation into the effects of integrating multiple layers along with NAP in detecting OOD data.
      \item \textbf{On transferability to other architectures (Section~\ref{appendix:transfer}):} To ascertain the versatility and robustness of NAP across different CNN architectures, we conducted extensive experiments on various backbones, including VGG, DenseNet, and ResNet.
      \item \textbf{Pareto frontier of ID accuracy and OOD detection performance (Section~\ref{appendix:pareto}):}Evaluation on Pareto Frontier of ID accuracy and OOD Detection Performance.
      \item \textbf{Full CIFAR benchmark results: enhancing methods with NAP (Section~\ref{appendix:cifar}):} Comprehensive evaluation of NAP's effectiveness in enhancing existing models, as demonstrated through detailed results on the CIFAR benchmark.
      \item \textbf{How to find an optimal parameter $w$? (Section~\ref{appendix:w}):} A guide on determining the optimal parameter $w$ for NAP.
      \item \textbf{Performance on Near-OOD detection (Section~\ref{appendix:near}):} Investigating the capability of the NAP in distinguishing between closely related datasets provides insight into its utility in nuanced OOD detection scenarios.
      \item \textbf{More examples of activation map visualizaiton. (Section~\ref{appendix:vis}):} Additional visual examples showcasing the activation maps.
      \item \textbf{Limitations (Section~\ref{appendix:limit}):} A critical analysis of the limitations of our approach.
    \item \textbf{Discussion (Section~\ref{appendix:discuss}):} A concluding section that summarizes the key findings and outlines future directions for research based on our work.
    \item \textbf{Licenses for existing assets (Section~\ref{appendix:license}):} Credits and licenses for all existing assets used in this research.
\end{itemize}
%%%%%%%%%%%%%%%%%%%%%%%%%%%%%%%%%%%%%%%%%%%%%%%%%%%%%%%%%%%%%%%%%%%%%%%%%%
\section{Broder impacts}
\label{appendix:impacts}
The proposed Neural Activation Prior for OOD detection has significant implications for the deployment of machine learning models in real-world scenarios. By enhancing the ability to detect OOD samples, our method contributes to improving the reliability and safety of AI systems, particularly in critical applications such as autonomous driving, healthcare, and security, where encountering unexpected inputs could have severe consequences.
%%%%%%%%%%%%%%%%%%%%%%%%%%%%%%%%%%%%%%%%%%%%%%%%%%%%%%%%%%%%%%%%%%%%%%%%%%%
\section{Pseudo code for NAP}
\label{appendix:pseudo}
As illustrated in Algorithm~\ref{algo}, we present a detailed pseudo-code representation of our proposed method for OOD detection, which is integrated into the DenseNet architecture. The key modification involves the calculation of NAP score $\mathcal{S}$ within the DenseNet's processing pipeline (\textcolor{highlightcolor}{highlighted in green font in the algorithm}), which is then followed by calculating the OOD score using the model's logits together with $\mathcal{S}$. These calculations do not alter the logits output by the model, thereby ensuring no degradation in classification accuracy for the ID dataset. 

\begin{algorithm}[!h]
\caption{OOD Detection Using Neural Activation Prior on DenseNet}
\begin{algorithmic}[1]
\REQUIRE Image $x$, Weight $w$ (for OOD score calculation)
\ENSURE Output logits, OOD Score

\STATE Apply initial layers of DenseNet on $x$ to obtain intermediate output:
\STATE \hspace{\algorithmicindent} $out \leftarrow \text{conv1}(x)$
\STATE \hspace{\algorithmicindent} $out \leftarrow \text{trans1}(\text{block1}(out))$
\STATE \hspace{\algorithmicindent} $out \leftarrow \text{trans2}(\text{block2}(out))$
\STATE \hspace{\algorithmicindent} $out \leftarrow \text{block3}(out)$
\STATE \hspace{\algorithmicindent} $out \leftarrow \text{relu}(\text{bn1}(out))$
\STATE \hspace{\algorithmicindent} Let $\mathcal{A}$ denote this intermediate output.

\STATE \textcolor{highlightcolor}{Compute NAP score $\mathcal{S}$ from $\mathcal{A}$:}
\STATE \textcolor{highlightcolor}{\hspace{\algorithmicindent} Flatten $\mathcal{A}$ across dimensions 2 and 3.}
\STATE \textcolor{highlightcolor}{\hspace{\algorithmicindent} Compute Max of $\mathcal{A}$ across flattened dimensions.}
\STATE \textcolor{highlightcolor}{\hspace{\algorithmicindent} Compute Mean of $\mathcal{A}$ over dims 2, 3.}
\STATE \textcolor{highlightcolor}{\hspace{\algorithmicindent} Calculate $\mathcal{S}$ as $(\text{Max of } \mathcal{A} / (\text{Mean of } \mathcal{A} + 1))^2$.}
\STATE \textcolor{highlightcolor}{\hspace{\algorithmicindent} Compute the mean of $\mathcal{S}$ across dimension 1.}

\STATE Continue with DenseNet forward pass:
\STATE \hspace{\algorithmicindent} Apply average pooling and reshape on $\mathcal{A}$.
\STATE \hspace{\algorithmicindent} Get logits from fully connected layer.

\STATE \textcolor{highlightcolor}{Calculate OOD Score:}
\STATE \textcolor{highlightcolor}{\hspace{\algorithmicindent} Compute log-sum-exp of the logits.}
\STATE \textcolor{highlightcolor}{\hspace{\algorithmicindent} Calculate OOD Score as $(\text{log-sum-exp}^w) * (\mathcal{S}^{1-w})$.}

\RETURN Output logits, OOD Score
\end{algorithmic}
\label{algo}
\end{algorithm}

%%%%%%%%%%%%%%%%%%%%%%%%%%%%%%%%%%%%%%%%%%%%%%%%%%%%%%%%%%%%%%%%%%%%%%%%%%%
\begin{figure}[!t]
  \centering
  \begin{subfigure}{\linewidth}
    \includegraphics[width=0.24\linewidth]{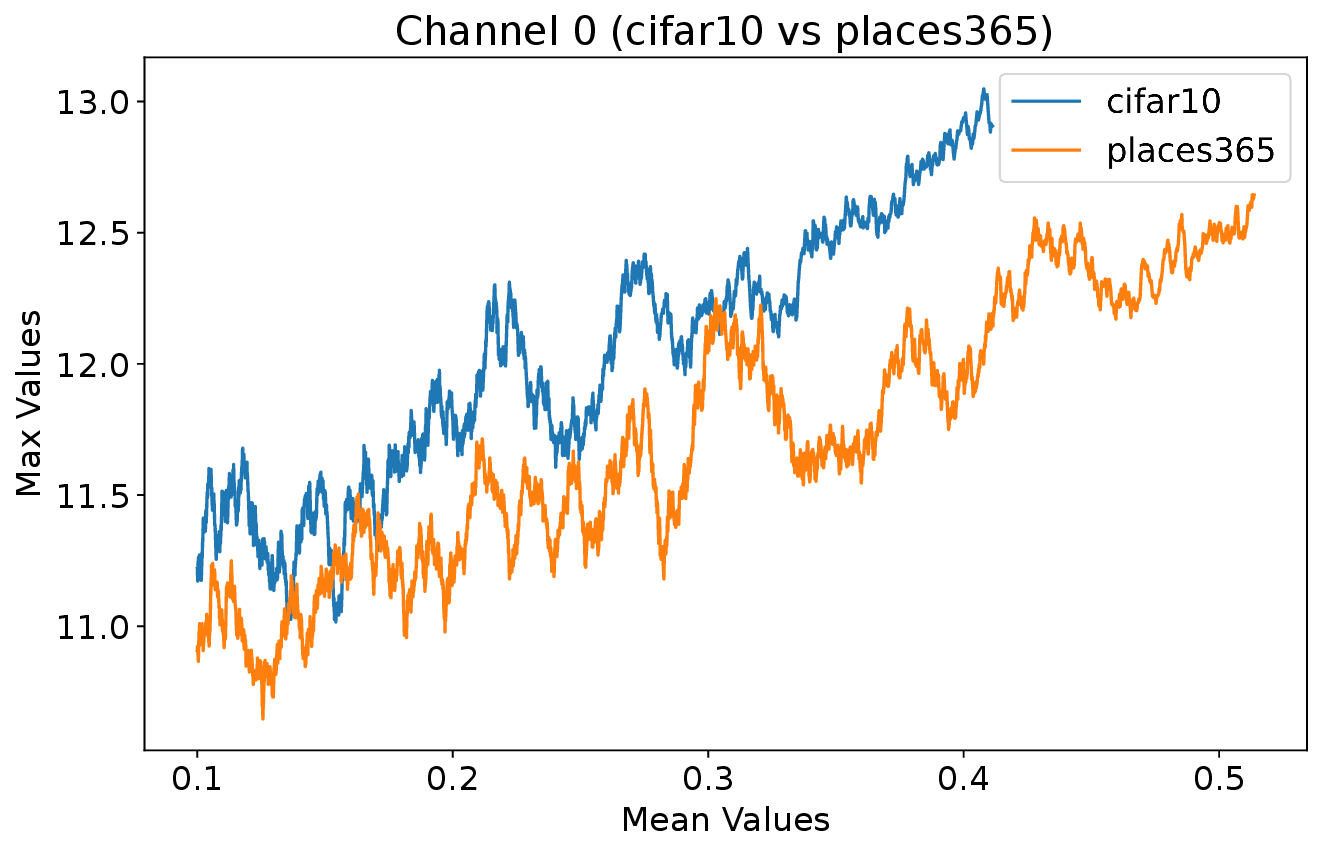}
    \includegraphics[width=0.24\linewidth]{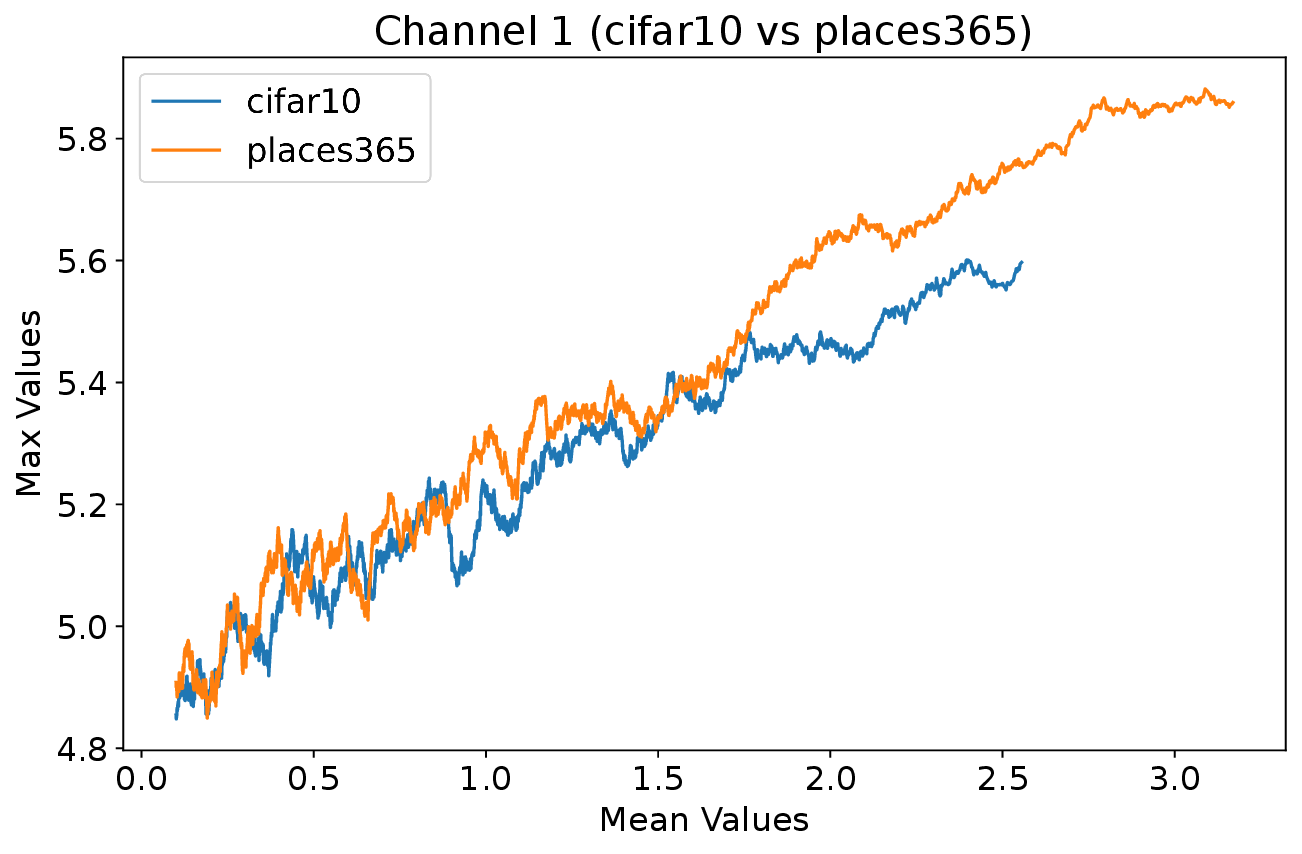}
    \includegraphics[width=0.24\linewidth]{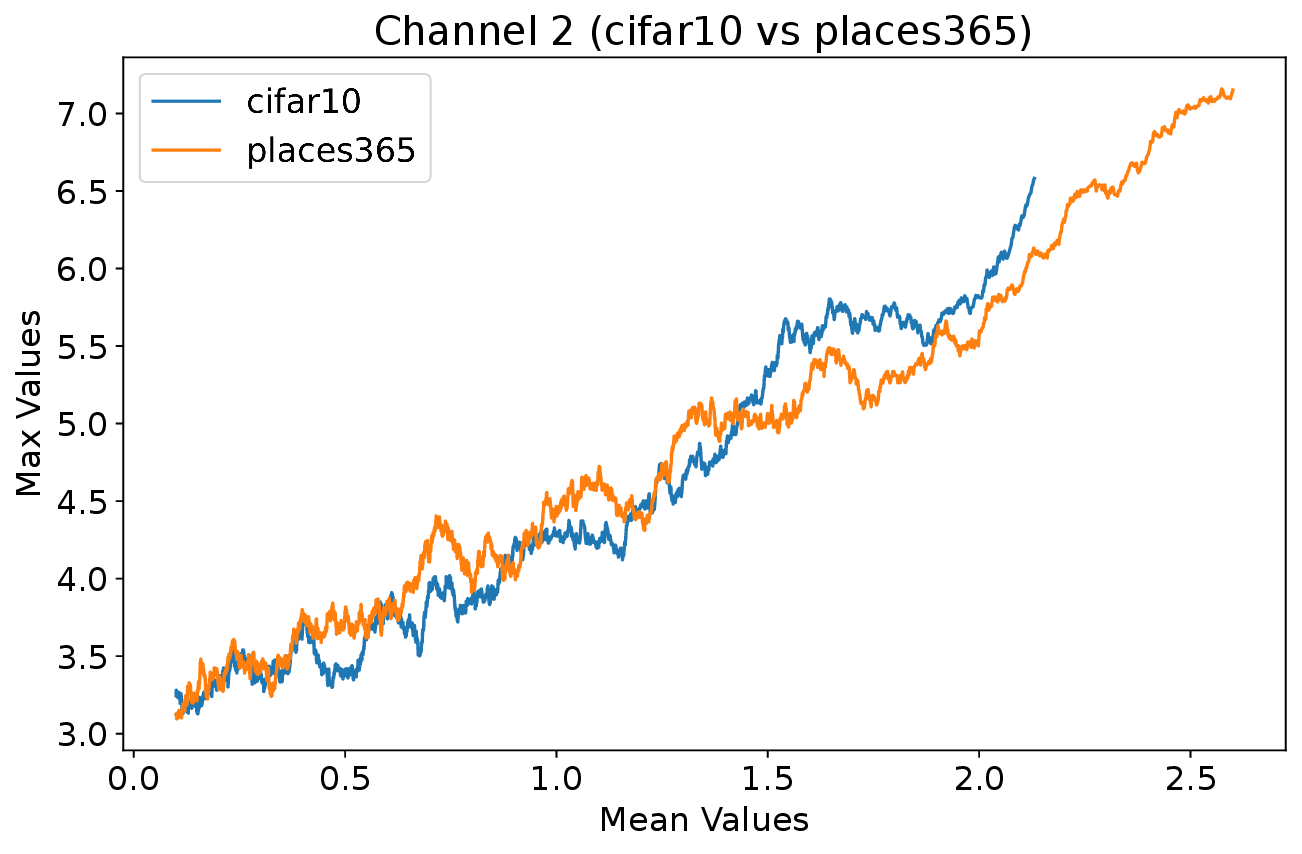}
    \includegraphics[width=0.24\linewidth]{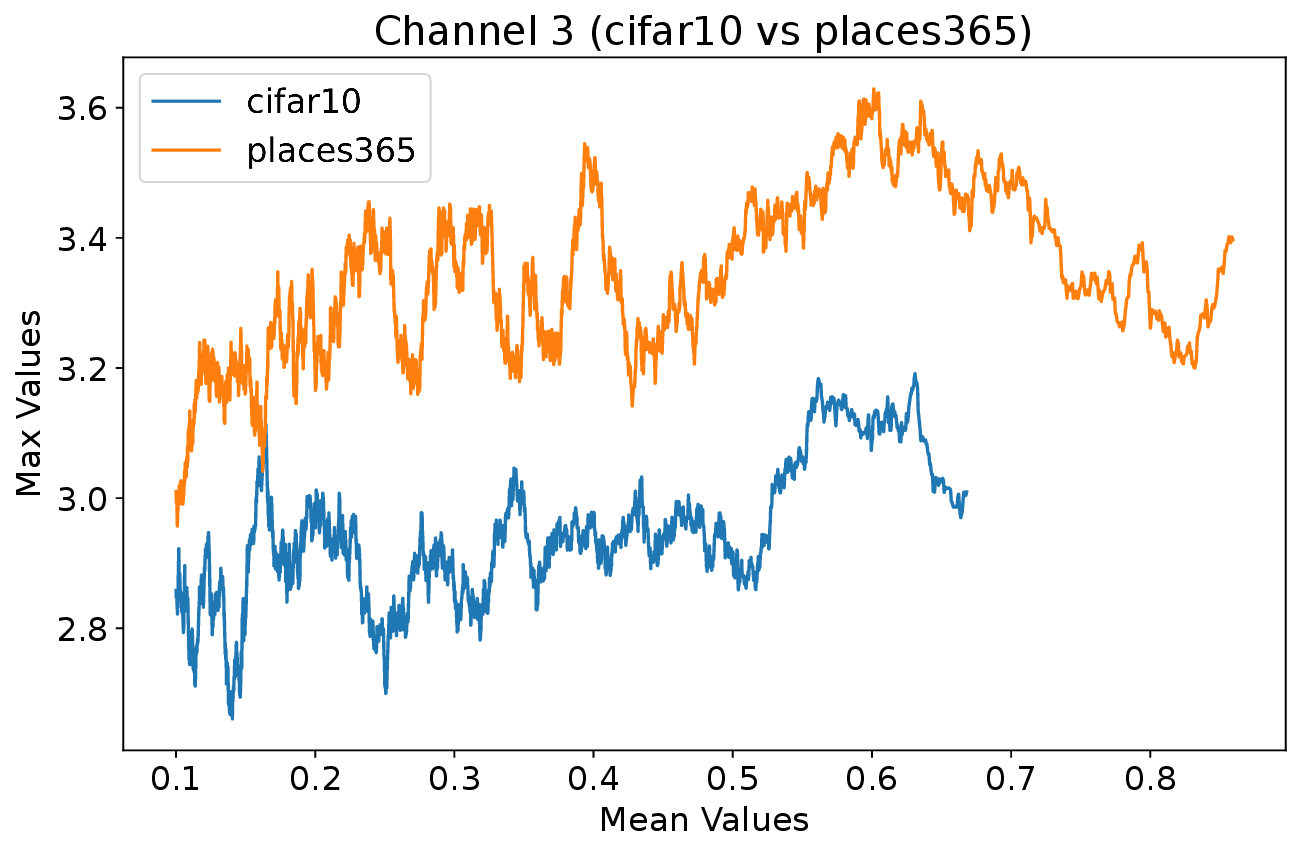}
    \caption{After first convolution layer}
    \label{fig:places3650}
  \end{subfigure}

  \begin{subfigure}{\linewidth}
    \includegraphics[width=0.24\linewidth]{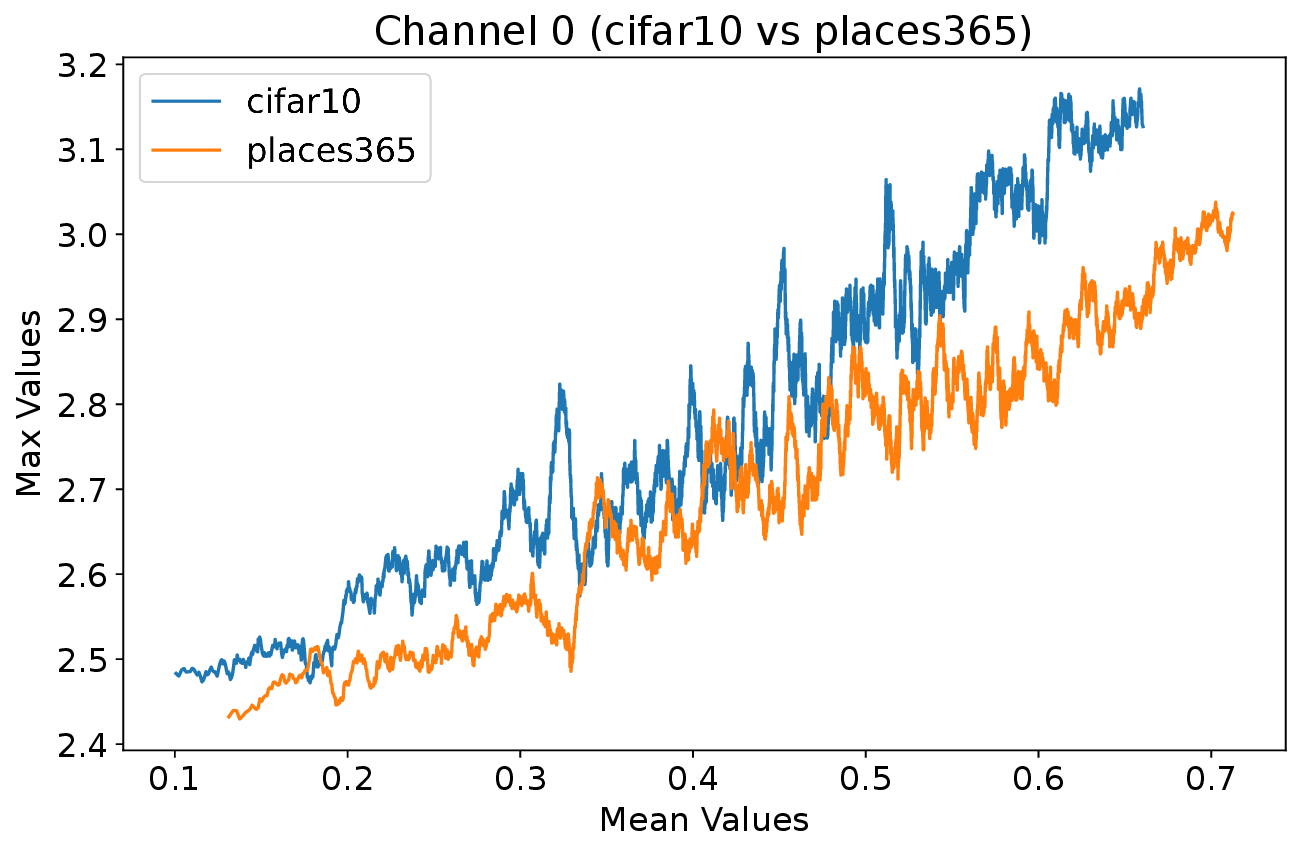}
    \includegraphics[width=0.24\linewidth]{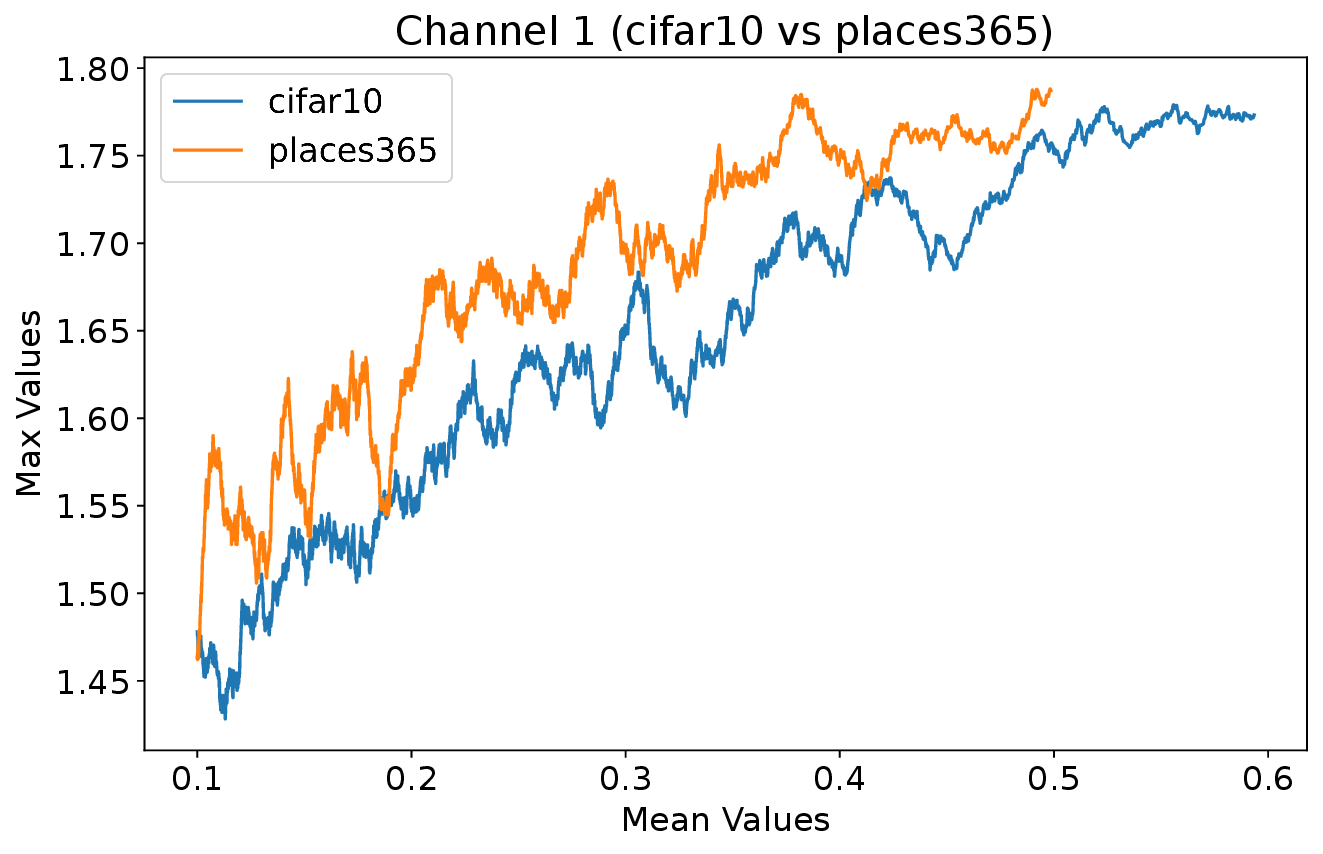}
    \includegraphics[width=0.24\linewidth]{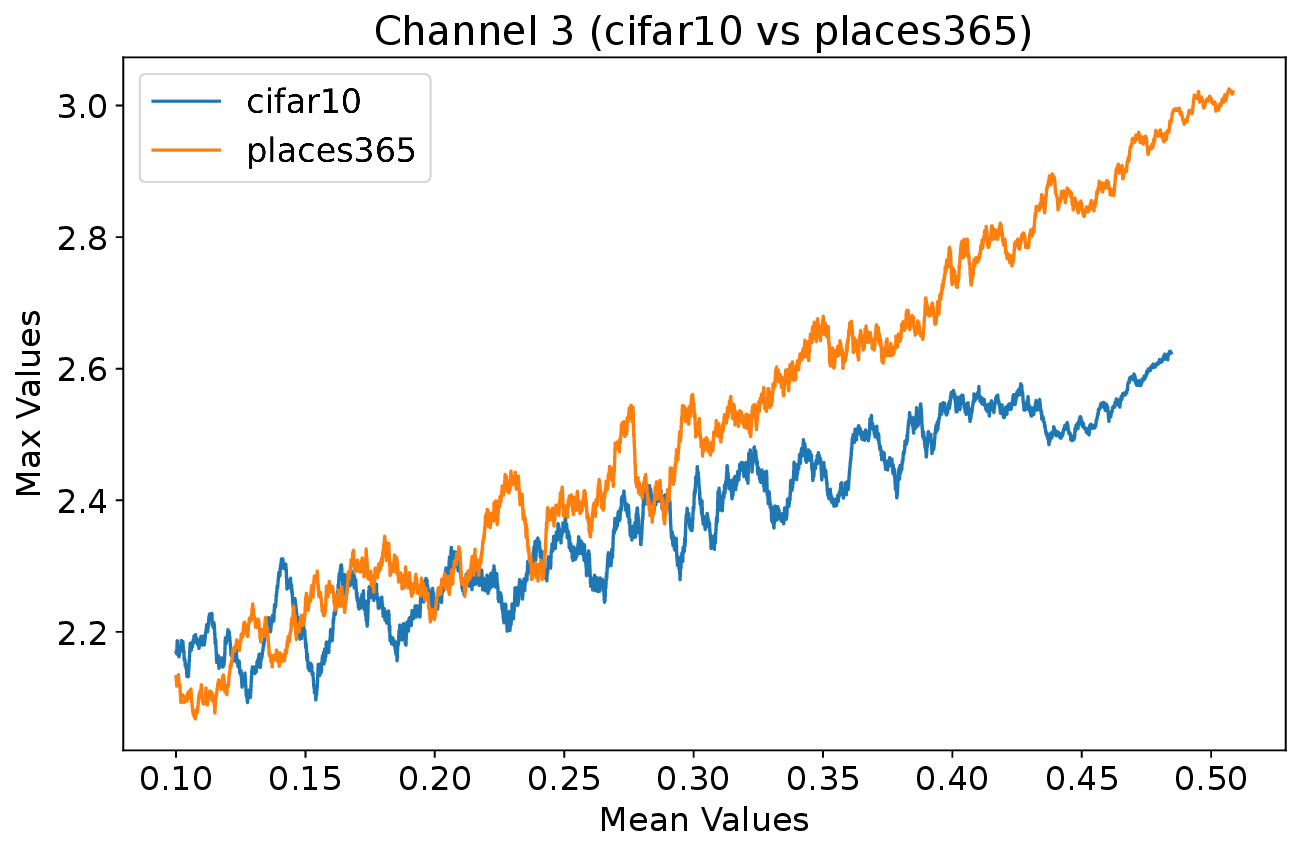}
    \includegraphics[width=0.24\linewidth]{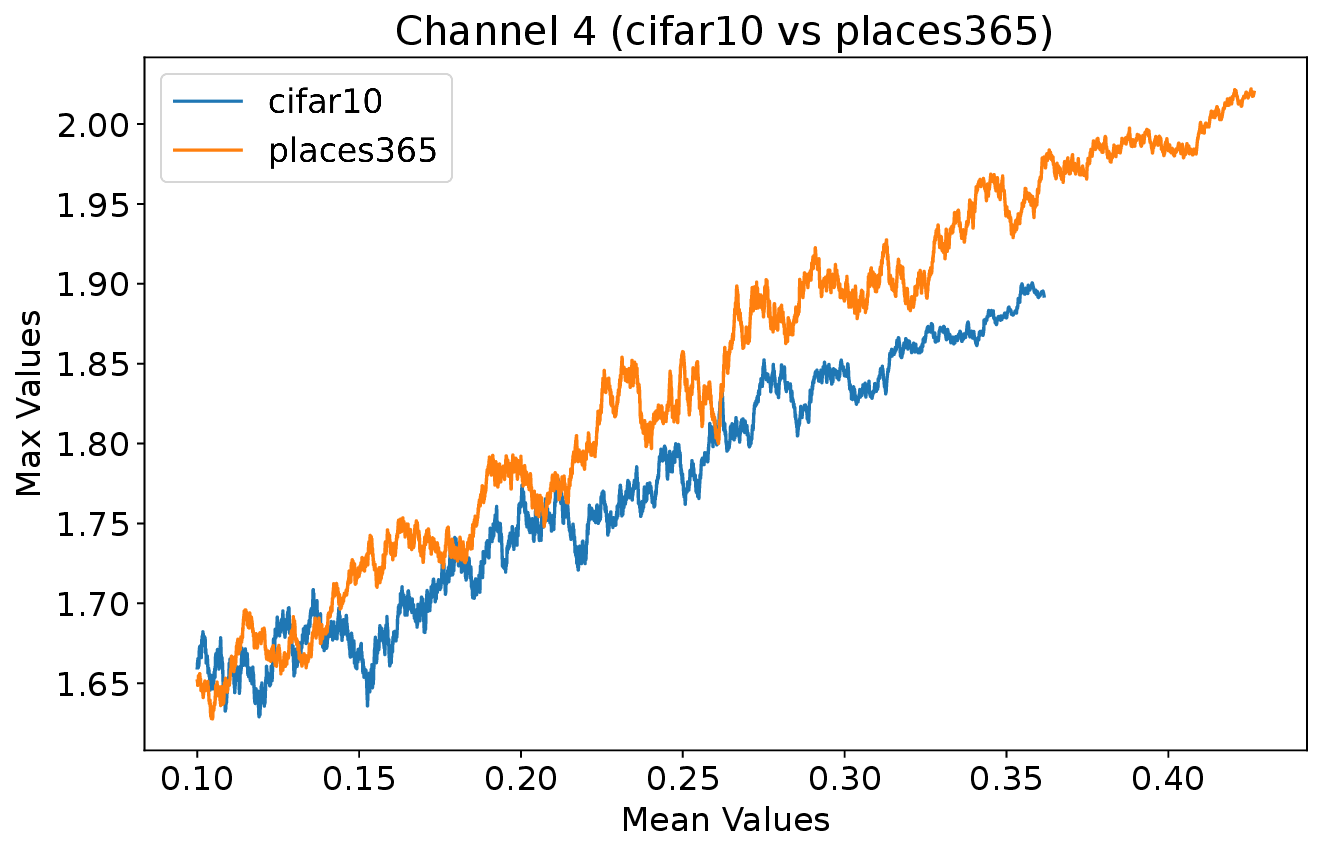}
    \caption{In Transaction Block 1}
    \label{fig:places3651}
  \end{subfigure}

  \begin{subfigure}{\linewidth}
    \includegraphics[width=0.24\linewidth]{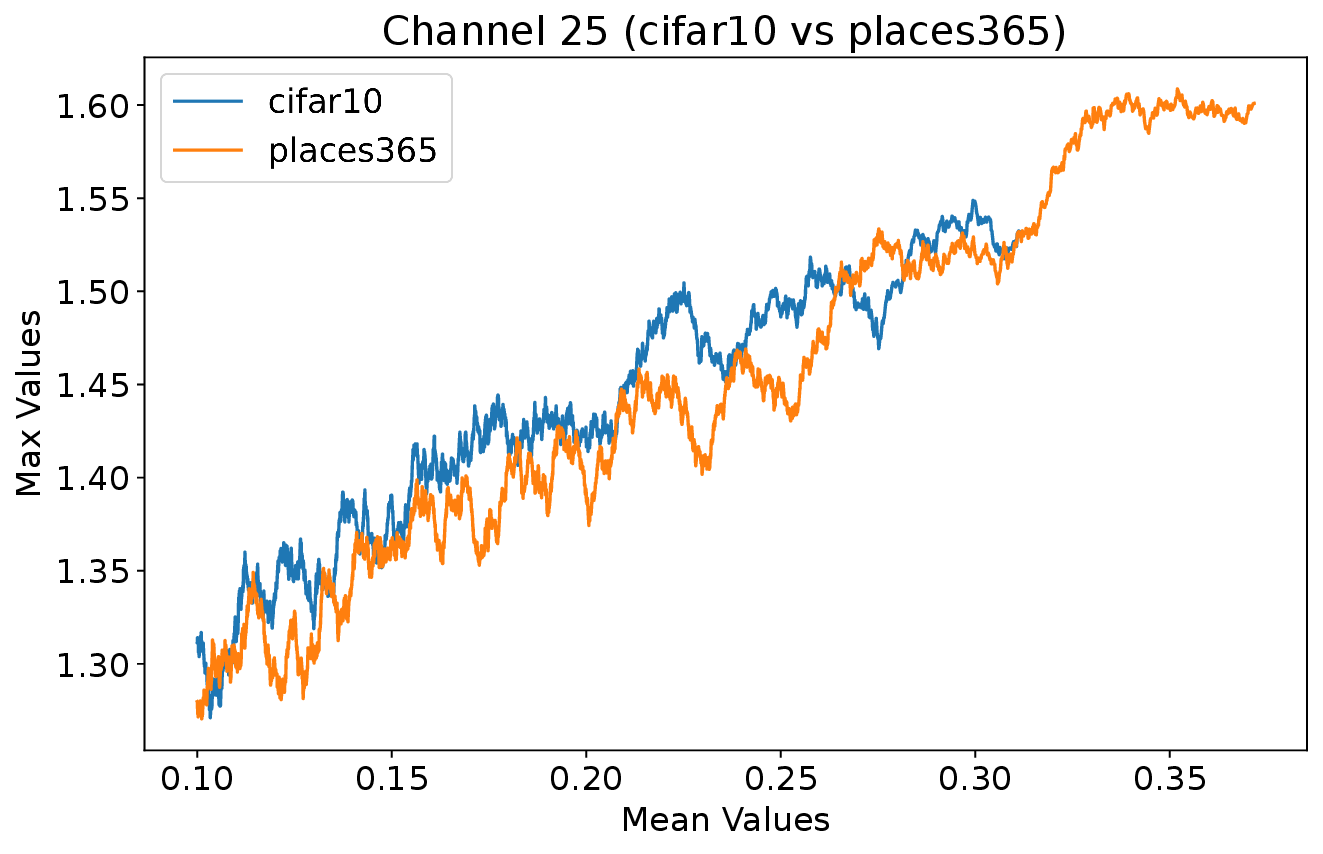}
    \includegraphics[width=0.24\linewidth]{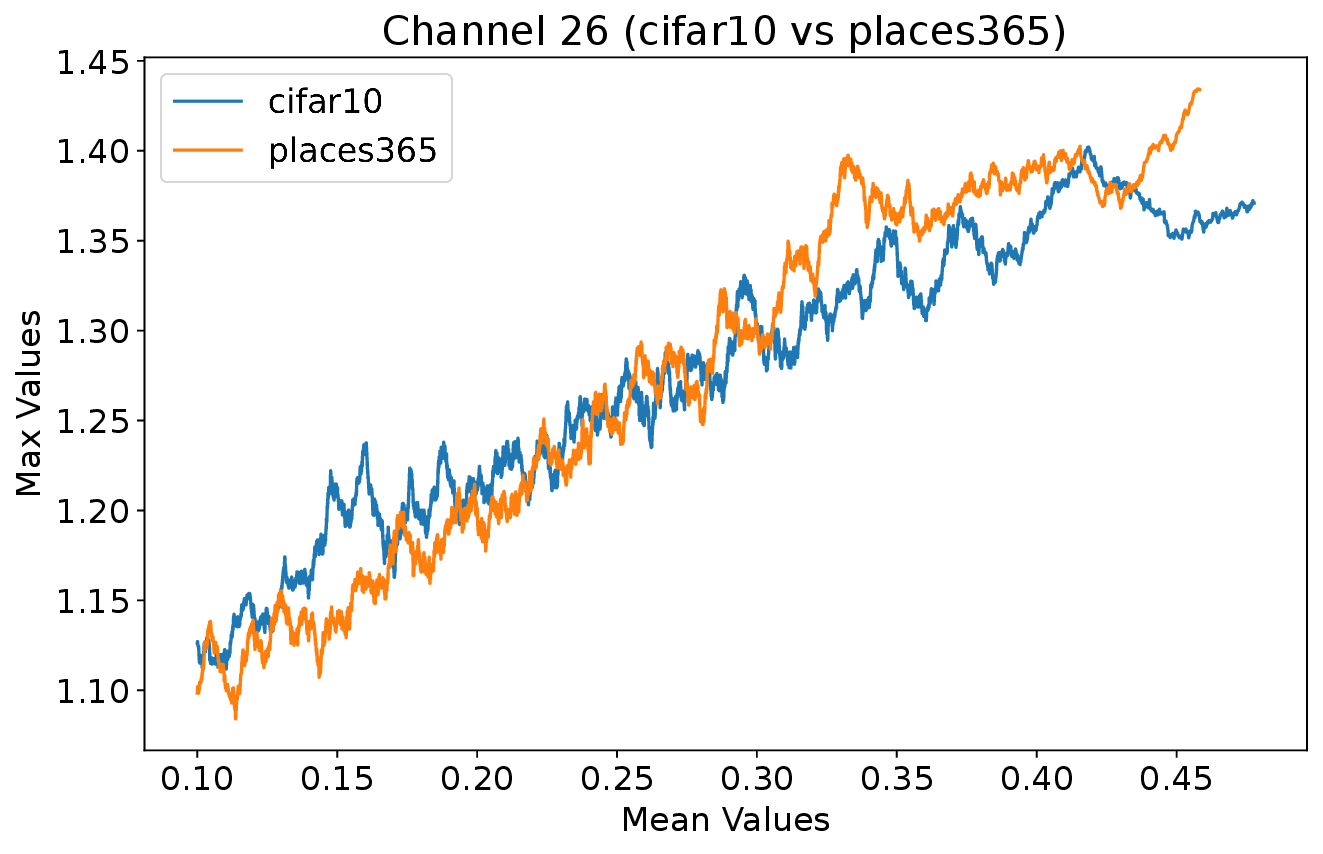}
    \includegraphics[width=0.24\linewidth]{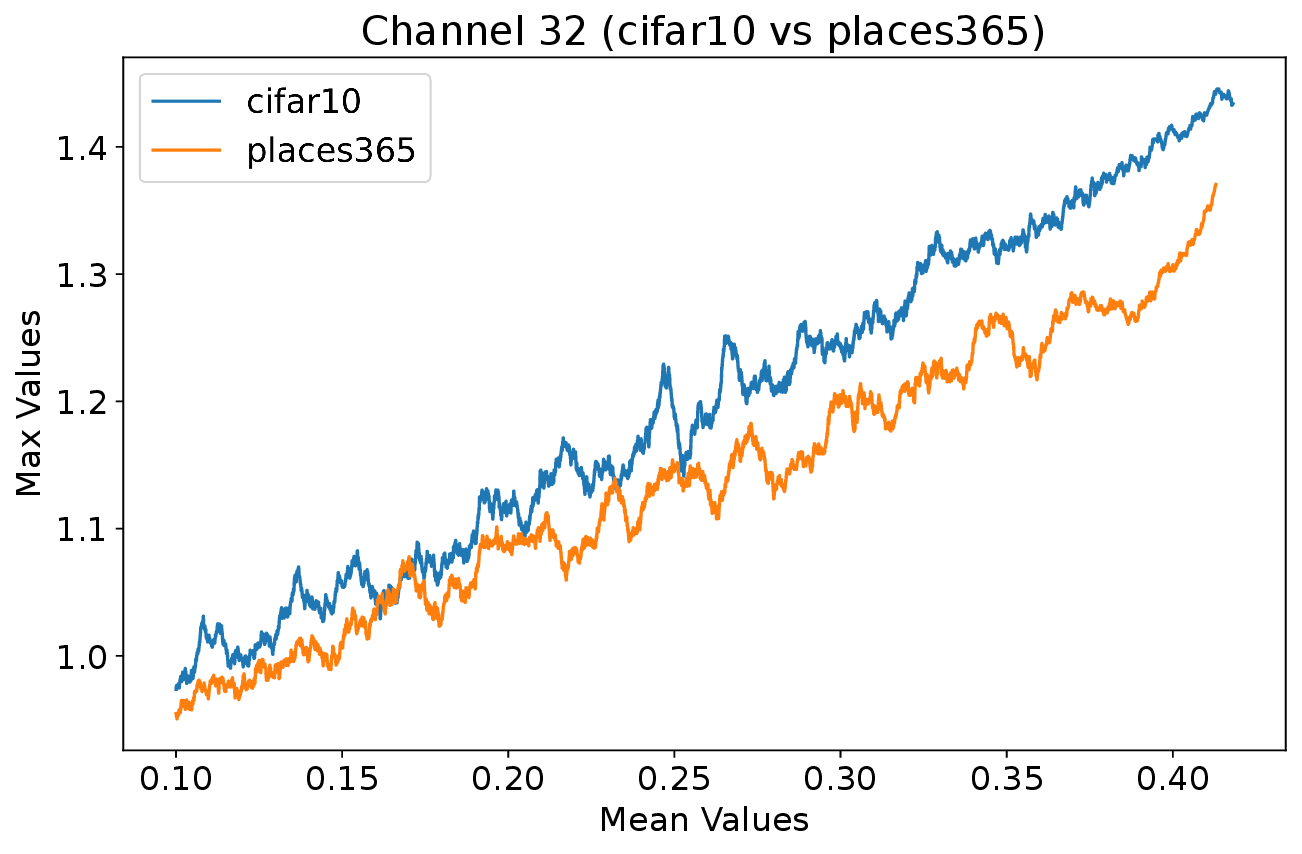}
    \includegraphics[width=0.24\linewidth]{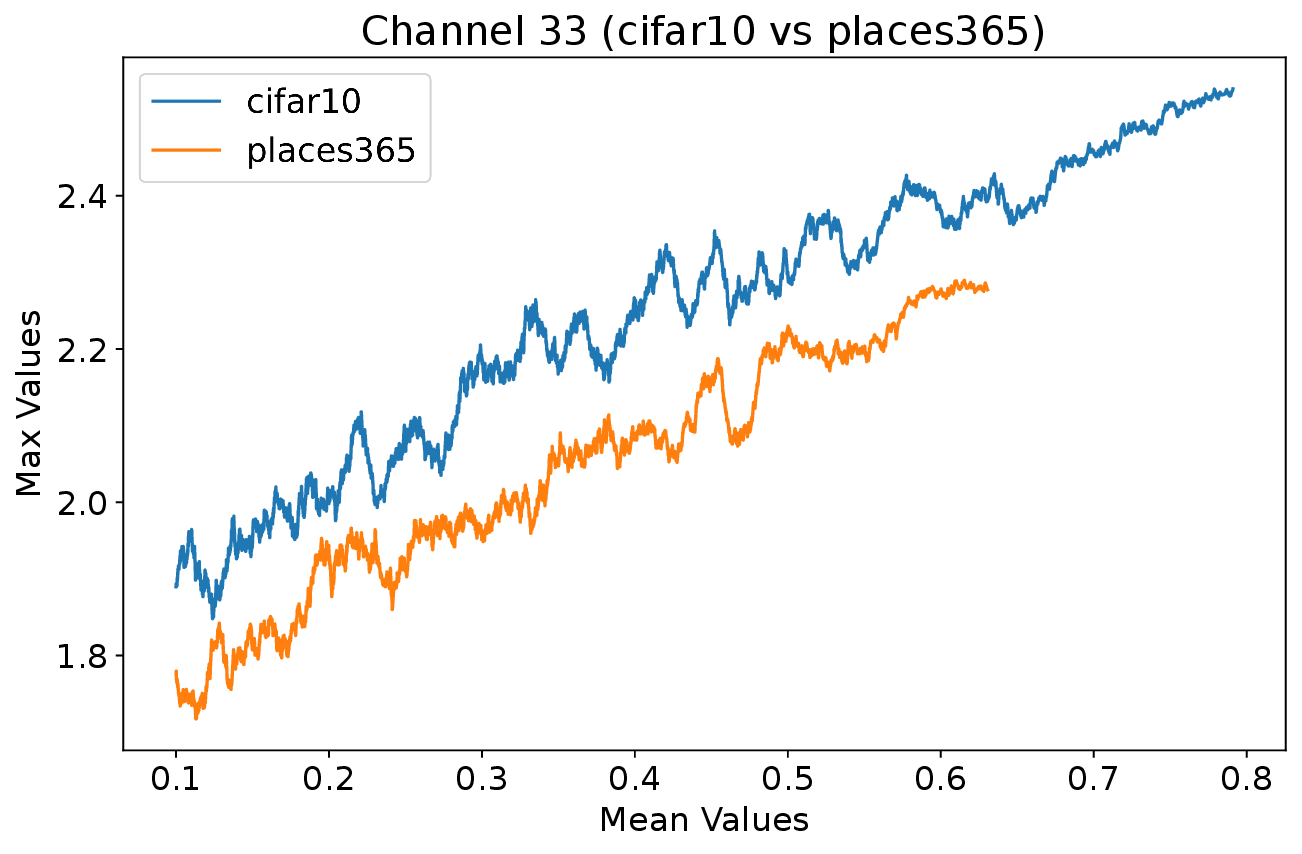}
    \caption{In Transaction Block 2}
    \label{fig:places3652}
  \end{subfigure}

  \begin{subfigure}{\linewidth}
    \includegraphics[width=0.24\linewidth]{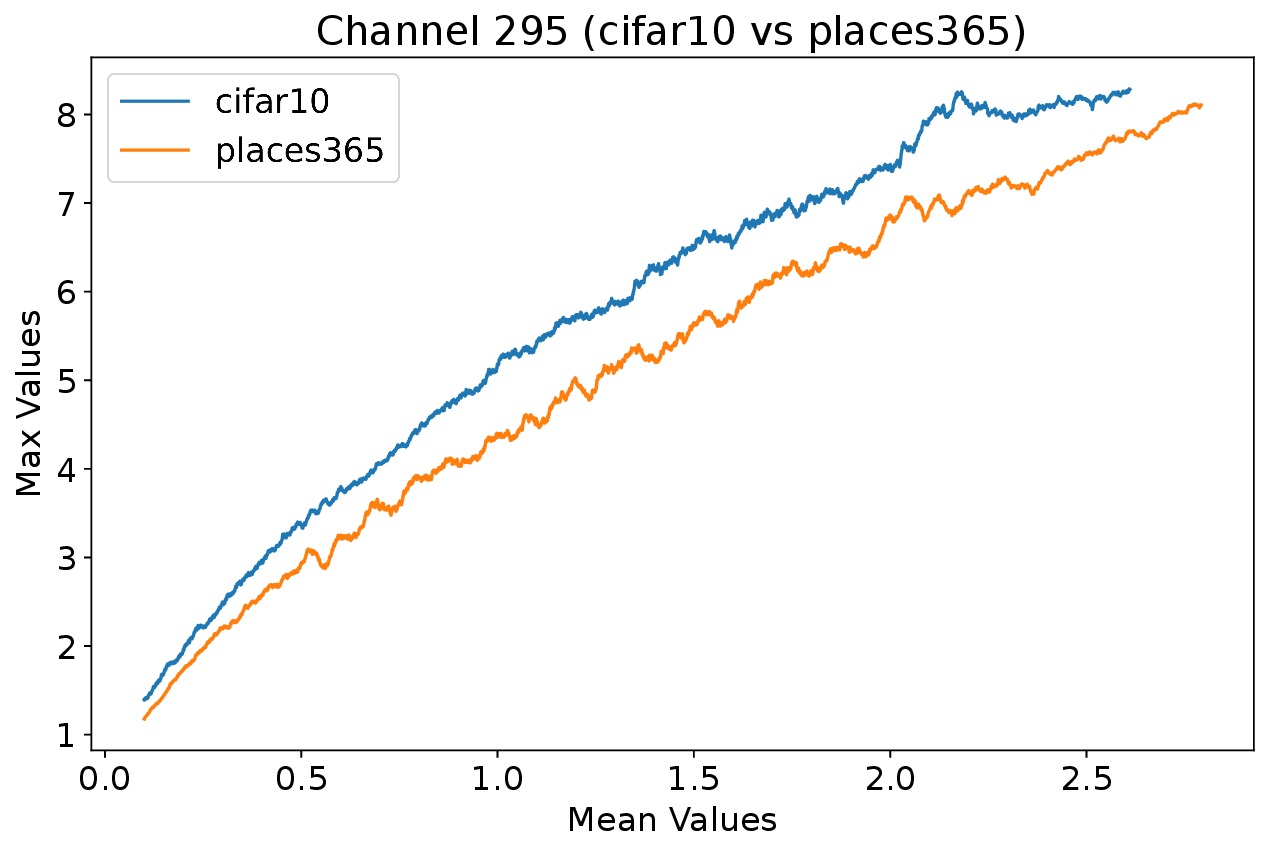}
    \includegraphics[width=0.24\linewidth]{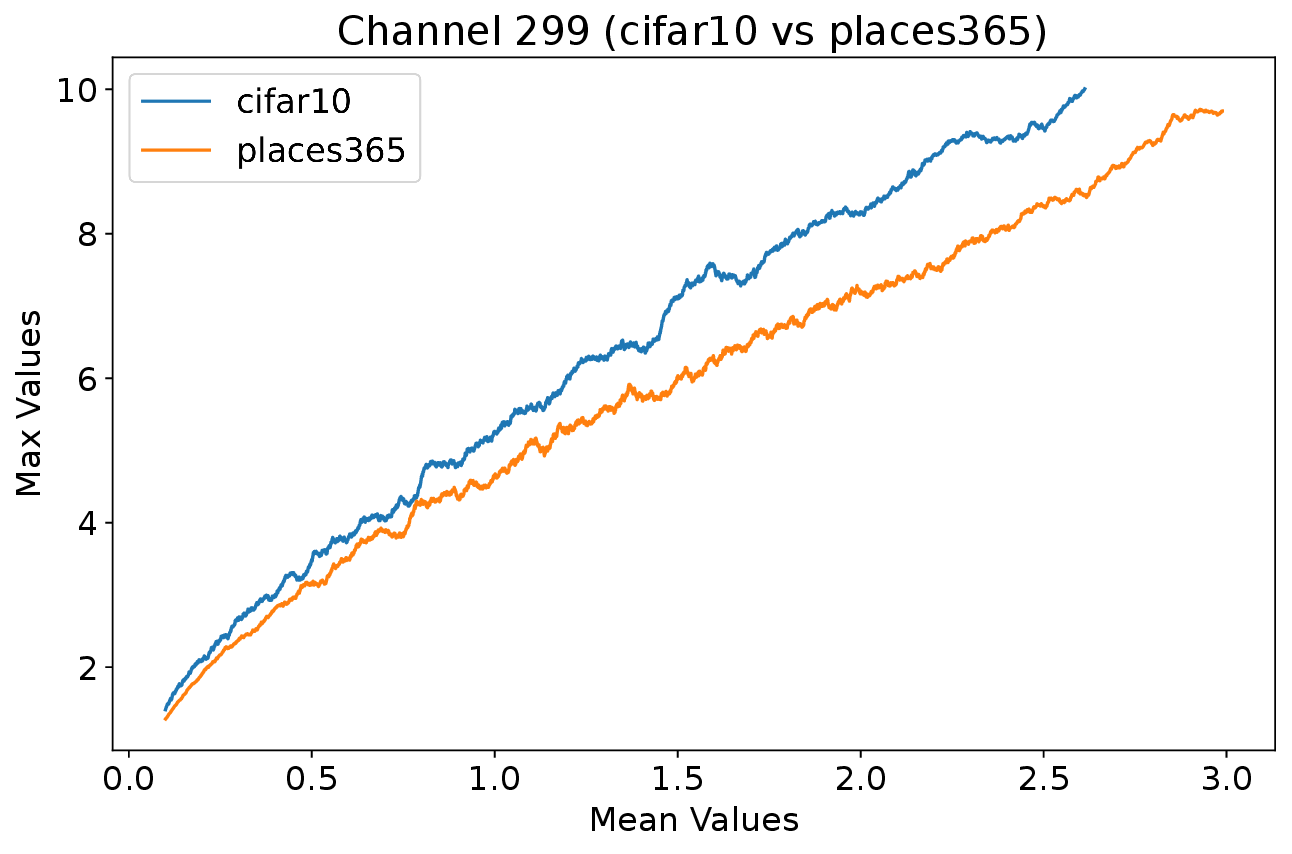}
    \includegraphics[width=0.24\linewidth]{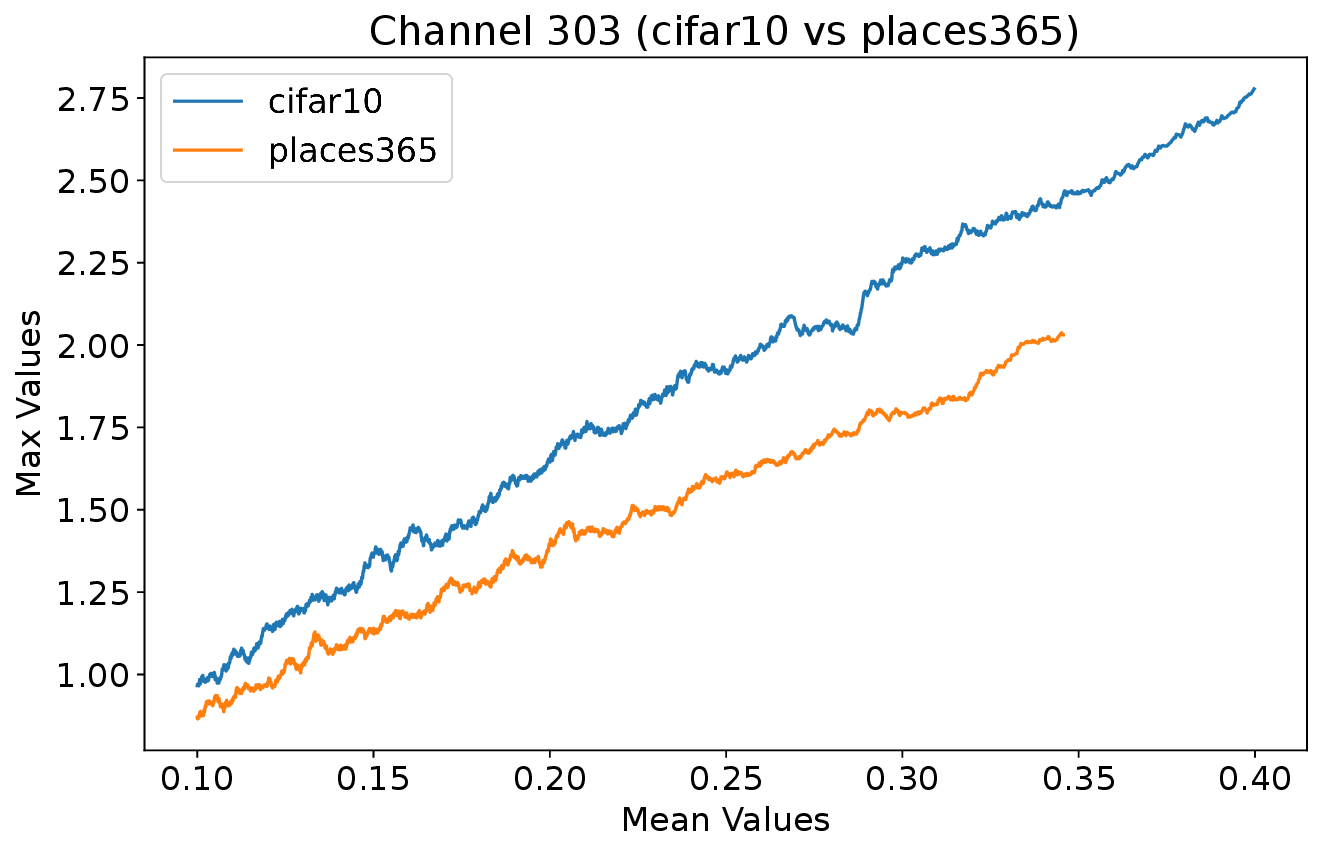}
    \includegraphics[width=0.24\linewidth]{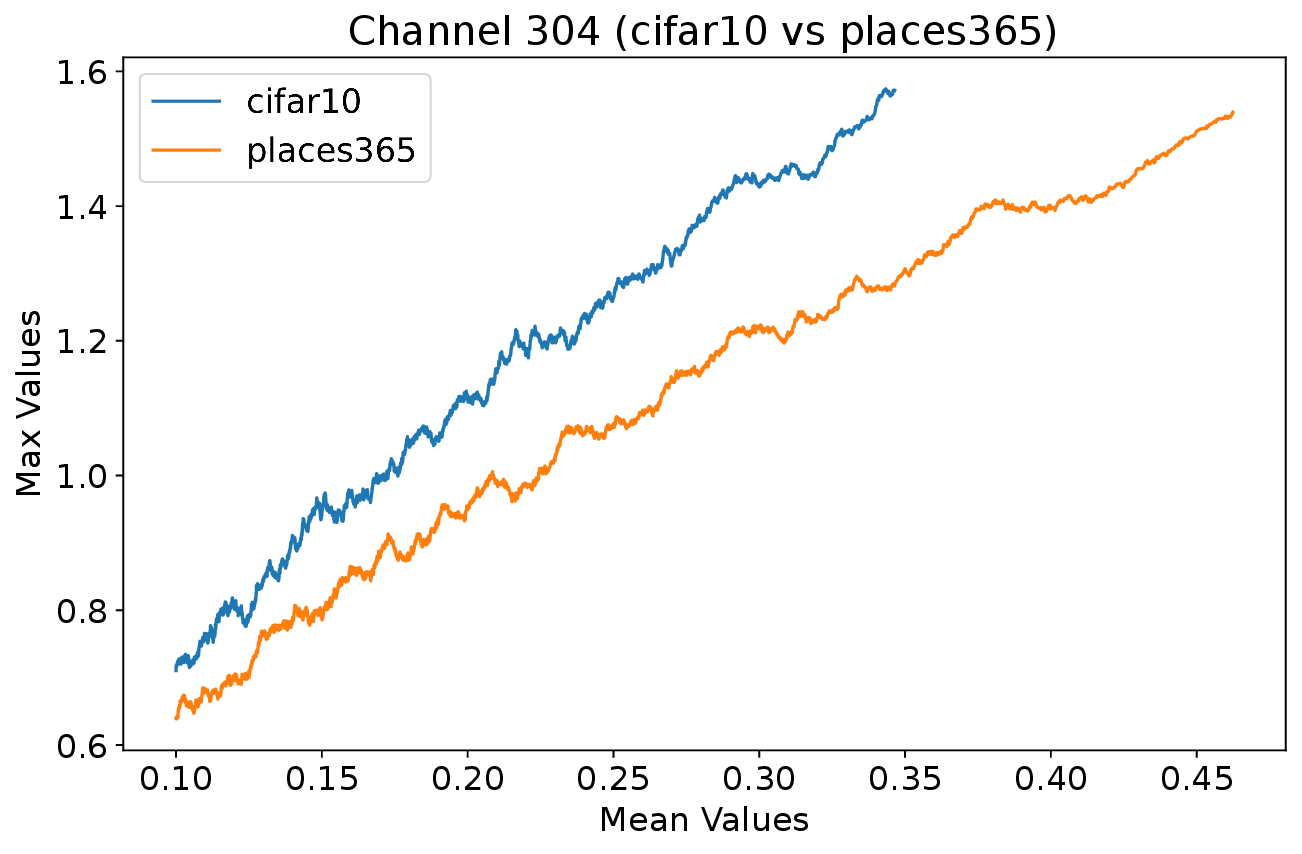}
    \caption{Before Global Pooling}
    \label{fig:places3653}
  \end{subfigure}

  \caption{\textbf{Activation distribution at different positions within the DenseNet architecture~\cite{huang2017densely} applied to CIFAR-10 and Places365 datasets.} For this analysis, four specific locations within the network were chosen: (a) after the first convolution layer, (b) just before the pooling operation in the first transition block, (c) just before the pooling operation in the second transition block, and (d) right before the final global pooling layer. Note that we only include data points with an average activation over $0.1$. As shown in the figure, the first three selected layers show a less marked distinction between ID and OOD samples, while the fourth layer, preceding the final global pooling layer, demonstrates clearer separability and enhanced stability. Therefore, the fourth selected layer (the penultimate layer) is more suitable for developing a scoring function for OOD detection.}
  \label{fig:appendix-places365}
\end{figure}

\section{Why is the penultimate layer more effective for NAP?}
\label{appendix:layer}
We provide an extensive collection of visualizations showcasing the activations within the DenseNet architecture when applied to the CIFAR-10 (ID) dataset and Places365 (OOD) dataset. These visualizations are crucial for understanding how the network processes both ID and OOD data, revealing the distinct patterns of neural activations at various layers of the network.
Our analysis focuses on four critical layers within DenseNet: (1) after the first convolutional layer, (2) before the pooling operation in the first transition block, (3) before pooling in the second transition block, and (4) before the final global pooling layer. Each layer offers four visualizations, providing a comprehensive view of the network's response to different datasets.

These detailed visualizations enhance the discussion in the main text, offering deeper insights into how NAP effectively distinguishes between ID and OOD samples within the network. As depicted in Figure~\ref{fig:appendix-places365}, the first three selected layers, which focus primarily on low-level features, exhibit a less pronounced distinction between ID and OOD samples. This is likely because low-level features, such as edges and textures, are common to both ID and OOD datasets, making them less distinctive. However, the contrast between ID and OOD samples becomes more evident and stable in the fourth selected layer, located before the final global pooling layer. This layer, concentrating on high-level semantic information, captures features more unique to the ID dataset, leading to clearer separability and enhanced stability in activation values compared to earlier layers. This layer's focus on distinctive semantic features makes it particularly suitable for developing a scoring function for OOD detection. 
Therefore, since the penultimate layer is the most informative layer in the neural network, we utilize this layer in our method to develop our scoring function.
%%%%%%%%%%%%%%%%%%%%%%%%%%%%%%%%%%%%%%%%%%%%%%%%%%%%%%%%%%%%%%%%%%%%%%%%%%%
\begin{table}[!t]
\footnotesize
\centering
\caption{\textbf{Experimental results of OOD detection using NAP at different layers in DenseNet~\cite{huang2017densely} on CIFAR-10 and CIFAR-100 datasets.} NAP(x) indicates the computation of OOD score at the layer 'x', where 'c1' corresponds to after the first convolutional layer, 't1' before the pooling operation in the first transition block, 't2' before pooling in the second transition block, and 'p' before the final global pooling layer. Combinations of layers, indicated by commas in NAP(..), represent the multiplication of OOD scores from respective layers. These selected layers are consistent with those used for visualizations described earlier in the paper. Notably, NAP(p) is the approach actually utilized in our paper.\\}
\label{tab:multi-layer}
\begin{tabular}{@{}lcc|cc@{}}
\toprule
\multirow{2}{*}{\textbf{Method}} & \multicolumn{2}{c|}{\textbf{CIFAR-10}} & \multicolumn{2}{c}{\textbf{CIFAR-100}} \\ \cmidrule{2-5} 
                                 & FPR95              & AUROC             & FPR95              & AUROC             \\ \midrule
NAP(c1)                          & 83.22              & 51.99             & 84.13              & 50.34             \\
NAP(t1)                          & 69.10              & 50.47             & 86.82              & 54.92             \\
NAP(t2)                          & 56.53              & 78.44             & 88.85              & 53.08             \\
NAP(c1,t1,t2,p)                  & 68.33              & 58.99             & 82.66              & 56.96             \\
NAP(t1,t2,p)                     & 56.84              & 64.26             & 83.35              & 57.97             \\
NAP(t2,p)                        & 34.41              & 87.81             & 82.74              & 58.31             \\
\textbf{NAP(p)}                  & \textbf{26.57}     & \textbf{92.45}    & \textbf{54.91}     & \textbf{85.86}    \\ \bottomrule
\end{tabular}
\end{table}

\section{Evaluating multi-layer integration with NAP for OOD detection}
\label{appendix:integration}
In an additional exploration presented in the appendix, we investigate the effects of incorporating values from both the penultimate layer and earlier layers on OOD detection. Our experiments, detailed in Table~\ref{tab:multi-layer}, suggest that the integration of earlier layers with the penultimate layer, where NAP is primarily applied, does not yield significant improvements in OOD detection. This phenomenon could partly stem from the inherent limitations of earlier layers in differentiating between ID and OOD data. Additionally, there is a possibility that the scoring function, specifically optimized for the penultimate layer, may not align optimally with the feature representation characteristics of the preceding layers. Considering the constraints of space in this paper, a comprehensive analysis of multi-layer integration using NAP is not presented. Nonetheless, the potential of combining multiple layers in OOD detection, especially in the context of NAP, remains an intriguing aspect for future research. We anticipate that further investigations, potentially involving the creation of new scoring functions suitable for a broader range of layers, could provide substantial contributions to the field. Thus, we propose this as an avenue for future work, aiming to stimulate further advancements within the research community.
%%%%%%%%%%%%%%%%%%%%%%%%%%%%%%%%%%%%%%%%%%%%%%%%%%%%%%%%%%%%%%%%%%%%%%%%%%%
\section{On transferability to other architectures}
\label{appendix:transfer}
In Figure~\ref{fig:appendix-e-mobilenet}, \ref{fig:appendix-e-resnet}, and \ref{fig:appendix-e-vgg}, we present detailed visualizations of the activation patterns within three distinct architectures: MobileNetV2~\cite{sandler2018mobilenetv2}, ResNet50~\cite{he2016deep}, and VGG16~\cite{simonyan2015very}. These visualizations clearly demonstrate a remarkable gap between ID samples, depicted with blue lines, and OOD samples, represented with orange lines. This distinction is evident across all three architectures, underscoring the versatility and effectiveness of the proposed NAP. The consistent separability observed in these diverse architectures confirms the adaptability and potential of NAP for broad application in different neural network models.

In order to validate the effectiveness of NAP across different Convolutional Neural Network (CNN) architectures, we conducted experiments on a variety of CNN backbones. As depicted in Table~\ref{tab:backbones}, our proposed NAP method significantly enhances the OOD detection performance across various CNN structures.
These results underscore the adaptability of NAP to various CNN models, demonstrating its potential as a versatile tool for enhancing the reliability and accuracy of OOD detection in neural network applications. 
\begin{table}[!h]
\scriptsize
\centering
\caption{Results on ImageNet with various backbones.\\\quad\\}
\label{tab:backbones}
\begin{tabular}{@{}lcc|cc|cc@{}}
\toprule
             & \multicolumn{2}{c|}{Energy} & \multicolumn{2}{c|}{NAP} & \multicolumn{2}{c}{NAP-E} \\ \cmidrule(l){2-7} 
             & FPR95        & AUROC        & FPR95       & AUROC      & FPR95       & AUROC       \\ \midrule
VGG          & 54.34             & 88.17             &  29.23           & 93.46           & \textbf{23.23}            & \textbf{95.00}            \\
DenseNet      & 50.40             & 87.66             & 49.89            & 88.40           & \textbf{32.95}            & \textbf{91.68}            \\
ResNet    & 57.47             & 87.05             & 48.77            & 82.76           & \textbf{32.12}            & \textbf{92.02}            \\ \bottomrule
\end{tabular}
% \vspace{-12pt}
\end{table}
%%%%%%%%%%%%%%%%%%%%%%%%%%%%%%%%%%%%%%%%%%%%%%%%%%%%%%%%%%%%%%%%%%%%%%%%%%%
\section{Pareto frontier of ID accuracy and OOD detection performance}
\label{appendix:pareto}
\begin{figure}[!h]
% \vspace{-15pt}
  \centering
  \begin{subfigure}{0.47\linewidth}
    \includegraphics[width=0.99\linewidth]{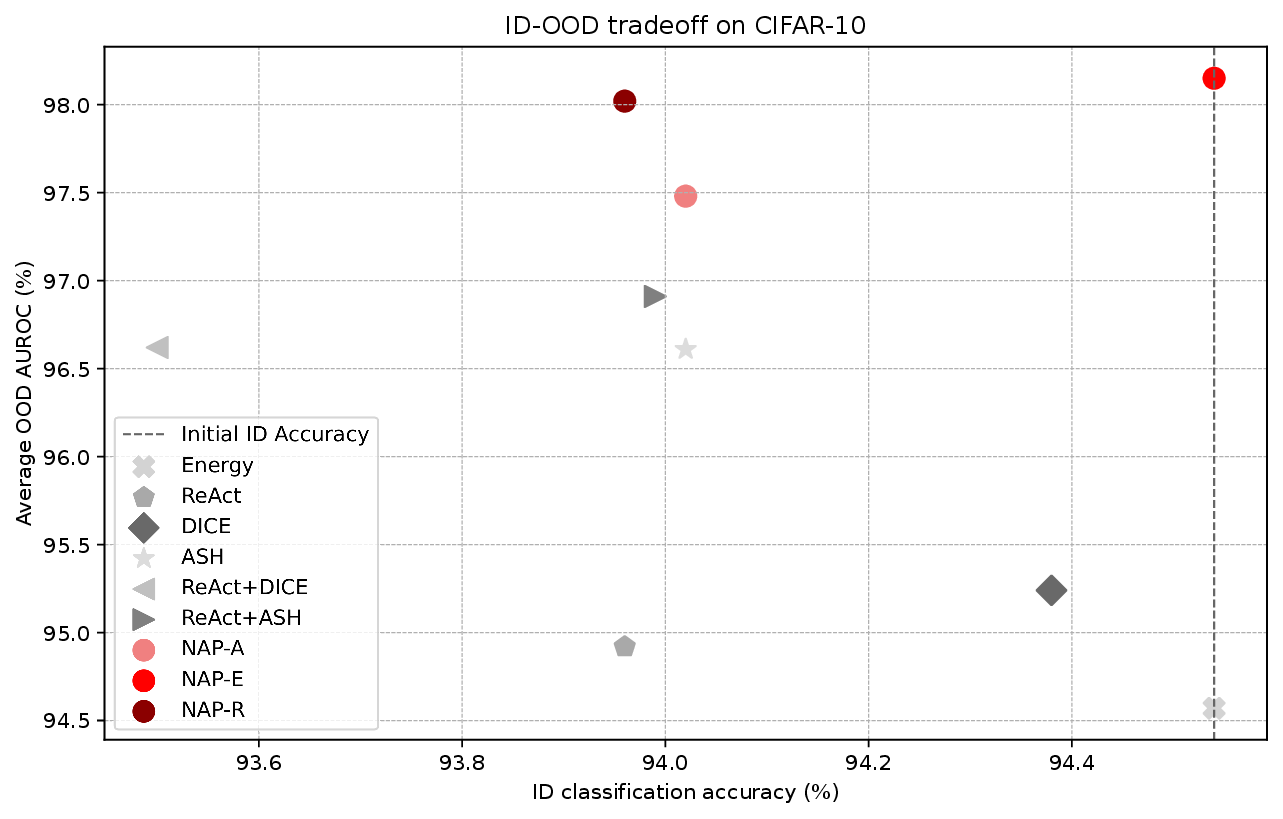}
    \caption{CIFAR-10}
    \label{fig:p1}
  \end{subfigure}
  \hfill
  \begin{subfigure}{0.46\linewidth}
    \includegraphics[width=0.99\linewidth]{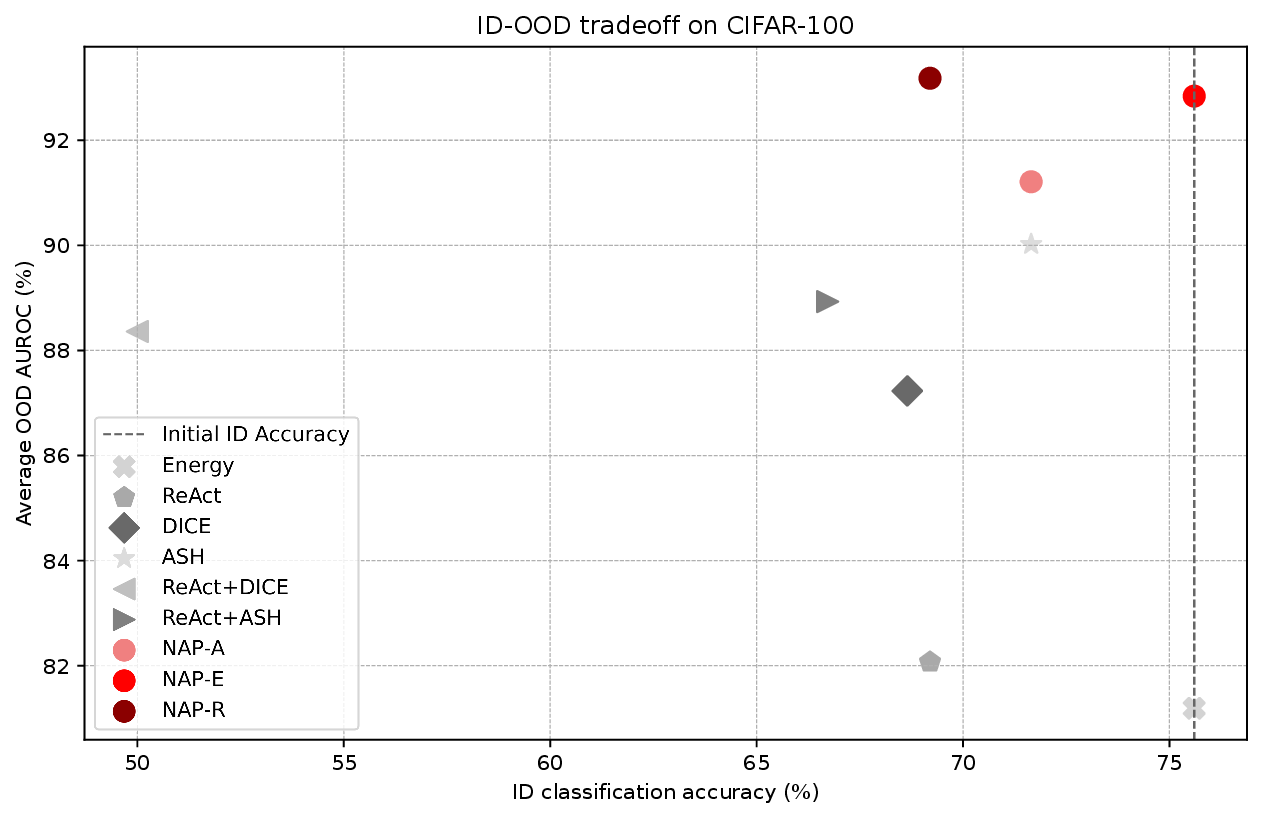}
    \caption{CIFAR-100}
    \label{fig:p2}
  \end{subfigure}
  \caption{\textbf{Investigating the trade-offs between ID classification accuracy and OOD detection AUROC on CIFAR benchmarks~\cite{krizhevsky2009cifar} across various methods.} All methods and experiments are implemented by us. Methods prefixed with 'NAP' are visually distinguished, highlighted in various shades of red in the figures.}
  \label{fig:pareto-cifar}
\end{figure}
Some existing methods negatively affect ID accuracy; however, we have found that integrating these methods with the NAP approach can mitigate or reduce this impact, achieving a more optimal balance. The NAP approach establishes an ideal equilibrium between ID classification accuracy and OOD detection efficacy (measured in AUROC), thereby positioning it favorably on the Pareto front for superior performance. This demonstrates our method's ability to enhance OOD detection without additional costs while maintaining the model’s classification performance, as illustrated in Figures~\ref{fig:p1} and ~\ref{fig:p2} using the CIFAR benchmark~\cite{krizhevsky2009cifar}.
%%%%%%%%%%%%%%%%%%%%%%%%%%%%%%%%%%%%%%%%%%%%%%%%%%%%%%%%%%%%%%%%%%%%%%%%%%%
\section{Full CIFAR benchmark results: enhancing methods with NAP}
\label{appendix:cifar}
This section focuses on the significant enhancements brought by the Neural Activation Prior (NAP) to existing out-of-distribution detection methods in the context of the CIFAR-10 (in Table~\ref{tab:cifar10}) and CIFAR-100 (in Table~\ref{tab:cifar100}) dataset. The incorporation of NAP into established methods like Energy~\cite{liu2020energy}, ASH~\cite{djurisic2022extremely}, DICE~\cite{sun2022dice}, KNN~\cite{sun2022knn}, MSP~\cite{hendrycks2016msp} and ReAct~\cite{sun2021react}, resulting in \textbf{NAP-E}, \textbf{NAP-A}, \textbf{NAP-D}, \textbf{NAP-K}, \textbf{NAP-M}and \textbf{NAP-R} respectively, showcases the potential of NAP in augmenting existing approaches.
Our experimental results demonstrate that NAP-based variants consistently outperform their corresponding traditional methods across all six OOD datasets. Notably, our experimental results reveal some extremely substantial decreases in FPR95 values, indicative of the profound impact of NAP integration. For instance, on the CIFAR-100 dataset, the FPR95 value of NAP-R, compared to ReAct, dropped by 83.06\% (from 83.81 to 14.19), highlighting a notable reduction in false alarms while maintaining high detection accuracy and affirming the enhanced capability of these methods in distinguishing between in-distribution and OOD samples.

\begin{table}[!h]
\tiny
\centering
\caption{\textbf{Scoring function proposed based on NAP is compatible with and improves on existing methods on CIFAR-10 dataset.}  All methods and experiments are implemented by us. All values in this table are percentages. The average over six OOD test datasets is also reported. The methods with prefix 'NAP-' (e.g., NAP-E, NAP-A, NAP-R) represent the integration of NAP with various existing methods (Energy~\cite{liu2020energy}, ASH-S~\cite{djurisic2022extremely}, ReAct~\cite{sun2021react} respectively).\\}
\label{tab:cifar10}
\resizebox{\textwidth}{!}{
\begin{tabular}{LcccccccccccccR}
\toprule
\multicolumn{1}{c}{\multirow{3}{*}{\textbf{Method}}} & \multicolumn{12}{c}{\textbf{OOD Datasets}}                                                                                                                                                                                                                                                                                                    & \multicolumn{2}{c}{\multirow{2}{*}{\textbf{Average}}} \\ \cline{2-13}
\multicolumn{1}{c}{}                                 & \multicolumn{2}{c}{\textbf{SVHN}}                     & \multicolumn{2}{c}{\textbf{Textures}}                 & \multicolumn{2}{c}{\textbf{iSUN}}                     & \multicolumn{2}{c}{\textbf{LSUN}}                     & \multicolumn{2}{c}{\textbf{LSUN-Crop}}                & \multicolumn{2}{c}{\textbf{Places365}}                & \multicolumn{2}{c}{}                                  \\
\multicolumn{1}{c}{}                                 & \multicolumn{1}{c}{FPR95} & \multicolumn{1}{c}{AUROC} & \multicolumn{1}{c}{FPR95} & \multicolumn{1}{c}{AUROC} & \multicolumn{1}{c}{FPR95} & \multicolumn{1}{c}{AUROC} & \multicolumn{1}{c}{FPR95} & \multicolumn{1}{c}{AUROC} & \multicolumn{1}{c}{FPR95} & \multicolumn{1}{c}{AUROC} & \multicolumn{1}{c}{FPR95} & \multicolumn{1}{c}{AUROC} & \multicolumn{1}{c}{FPR95} & \multicolumn{1}{c}{AUROC} \\ \hline
ASH                            & 6.51                      & 98.65                     & 24.34                     & 95.09                     & 5.17                      & 98.90                     & 4.96                      & 98.92                     & 0.90                      & 99.73                     & 48.45                     & 88.34                     & 15.05                     & 96.91                     \\ 
\rowcolor[HTML]{C0C0C0}
\textbf{NAP-A}            & 5.55                      & 98.86                     & 10.51                     & 97.90                     & 3.04                      & 99.32                     & 2.68                      & 99.40                     & 0.80                      & 99.80                     & 44.28                     & 89.59                     & 11.14                     & 97.48                    \\ % \midrule
DICE                      & 29.62                     &      94.66                     & 0.38                      &      99.90                     & 4.43                      &      99.03                     & 5.14                      &      98.97                     & 45.87                     &      86.97                     & 45.32                     &      90.29                     & 21.79                     &      94.97                 \\
\rowcolor[HTML]{C0C0C0}
\textbf{NAP-D}            & 10.60                      &     97.75                     & 0.41                       &     99.88                     & 2.03                       &     99.48                     & 2.69                       &     99.41                     & 13.85                      &     96.98                     & 40.40                      &     91.31                     & 11.66                      &     97.47                  \\ % \midrule
Energy                           & 40.57                     & 93.99                     & 56.29                     & 86.42                     & 10.07                     & 98.07                     & 9.28                      & 98.12                     & 3.81                      & 99.15                     & 39.50                     & 92.01                     & 26.59                     & 94.63                     \\
\rowcolor[HTML]{C0C0C0}
\textbf{NAP-E}            & 8.32                      & 98.36                     & 11.65                     & 97.72                     & 1.77                      & 99.57                     & 1.50                      & 99.60                     & 0.99                      & 99.76                     & 29.89                     & 93.91                     & 9.02                      & 98.15                     \\ % \midrule
KNN                       & 4.31                      &      99.20                     & 7.71                      &      98.62                     & 9.45                      &      98.22                     & 10.08                     &      98.15                     & 19.31                     &      96.46                     & 45.83                     &      90.09                     & 16.12                     &      96.79                 \\
\rowcolor[HTML]{C0C0C0}
\textbf{NAP-K}            & 2.39                      &      99.56                     & 2.29                      &      99.55                     & 1.76                      &      99.57                     & 2.45                      &      99.47                     & 3.58                      &      99.34                     & 34.27                     &      92.80                     & 7.79                      &      98.38                 \\ % \midrule
MSP                       & 47.34                     &      93.48                     & 33.66                     &      95.54                     & 42.21                     &      94.51                     & 42.42                     &      94.52                     & 64.52                     &      88.14                     & 61.98                     &      88.95                     & 48.69                     &      92.52                 \\
\rowcolor[HTML]{C0C0C0}
\textbf{NAP-M}            & 14.09                     &      96.05                     & 7.33                      &      98.35                     & 10.91                     &      97.72                     & 11.20                     &      97.55                     & 16.42                     &      96.23                     & 54.61                     &      84.76                     & 19.09                     &      95.11                 \\ % \midrule
ReAct                            & 41.64                     & 93.87                     & 43.58                     & 92.47                     & 12.72                     & 97.72                     & 11.46                     & 97.87                     & 5.96                      & 98.84                     & 43.31                     & 91.03                     & 26.45                     & 94.67                     \\
\rowcolor[HTML]{C0C0C0}
\textbf{NAP-R}            & 8.07                      & 98.31                     & 8.10                      & 98.17                     & 2.81                      & 99.35                     & 2.35                      & 99.43                     & 3.04                      & 99.33                     & 30.70                     & 93.50                     & 9.18                      & 98.02                    \\ \bottomrule
\end{tabular}
}
\end{table}

\begin{table}[!h]
\tiny
\centering
\caption{\textbf{Scoring function proposed based on NAP is compatible with and improves on existing methods on CIFAR-100 dataset.}  All methods and experiments are implemented by us. All values in this table are percentages. The average over six OOD test datasets is also reported. The methods with prefix 'NAP-' (e.g., NAP-E, NAP-A, NAP-R) represent the integration of NAP with various existing methods (Energy~\cite{liu2020energy}, ASH-S~\cite{djurisic2022extremely}, ReAct~\cite{sun2021react} respectively).\\}
\label{tab:cifar100}
\resizebox{\textwidth}{!}{
\begin{tabular}{LcccccccccccccR}
\toprule
\multicolumn{1}{c}{\multirow{3}{*}{\textbf{Method}}} & \multicolumn{12}{c}{\textbf{OOD Datasets}}                                                                                                                                                                                                                                                                                                    & \multicolumn{2}{c}{\multirow{2}{*}{\textbf{Average}}} \\ \cline{2-13}
\multicolumn{1}{c}{}                                 & \multicolumn{2}{c}{\textbf{SVHN}}                     & \multicolumn{2}{c}{\textbf{Textures}}                 & \multicolumn{2}{c}{\textbf{iSUN}}                     & \multicolumn{2}{c}{\textbf{LSUN}}                     & \multicolumn{2}{c}{\textbf{LSUN-Crop}}                & \multicolumn{2}{c}{\textbf{Places365}}                & \multicolumn{2}{c}{}                                  \\
\multicolumn{1}{c}{}                                 & \multicolumn{1}{c}{FPR95} & \multicolumn{1}{c}{AUROC} & \multicolumn{1}{c}{FPR95} & \multicolumn{1}{c}{AUROC} & \multicolumn{1}{c}{FPR95} & \multicolumn{1}{c}{AUROC} & \multicolumn{1}{c}{FPR95} & \multicolumn{1}{c}{AUROC} & \multicolumn{1}{c}{FPR95} & \multicolumn{1}{c}{AUROC} & \multicolumn{1}{c}{FPR95} & \multicolumn{1}{c}{AUROC} & \multicolumn{1}{c}{FPR95} & \multicolumn{1}{c}{AUROC} \\ \hline
ASH                                                 & 25.02                     & 95.76                     & 34.02                     & 92.35                     & 46.67                     & 91.30                     & 51.33                     & 90.12                     & 5.52                      & 98.84                     & 85.86                     & 71.62                     & 41.40                     & 90.02                     \\
\rowcolor[HTML]{C0C0C0}
\textbf{NAP-A}                                 & 17.41                     & 96.72                     & 22.70                     & 94.99                     & 38.22                     & 93.34                     & 43.05                     & 92.17                     & 5.25                      & 98.94                     & 85.76                     & 72.08                     & 35.40                     & 91.32                     \\
DICE                           & 59.25                      & 88.57                      & 0.91                      & 99.74                        & 51.63                      & 89.32                      & 49.48                      & 89.51                       & 61.42                      & 77.12                      & 80.29                      & 77.08                      & 50.50                      & 86.89                       \\
\rowcolor[HTML]{C0C0C0}
\textbf{NAP-D}            & 23.63                       & 95.28                      & 1.22                      & 99.68                      & 33.86                       & 94.25                      & 28.56                       & 95.02                      & 24.59                       & 92.04                      & 82.12                      & 77.13                      & 32.34                       & 92.23                      \\ % \midrule
Energy                                                & 87.46                     & 81.85                     & 84.15                     & 71.03                     & 74.54                     & 78.95                     & 70.65                     & 80.14                     & 14.72                     & 97.43                     & 79.20                     & 77.72                     & 68.45                     & 81.19                     \\
\rowcolor[HTML]{C0C0C0}
\textbf{NAP-E}                                 & 19.03                     & 96.40                     & 21.72                     & 95.47                     & 33.24                     & 94.15                     & 43.38                     & 92.11                     & 2.60                      & 99.38                     & 75.70                     & 79.54                     & 32.61                     & 92.84                     \\
KNN                           & 16.27                      & 96.65                      & 28.06                      & 92.69                      & 58.74                      & 82.09                      & 52.77                       & 84.55                      & 26.01                      & 93.53                      & 87.59                      & 69.94                      & 44.91                      & 86.58                      \\
\rowcolor[HTML]{C0C0C0}
\textbf{NAP-K}            & 10.26                       & 97.84                      & 12.24                      & 97.76                      & 45.56                       & 91.57                      & 36.45                       & 93.22                      & 9.84                       & 98.04                      & 87.42                      & 70.83                      & 33.63                       & 91.54                      \\ % \midrule
MSP                         & 81.70                      & 75.40                      & 60.49                      & 85.60                      & 85.24                      & 69.18                      & 85.99                       & 70.17                      & 84.79                      & 71.48                      & 82.55                      & 74.31                      & 80.13                      & 74.36                      \\
\rowcolor[HTML]{C0C0C0}
\textbf{NAP-M}            & 35.58                       &    93.32                   & 15.29                      & 96.94                      & 66.86                       & 86.62                      & 57.64                       & 88.98                      & 27.85                       & 93.93                      & 86.00                      & 70.89                      & 48.20                       & 88.45                      \\ % \midrule
ReAct                                                 & 83.81                     & 81.41                     & 77.78                     & 78.95                     & 65.27                     & 86.55                     & 60.08                     & 87.88                     & 25.55                     & 94.92                     & 82.65                     & 74.04                     & 62.27                     & 84.47                     \\
\rowcolor[HTML]{C0C0C0}
\textbf{NAP-R}                                 & 14.19                     & 96.52                     & 17.22                     & 96.16                     & 16.72                     & 96.54                     & 17.16                     & 96.64                     & 5.73                      & 98.76                     & 82.54                     & 74.46                     & 25.71                     & 93.18                     \\ \bottomrule
\end{tabular}
}
\end{table}
%%%%%%%%%%%%%%%%%%%%%%%%%%%%%%%%%%%%%%%%%%%%%%%%%%%%%%%%%%%%%%%%%%%%%%%%%%%
\section{How to find an optimal parameter $w$?}
\label{appendix:w}
When combining NAP with different OOD detection methods, the optimal weight parameter \(w\) varies. To obtain the optimal parameter, we utilized a set of data transformation techniques (including Gaussian noise, shot noise, impulse noise, defocus blur, glass blur, motion blur, zoom blur, snow, frost, fog, brightness, contrast, elastic transform, pixelate, jpeg compression) to generate a corrupted dataset based on the ID dataset, serving as pseudo OOD data. Utilizing this set of OOD data, we employed a binary search method to find the optimal \(w\). Through experimentation with various datasets and methods, we found that this search approach quickly identifies the optimal \(w\), which generalizes well to real OOD datasets. The values of \(w\) used in our experiments are summarized in the Table~\ref{tab:w}.
\begin{table}[]
    \centering
    \footnotesize
    \caption{Optimal weight parameter \(w\) for different OOD detection methods across various datasets.\\\quad \\}
    \label{tab:w}
    \begin{tabular}{cc|c|c}
        \hline 
        Method & CIFAR-10 & CIFAR-100 & ImageNet-1k\tabularnewline
        \hline 
        ASH & 0.5 & 0.6 & 0.8\tabularnewline
        DICE & 0.5 & 0.6 & 0.6\tabularnewline
        Energy & 0.4 & 0.4 & 0.6\tabularnewline
        KNN & 0.8 & 0.8 & 0.6\tabularnewline
        MSP & 0.5 & 0.3 & 0.3\tabularnewline
        ReAct & 0.4 & 0.5 & 0.8\tabularnewline
        \hline 
        \end{tabular}
\end{table}

%%%%%%%%%%%%%%%%%%%%%%%%%%%%%%%%%%%%%%%%%%%%%%%%%%%%%%%%%%%%%%%%%%%%%%%%%%%
\section{Performance on Near-OOD detection}
\label{appendix:near}
Given the context of existing research, where CIFAR-10 is commonly used as the ID dataset and datasets such as SVHN and Texture are utilized as OOD datasets, the distinction in data distribution is markedly evident due to the difference in data sources. This conventional setup, however, does not adequately challenge the model with closely related distributions. Therefore, we embark on an experiment utilizing CIFAR-10 and CIFAR-100 as ID and OOD datasets, respectively, to explore the performance of NAP in scenarios where the data distributions are more closely aligned. This approach aims to assess the robustness of NAP in distinguishing between datasets with subtle differences in distribution yet distinct semantic features.

\textbf{Conclusion:} Our findings confirm that NAP is capable of effectively functioning in scenarios where the ID and OOD data distributions are closely related, showcasing its utility in near-OOD detection tasks. Table~\ref{tab:cifar10-cifar100} demonstrates the effectiveness of NAP variants (NAP-E, NAP-R, and NAP-A) in comparison to baseline methods (Energy, ReAct, and ASH) for the task of near-OOD detection between CIFAR-10 and CIFAR-100 datasets.

\begin{table}[!h]
\scriptsize
\centering
\caption{Result of CIFAR-10 vs. CIFAR-100\\\quad\\}
\label{tab:cifar10-cifar100}
\begin{tabular}{@{}lcclcclcc@{}}
\toprule
       & FPR95 & \multicolumn{1}{c|}{AUROC} &       & FPR95 & \multicolumn{1}{c|}{AUROC} & & FPR95 & AUROC\\ \midrule
Energy & 50.74      & \multicolumn{1}{c|}{89.76}      & ReAct   & 48.77      & \multicolumn{1}{c|}{90.55} & ASH   & 48.74      & 89.93     \\
NAP-E  & \textbf{44.38}      & \multicolumn{1}{c|}{\textbf{90.69}}      & NAP-R & \textbf{42.94}      & \multicolumn{1}{c|}{\textbf{90.66}} & NAP-A & \textbf{44.92}      & \textbf{90.07}     \\ \bottomrule
\end{tabular}
\end{table}

To understand the mechanism by which NAP achieves this, it is essential to delve into how neural networks process and distinguish between different types of data. Neural network classifiers are adept at detecting various semantic features through different channels in the penultimate layer. It is this capability that NAP leverages to differentiate between ID and OOD data.
NAP distinguishes between ID and OOD data based on the neural network's high response to specific semantic features of ID data. Thus, in principle, NAP is well-suited for semantic OOD detection, capable of effectively distinguishing between samples from closely related distributions but with different semantics (e.g., CIFAR-10 vs CIFAR-100, as shown in the table below). However, it is important to note that NAP is not intended for more fine-grained tasks, such as pixel-level industrial surface defect detection.

%%%%%%%%%%%%%%%%%%%%%%%%%%%%%%%%%%%%%%%%%%%%%%%%%%%%%%%%%%%%%%%%%%%%%%%%%%%
\section{More examples of activation map visualizaiton}
\label{appendix:vis}
\begin{figure}[!t]
  \centering
  \begin{subfigure}{\linewidth}
    \includegraphics[width=0.49\linewidth]{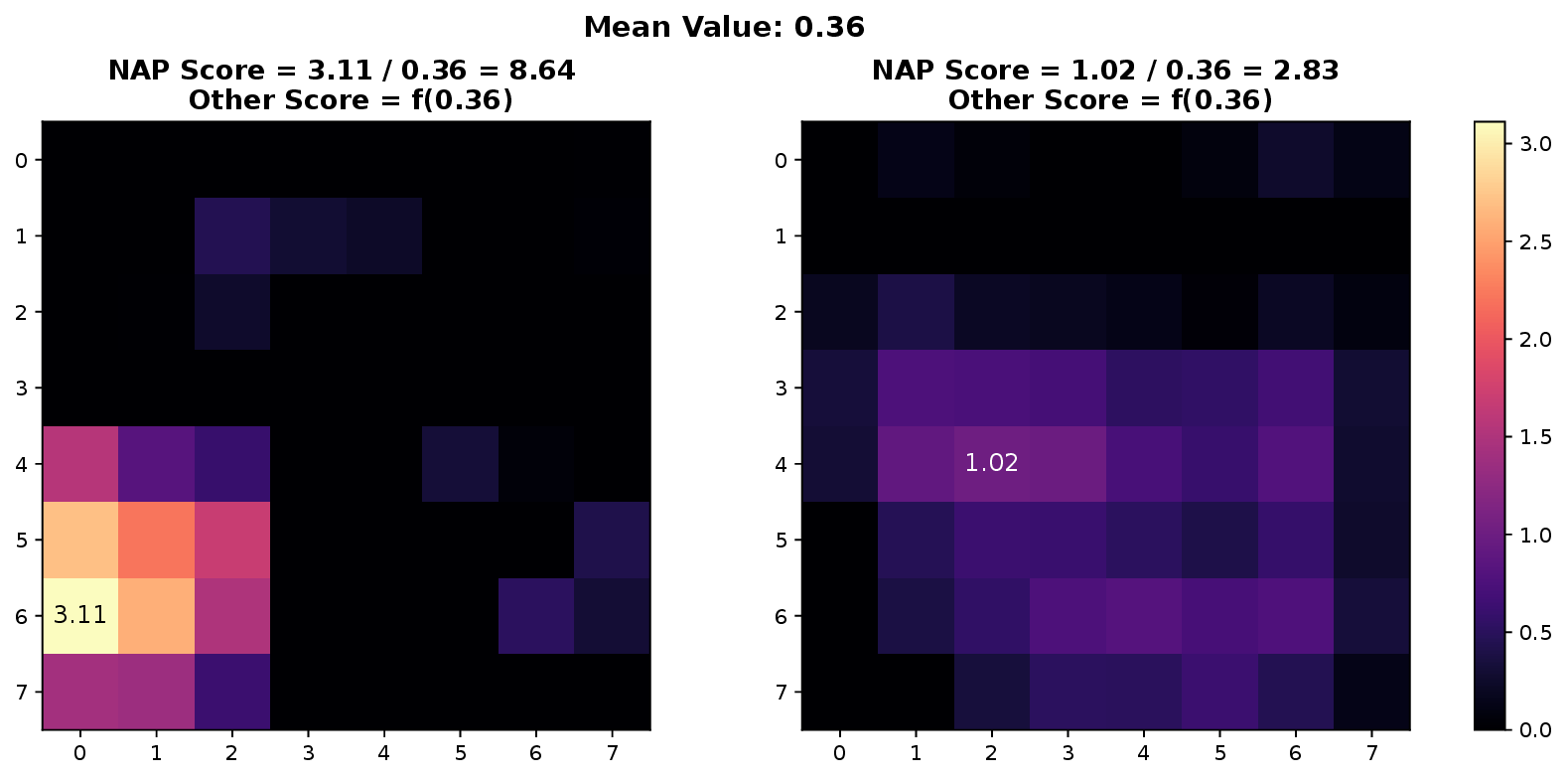}
    \includegraphics[width=0.49\linewidth]{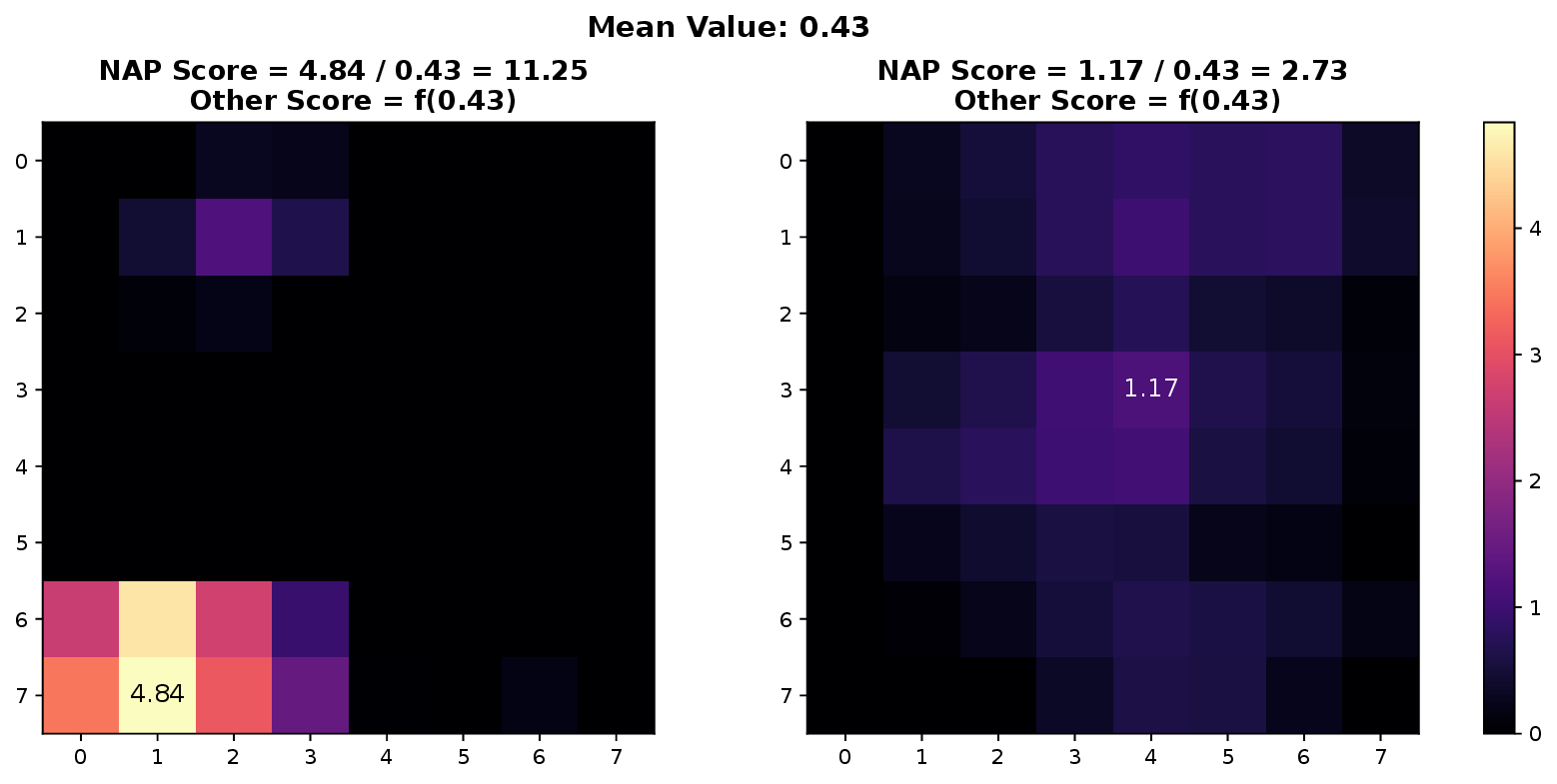}
    \label{fig:act_map0}
  \end{subfigure}

  \begin{subfigure}{\linewidth}
    \includegraphics[width=0.49\linewidth]{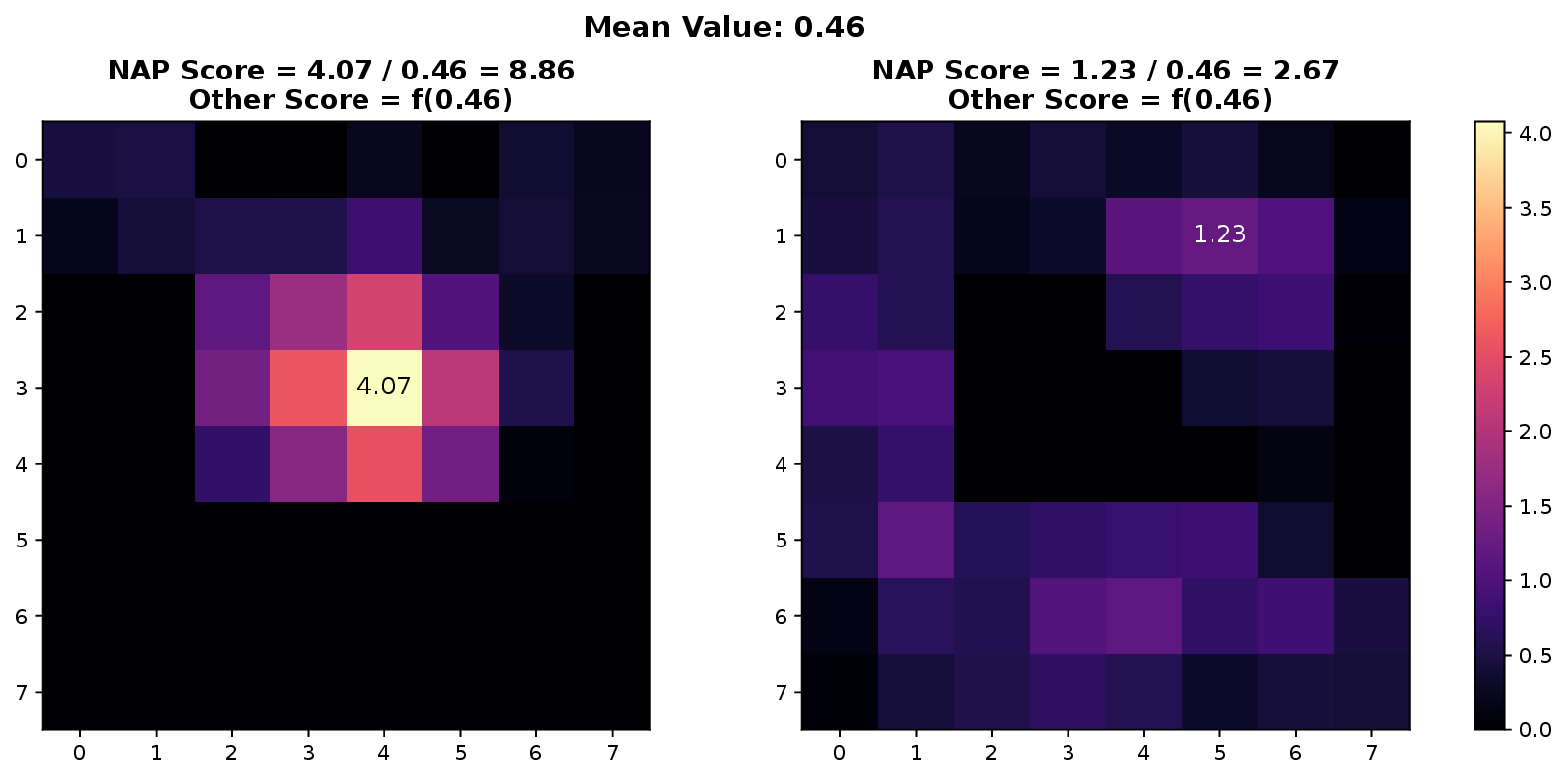}
    \includegraphics[width=0.49\linewidth]{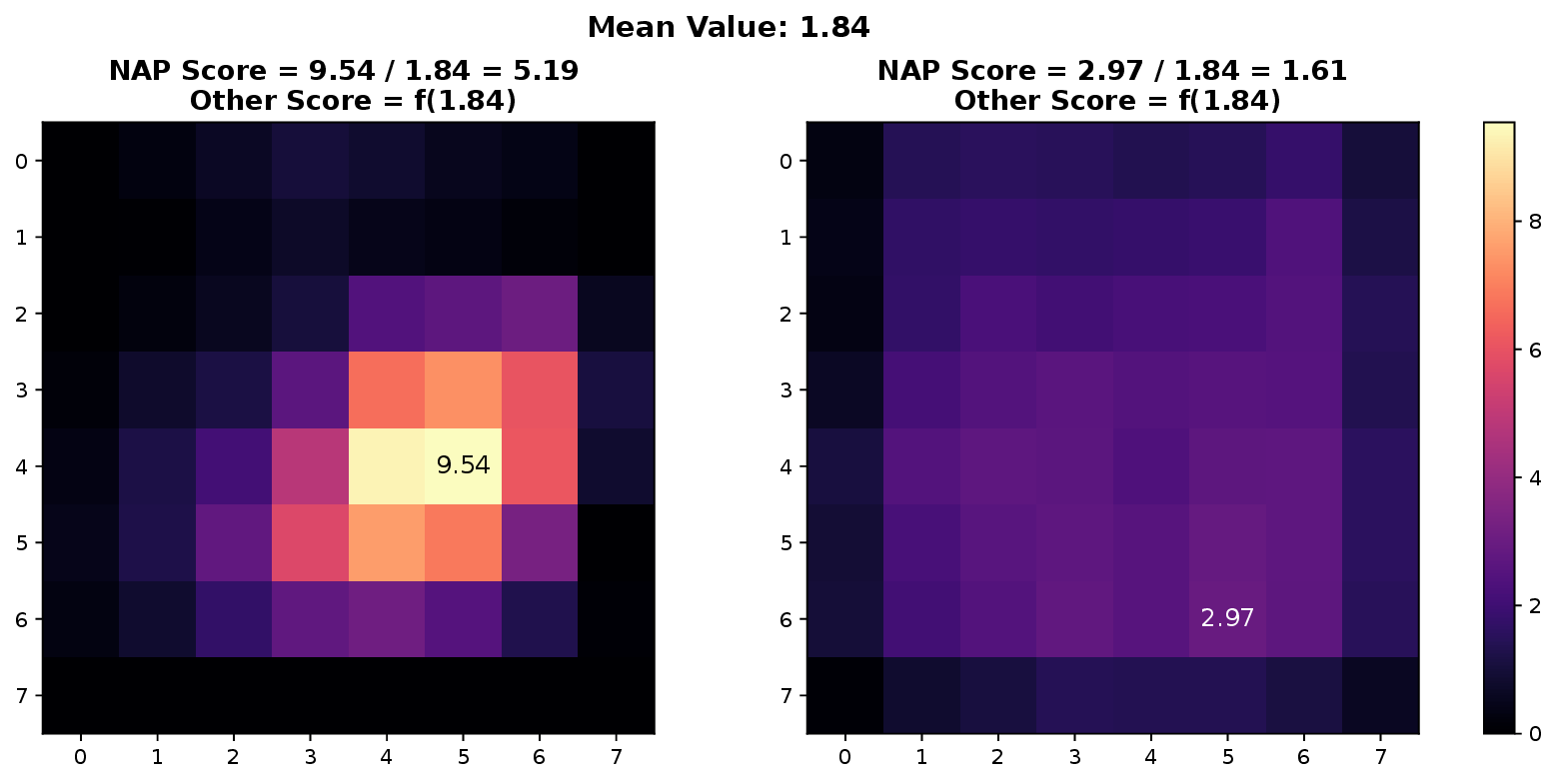}
    \label{fig:act_map1}
  \end{subfigure}

  \begin{subfigure}{\linewidth}
    \includegraphics[width=0.49\linewidth]{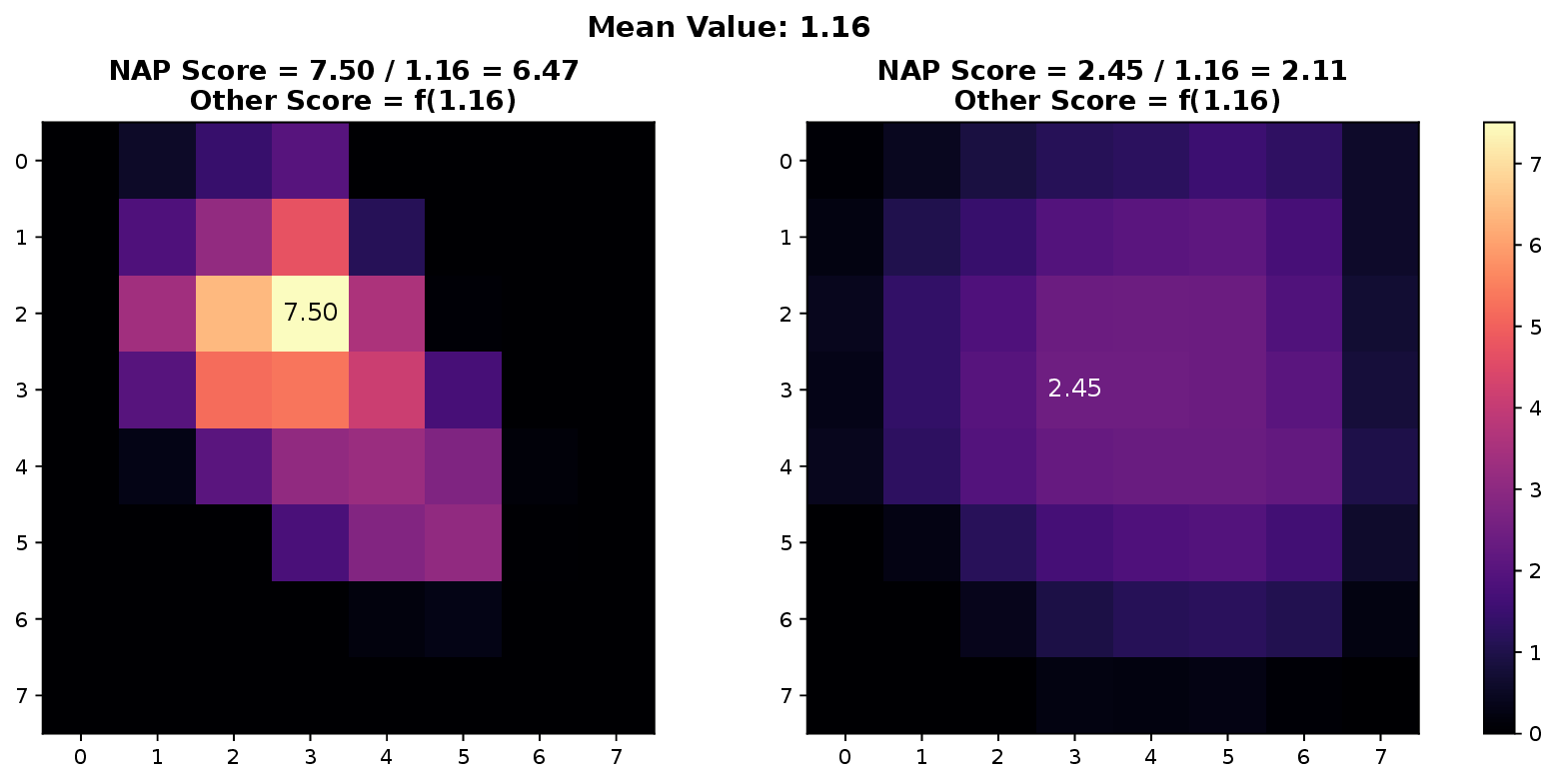}
    \includegraphics[width=0.49\linewidth]{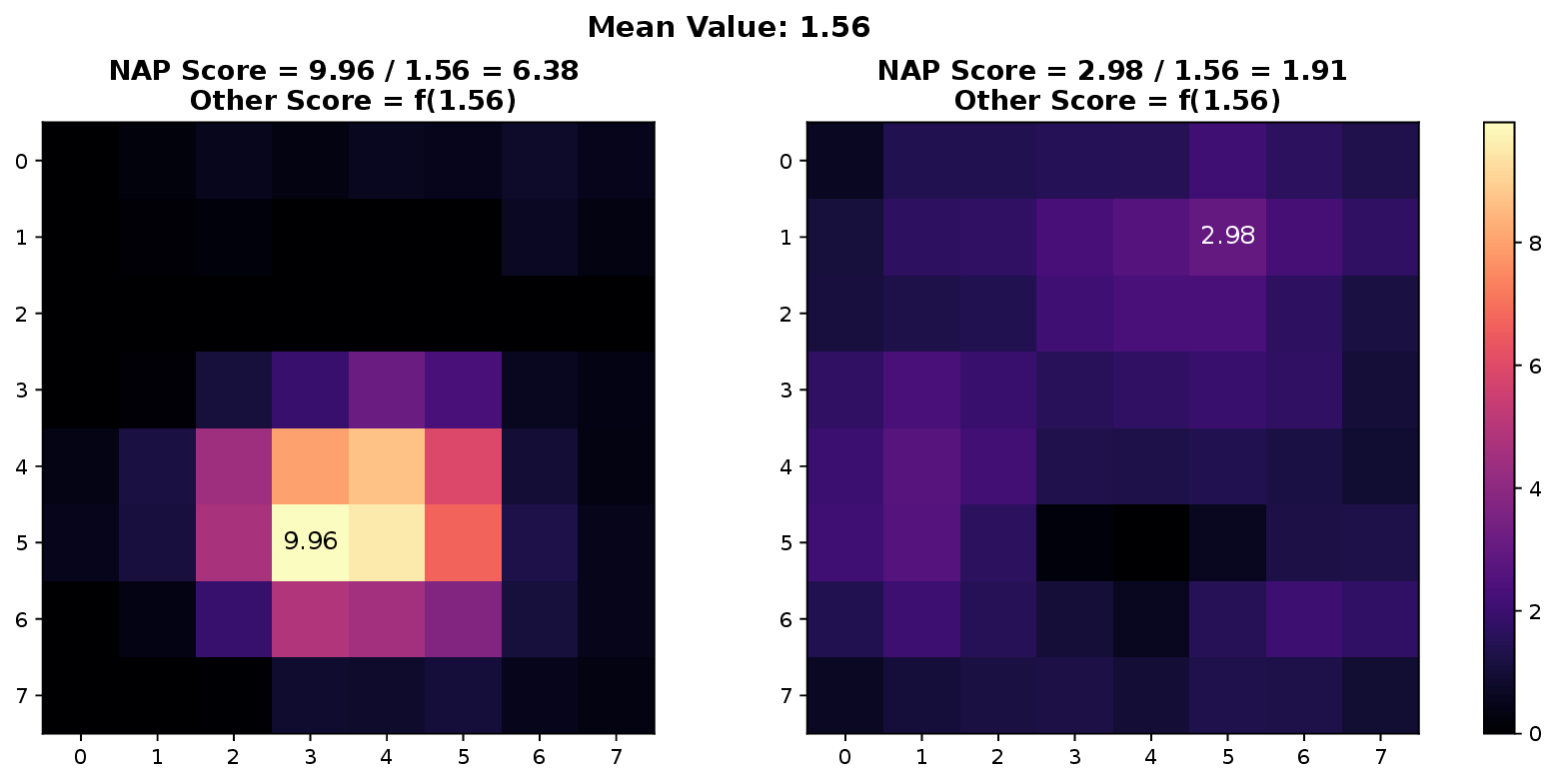}
    \label{fig:act_map2}
  \end{subfigure}

  \begin{subfigure}{\linewidth}
    \includegraphics[width=0.49\linewidth]{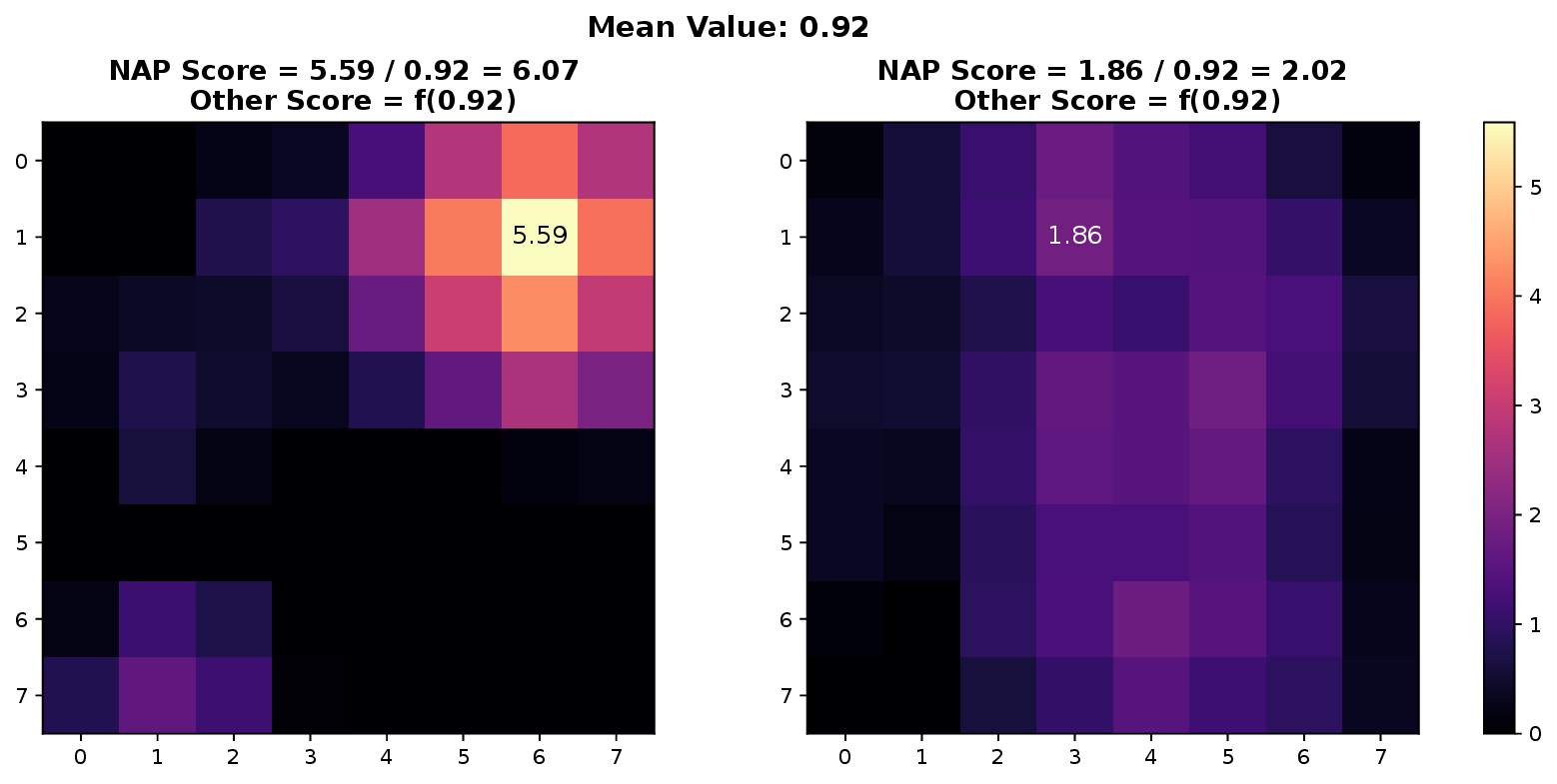}
    \includegraphics[width=0.49\linewidth]{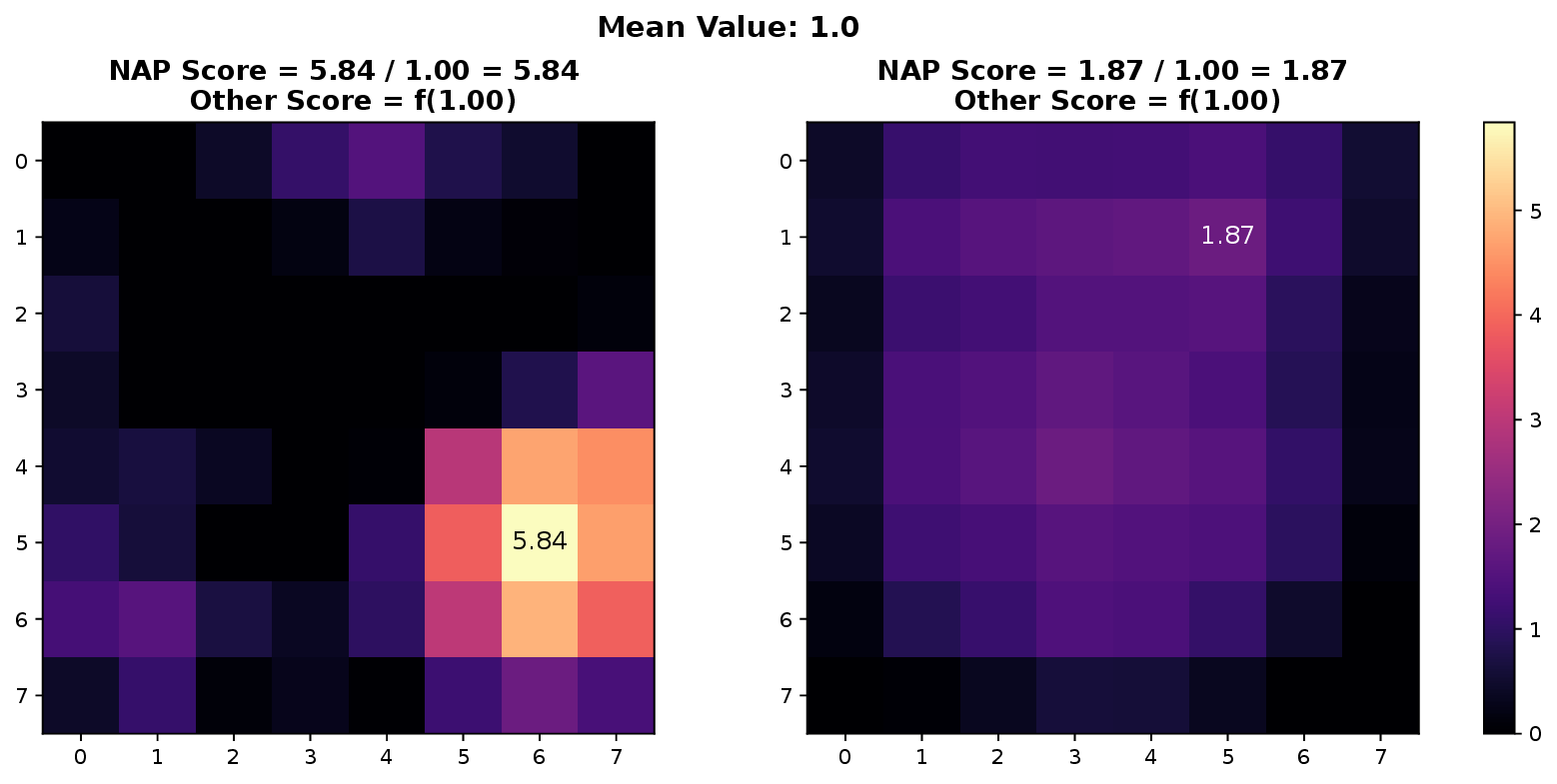}
    \label{fig:act_map3}
  \end{subfigure}

  \begin{subfigure}{\linewidth}
    \includegraphics[width=0.49\linewidth]{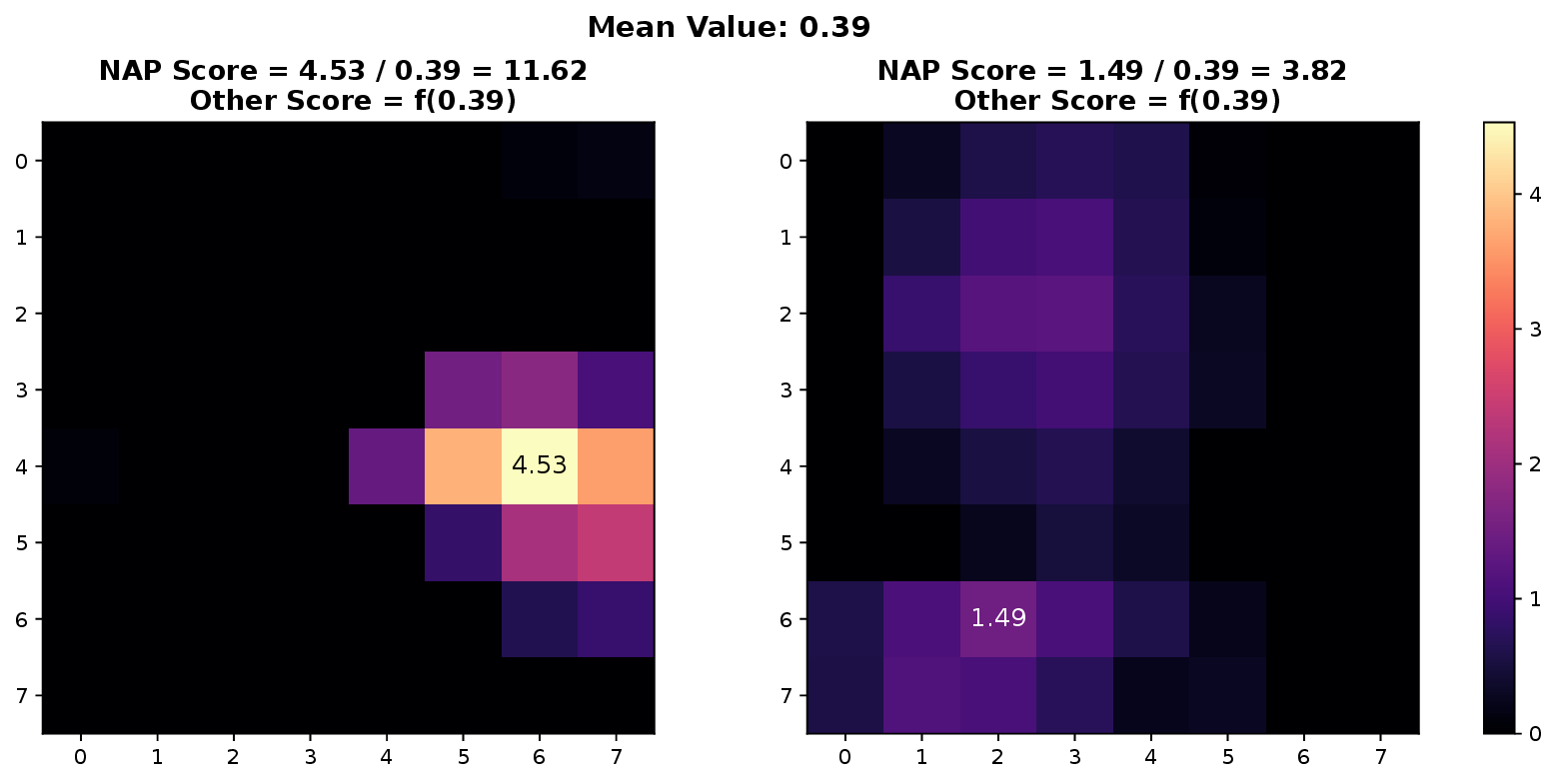}
    \includegraphics[width=0.49\linewidth]{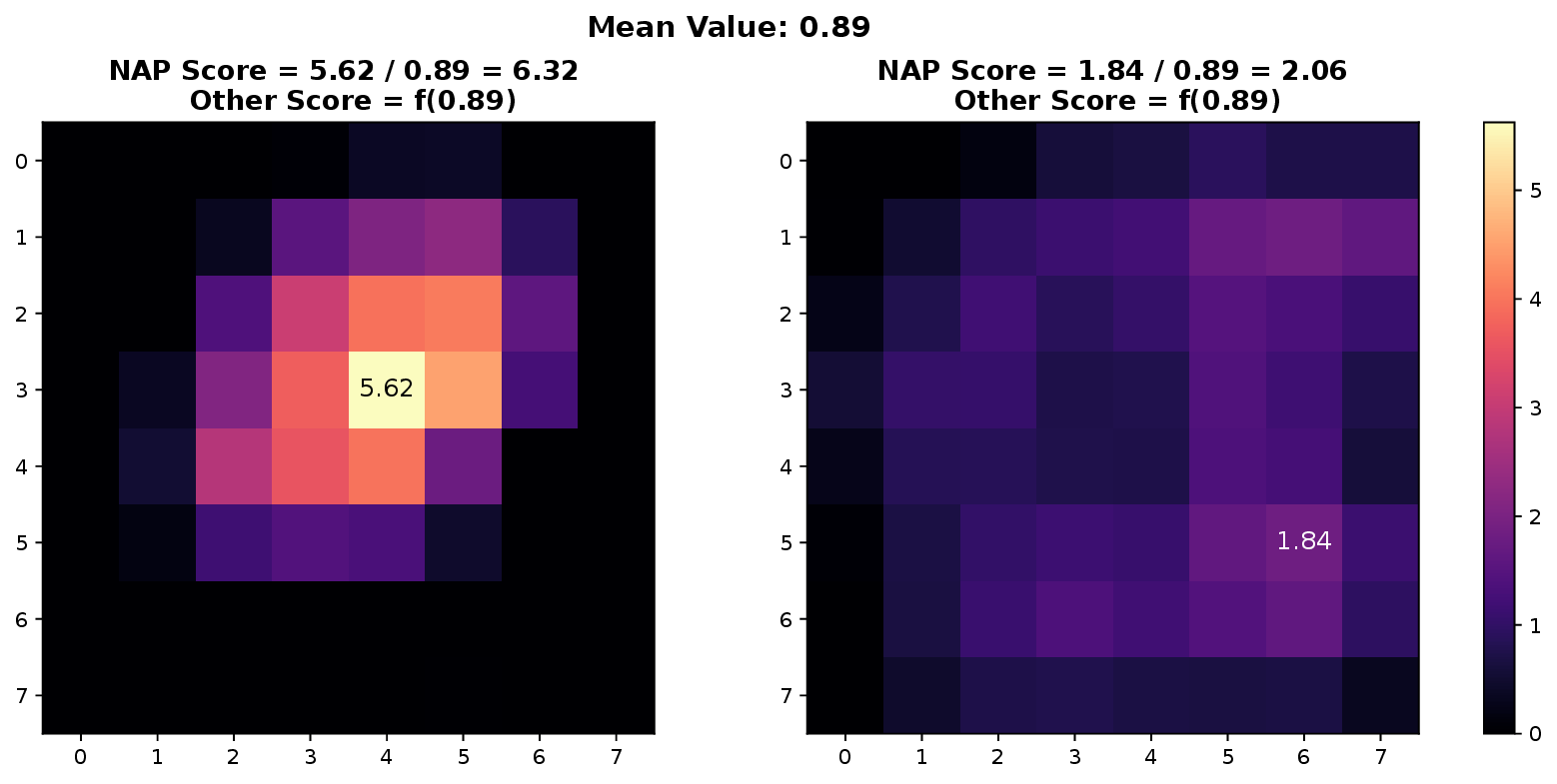}
    \label{fig:act_map4}
  \end{subfigure}

  \caption{\textbf{Global pooling disregards the distribution of activation values within channels, making it challenging to differentiate between ID and OOD samples.} Each pair of images in this figure illustrates the activation maps of the penultimate layer for in-distribution (ID) samples (left) and out-of-distribution (OOD) samples (right) within the same channel. In this layer, different channels typically focus on distinct semantic features. When specific features are present in the image, such as certain regions in the left images of each pair, these areas exhibit very high activation values. Although OOD samples lack these specific features, the model's unfamiliarity with OOD data can lead to unpredictable activations, potentially resulting in weak noise activations (right images). This phenomenon is discussed in detail by \cite{sun2021react}. Existing methods often rely on aggregating activation values for OOD detection. Consequently, the average channel activations of ID and OOD samples are not discriminative, making it difficult for existing methods to distinguish between them. However, the NAP score proposed in this paper effectively differentiates them.}
  \label{fig:act_maps0}
\end{figure}
In this section, we present additional examples of activation map visualizations to further illustrate the challenges and phenomena discussed in our work. Specifically, we examine the penultimate layer's activation maps for both in-distribution (ID) and out-of-distribution (OOD) samples. The visualizations provide insights into how global pooling methods can obscure important distinctions between ID and OOD samples by averaging out the spatial distribution of activation values within channels.

As shown in Figure~\ref{fig:act_maps0}, each pair of images displays the activation maps for a given channel. The left image in each pair corresponds to an ID sample, while the right image corresponds to an OOD sample. Different channels in the penultimate layer are typically tuned to capture specific semantic features present in the training data. For ID samples, these features often result in high activation values in particular regions of the activation map. For example, certain regions in the left images of each pair exhibit very high activation values when the corresponding semantic features are present in the ID sample.

In contrast, OOD samples, which do not contain these specific semantic features, may still produce activation responses due to the model's unfamiliarity with such data. This can result in weak noise activations, as observed in the right images of each pair. This phenomenon highlights the difficulty faced by traditional OOD detection methods that rely on aggregated activation values, as discussed by \cite{sun2021react}. These methods often fail to differentiate between ID and OOD samples because the average channel activations do not provide sufficient discriminative power.

However, the NAP score proposed in this paper addresses this issue by effectively distinguishing between ID and OOD samples based on a more nuanced analysis of activation patterns. The following visualizations exemplify the described behavior and underscore the importance of considering the distribution of activation values within channels for robust OOD detection.
%%%%%%%%%%%%%%%%%%%%%%%%%%%%%%%%%%%%%%%%%%%%%%%%%%%%%%%%%%%%%%%%%%%%%%%%%%%
\section{Limitations}
\label{appendix:limit}
The proposed method relies on the neural network's ability to effectively learn specific semantic features of the ID dataset in the penultimate layer, and the assumption that OOD samples do not possess these features. If OOD samples exhibit similar semantic features or if the neural network is not well-trained, the effectiveness of the proposed method may be compromised.
%%%%%%%%%%%%%%%%%%%%%%%%%%%%%%%%%%%%%%%%%%%%%%%%%%%%%%%%%%%%%%%%%%%%%%%%%%%

\begin{figure}[!h]
  \centering
  \begin{subfigure}{\linewidth}
    \includegraphics[width=0.24\linewidth]{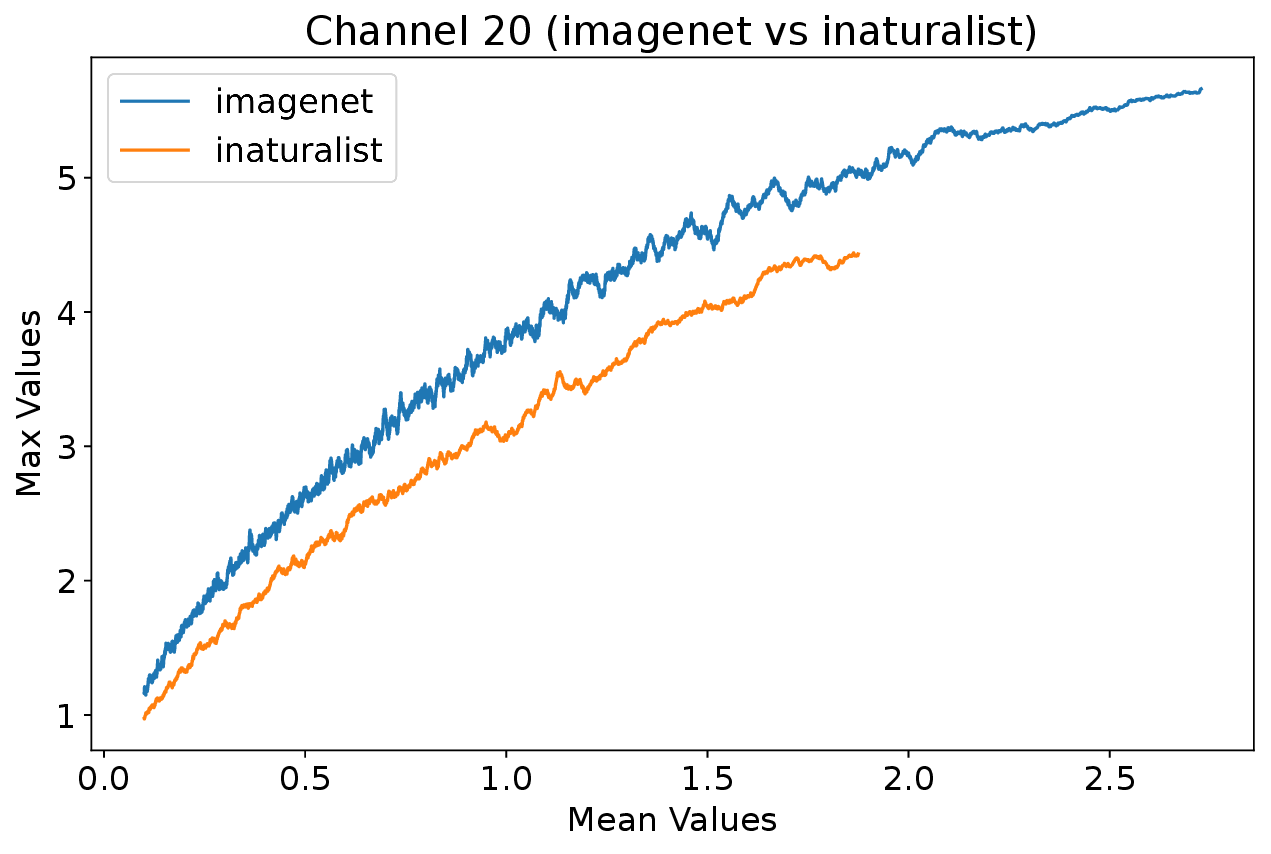}
    \includegraphics[width=0.24\linewidth]{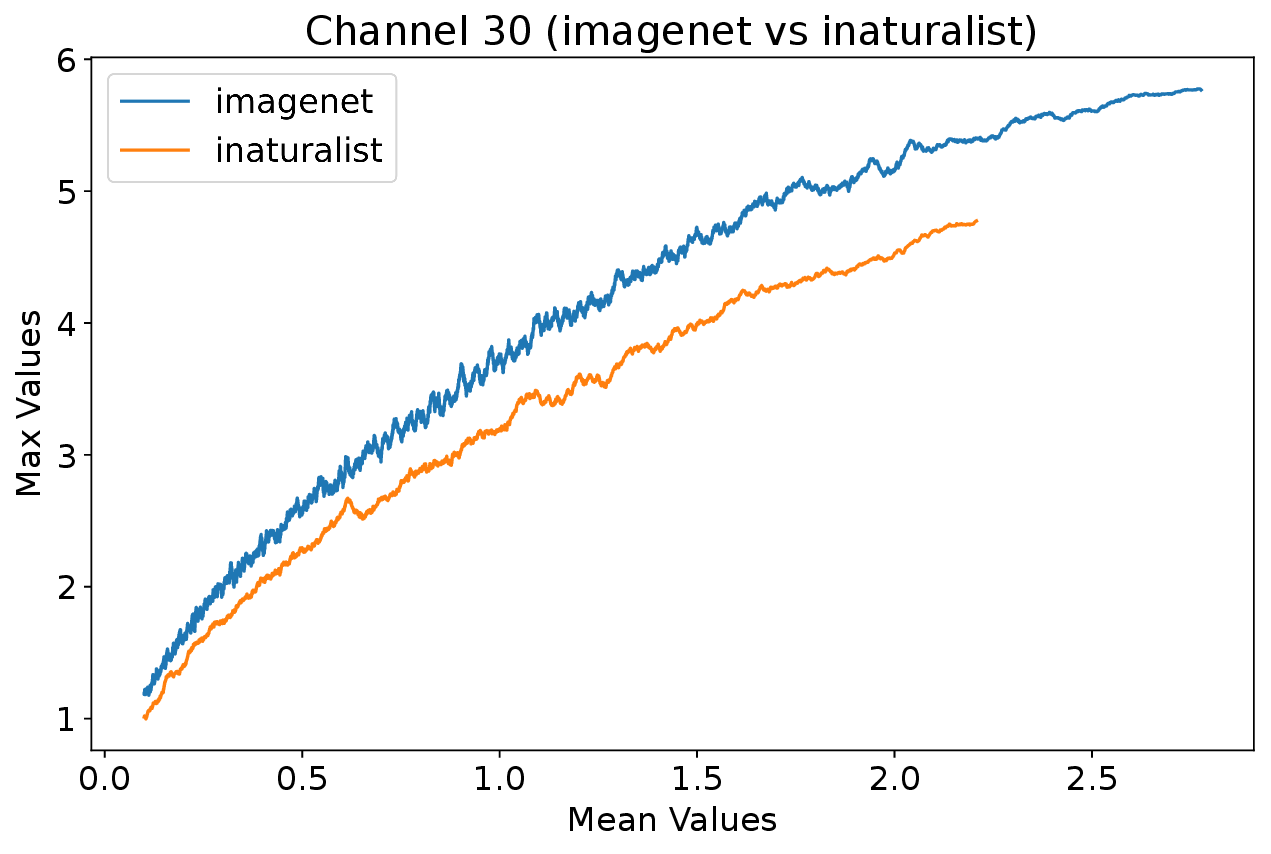}
    \includegraphics[width=0.24\linewidth]{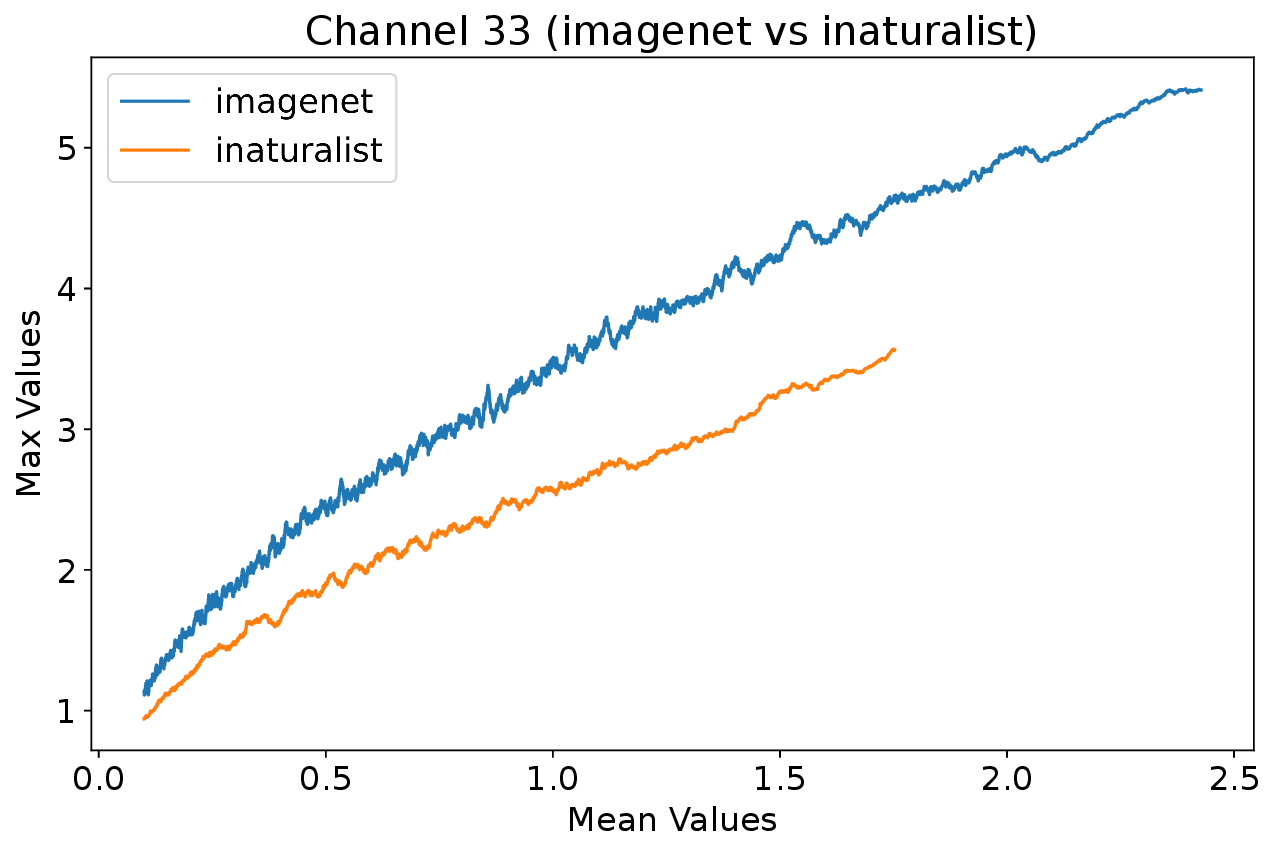}
    \includegraphics[width=0.24\linewidth]{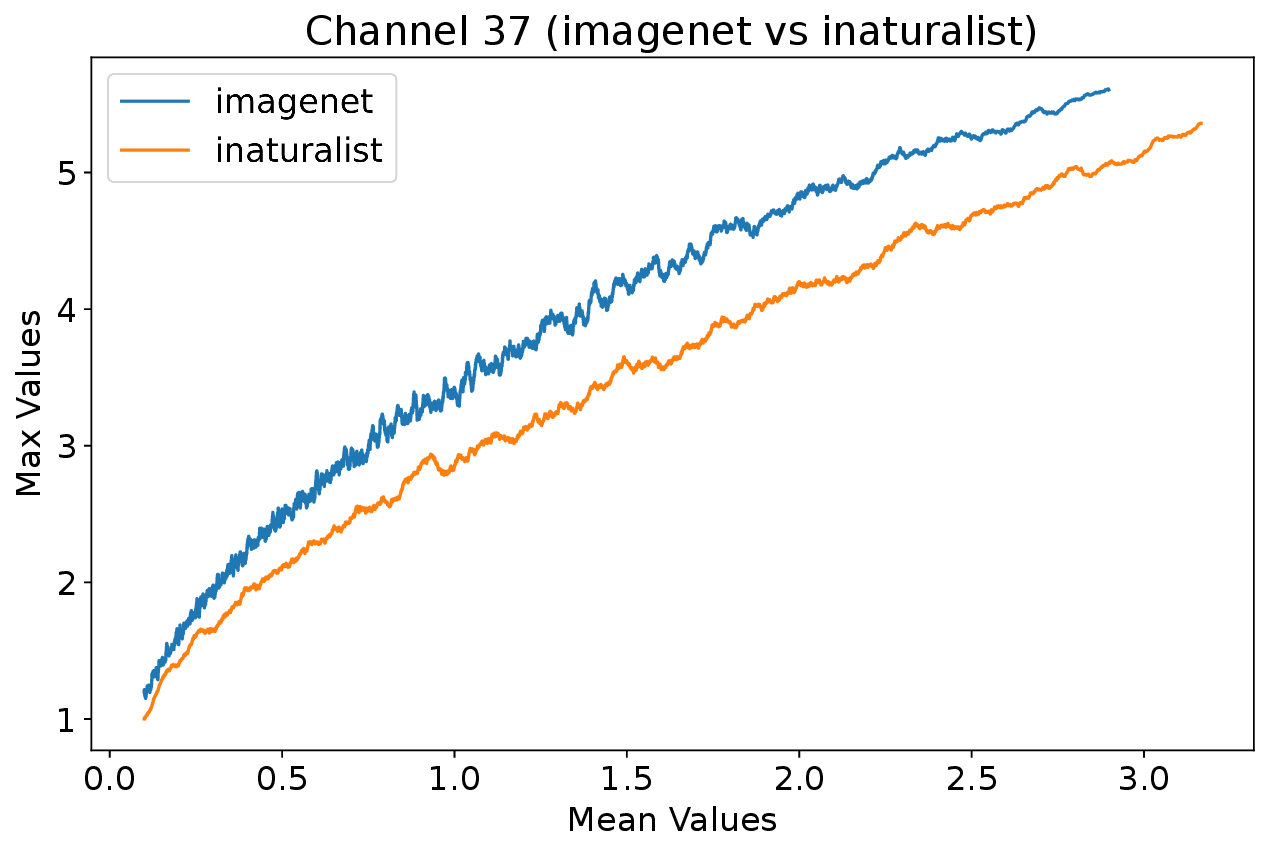}
    \label{fig:mobilenet-r1}
  \end{subfigure}

  \begin{subfigure}{\linewidth}
    \includegraphics[width=0.24\linewidth]{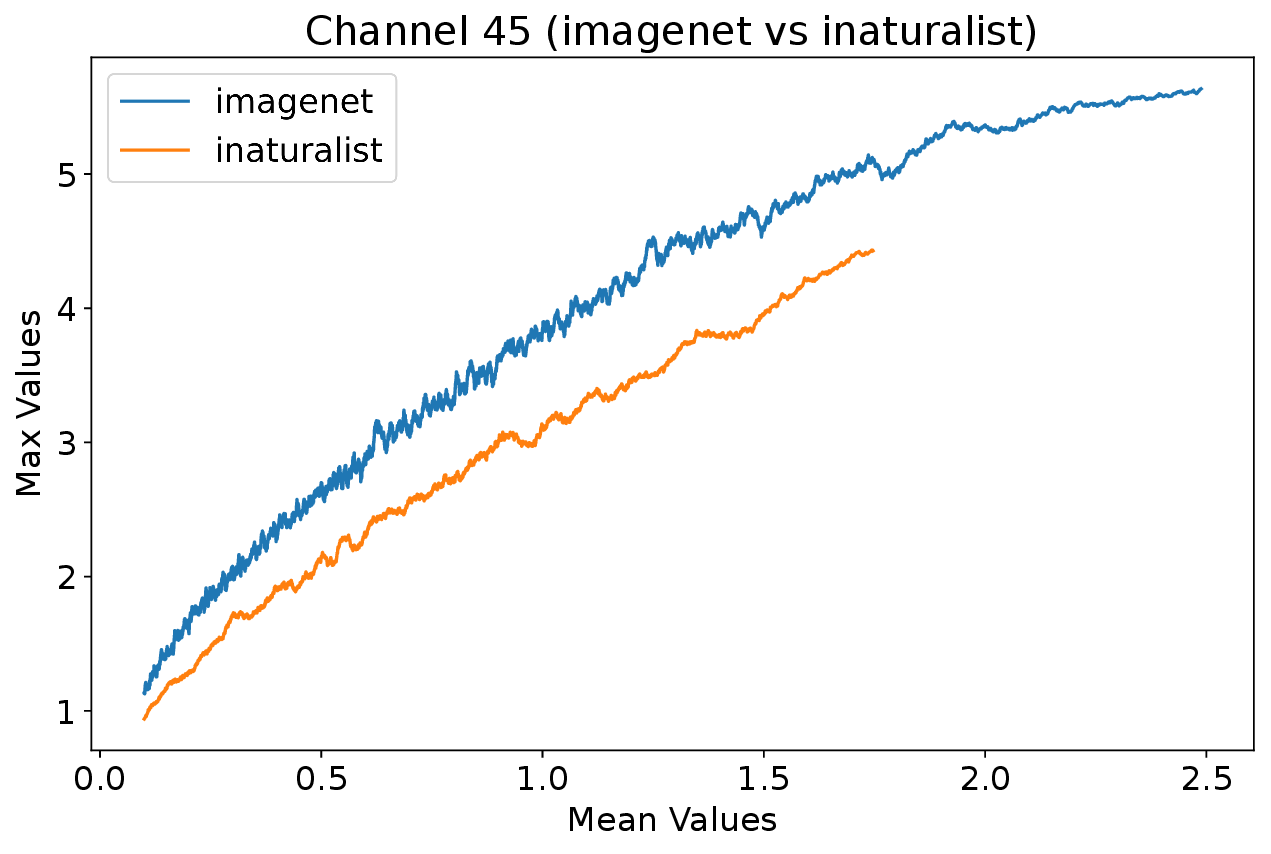}
    \includegraphics[width=0.24\linewidth]{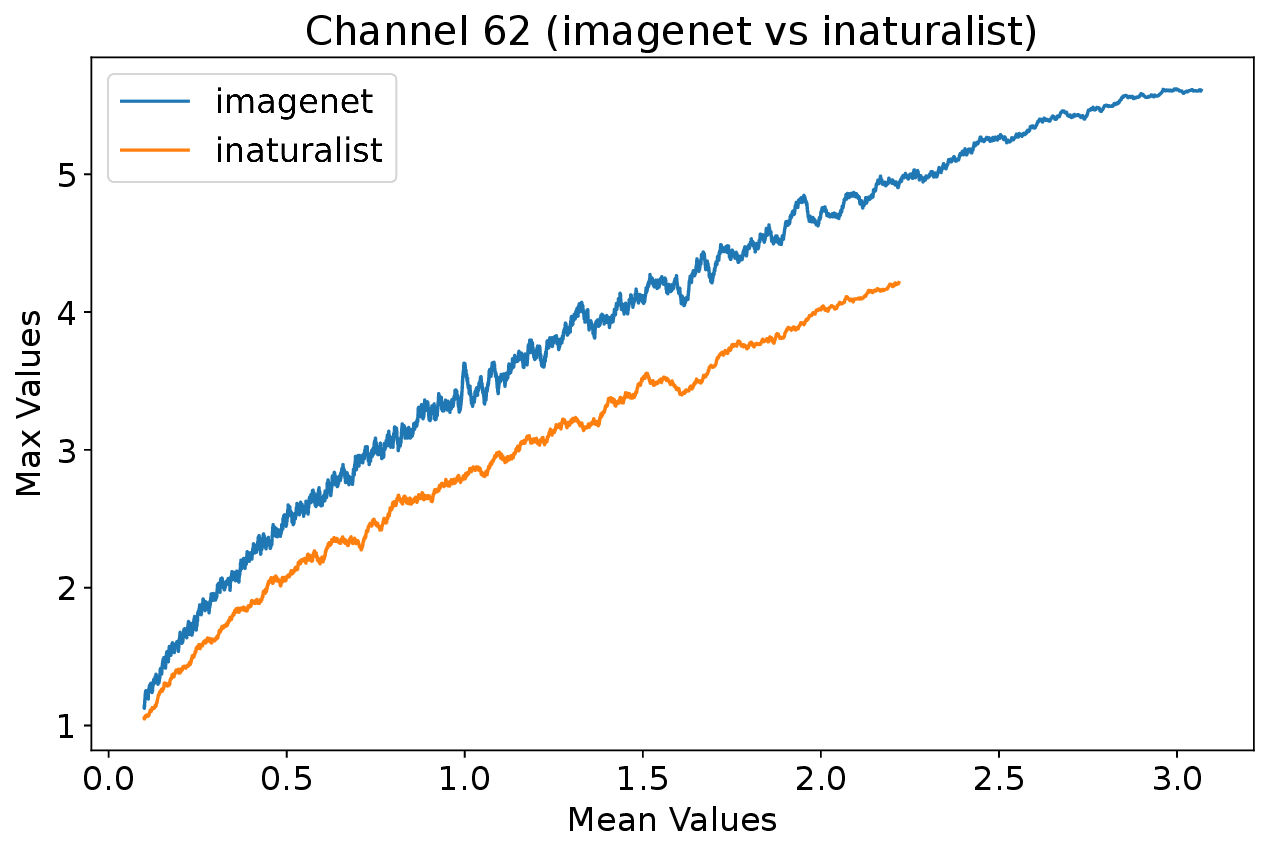}
    \includegraphics[width=0.24\linewidth]{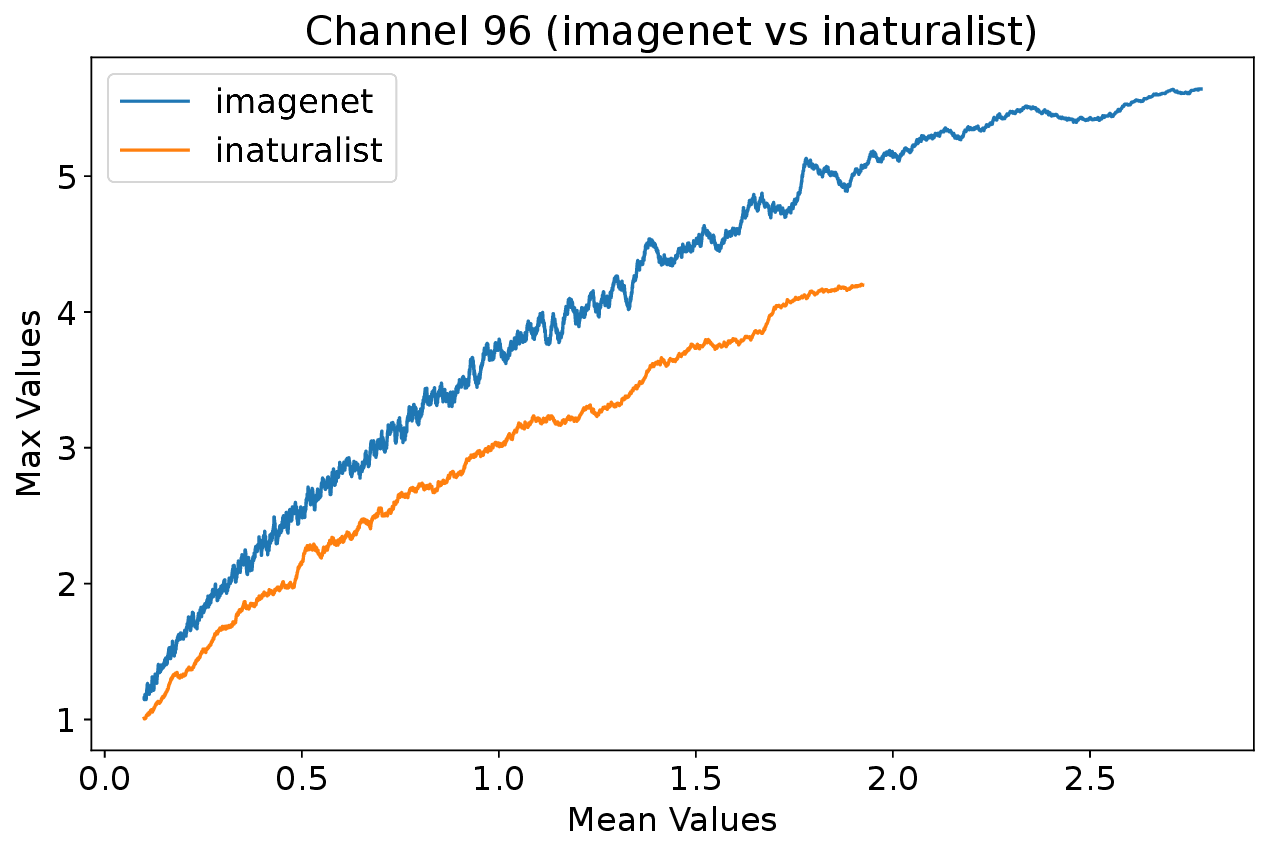}
    \includegraphics[width=0.24\linewidth]{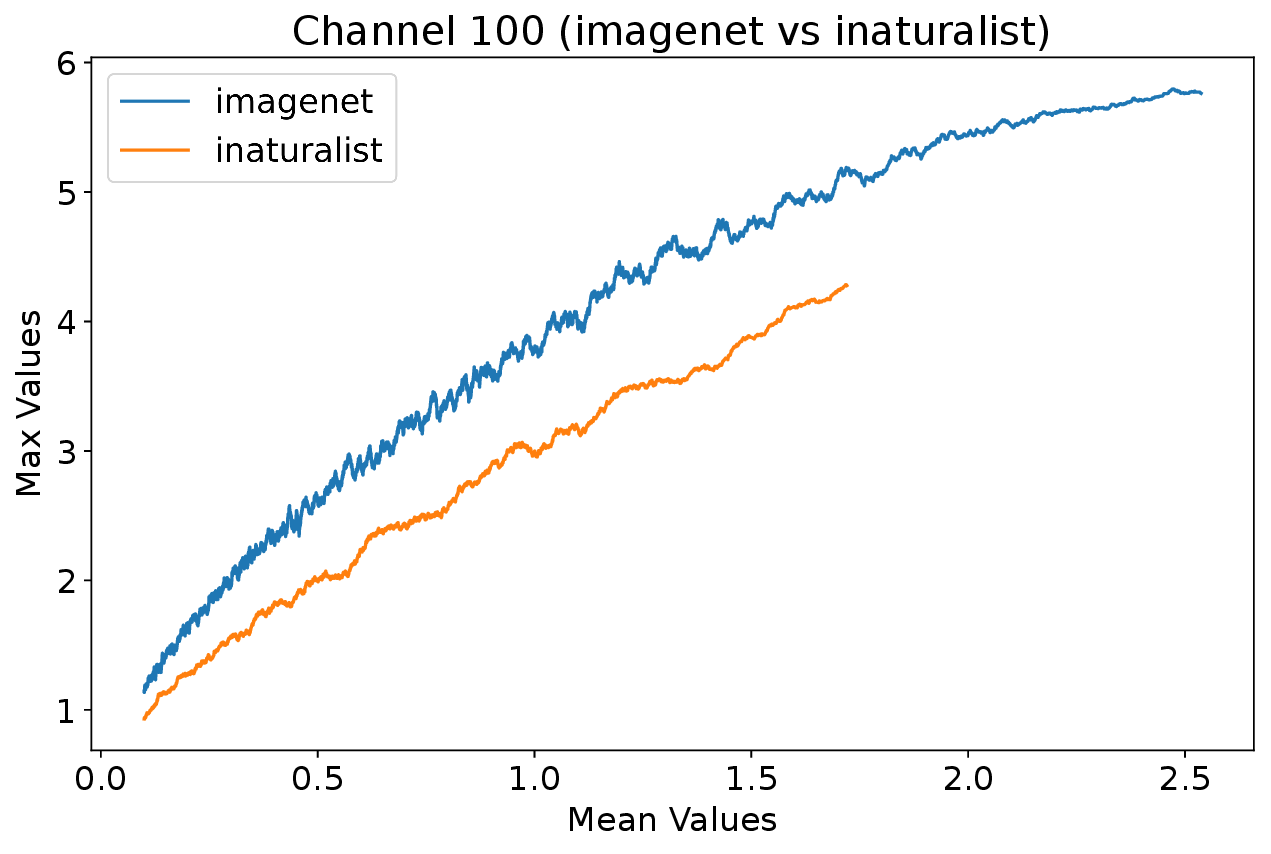}
    \label{fig:mobilenet-r2}
  \end{subfigure}

  \caption{\textbf{Activation distribution at penultimate layer before global pooling operation within the MobileNetV2 architecture~\cite{sandler2018mobilenetv2} applied to ImageNet-1k and iNaturalist datasets~\cite{van2018inaturalist}.} We only include data points with an average activation over $0.1$. The figures show that our Neural Activation Prior (NAP) method is also effective in MobileNetV2, proving that NAP can be applied to different architectures.}
  \label{fig:appendix-e-mobilenet}
\end{figure}

\begin{figure}[!h]
  \centering
  \begin{subfigure}{\linewidth}
  \includegraphics[width=0.24\linewidth]{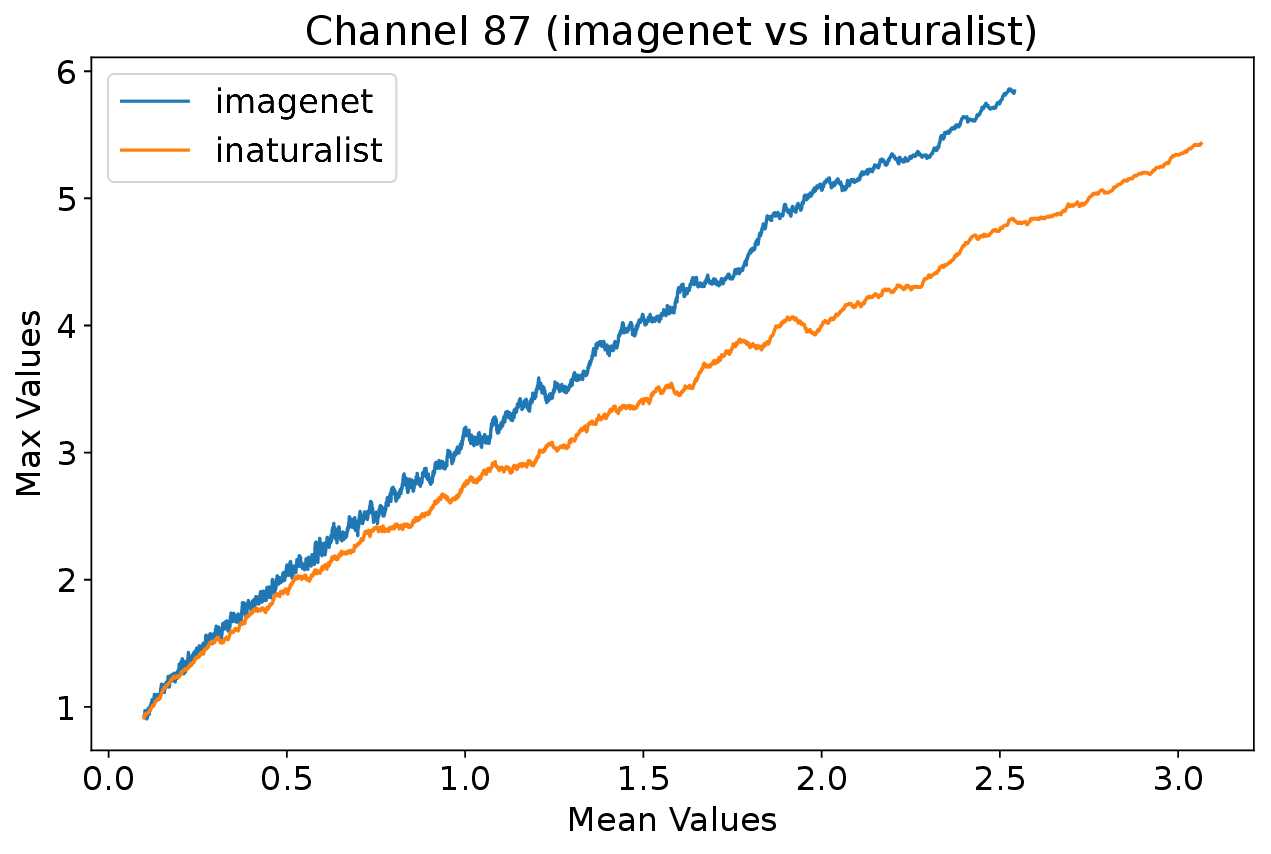}
    \includegraphics[width=0.24\linewidth]{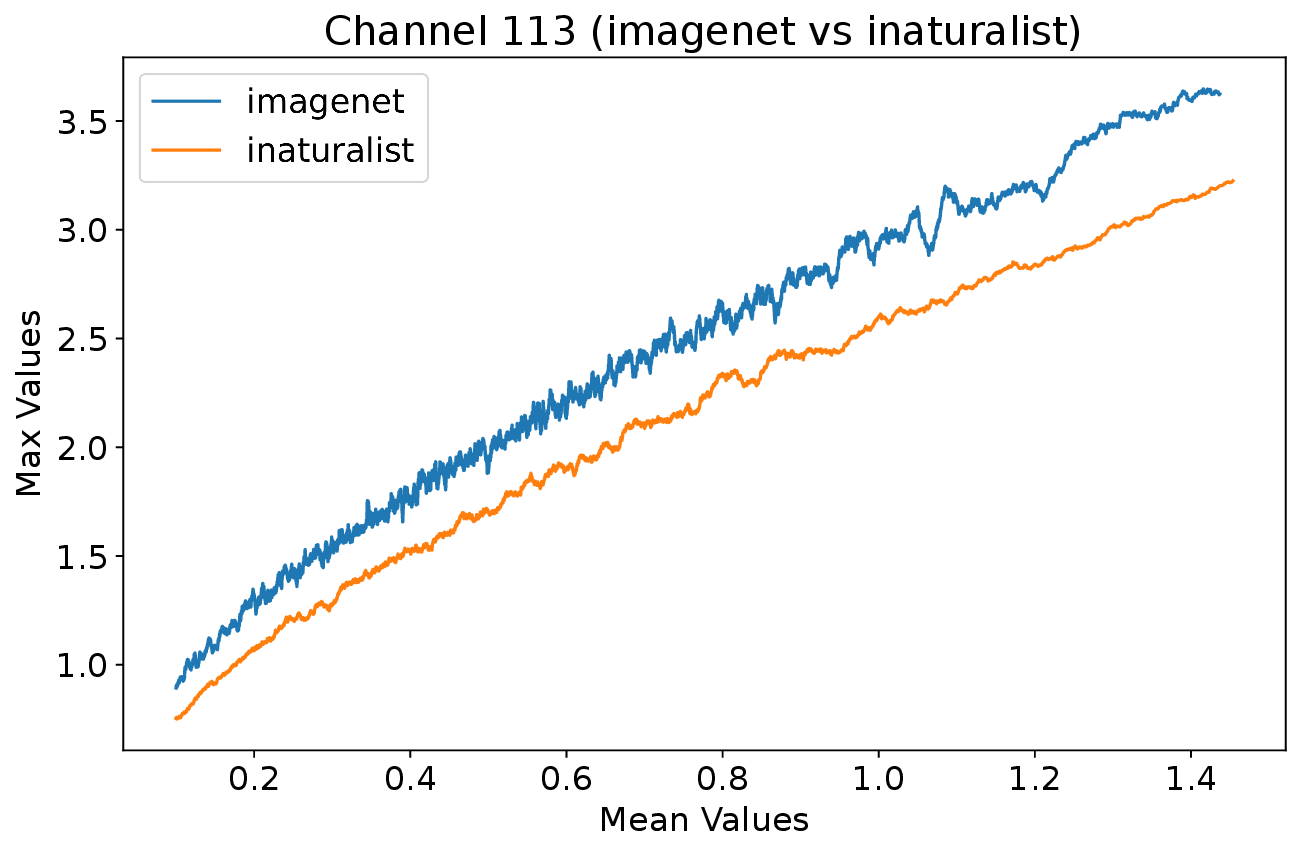}
    \includegraphics[width=0.24\linewidth]{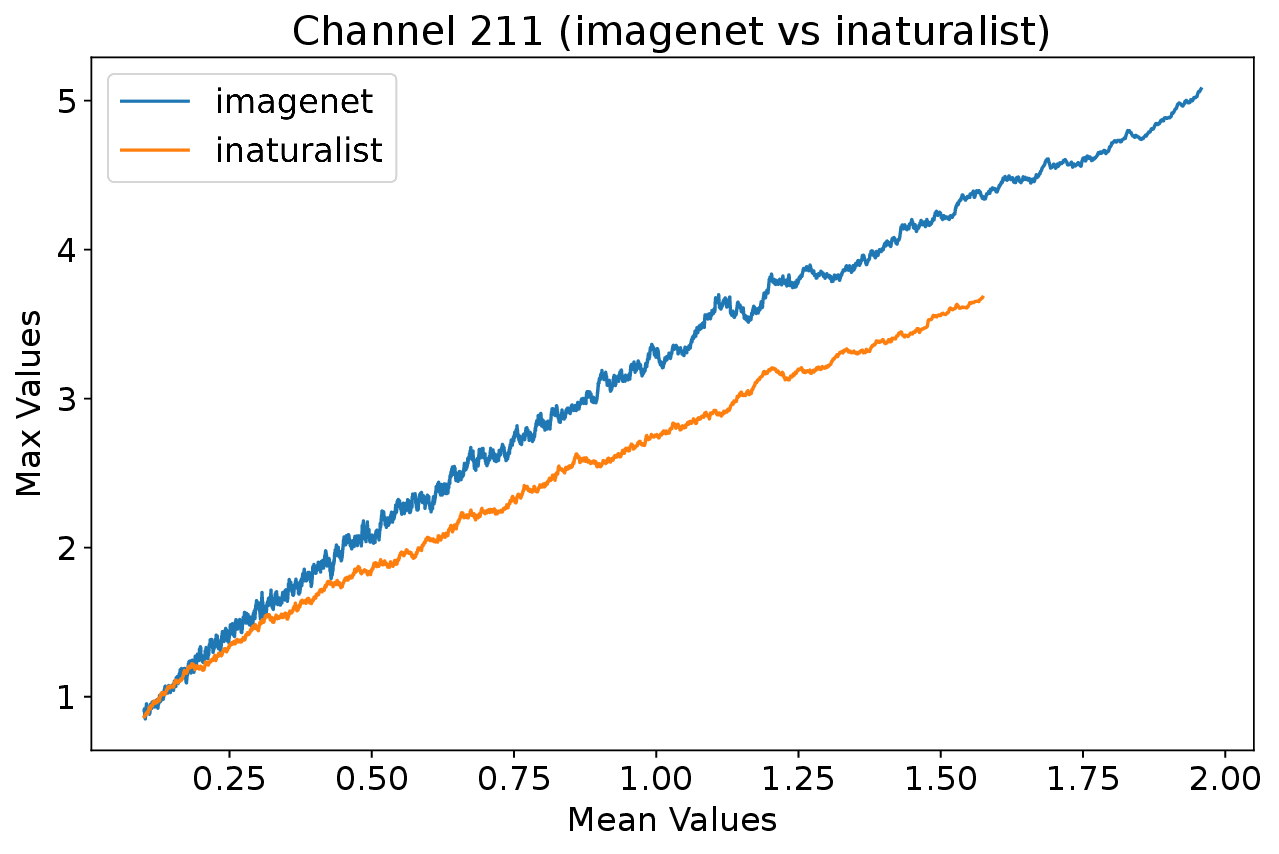}
    \includegraphics[width=0.24\linewidth]{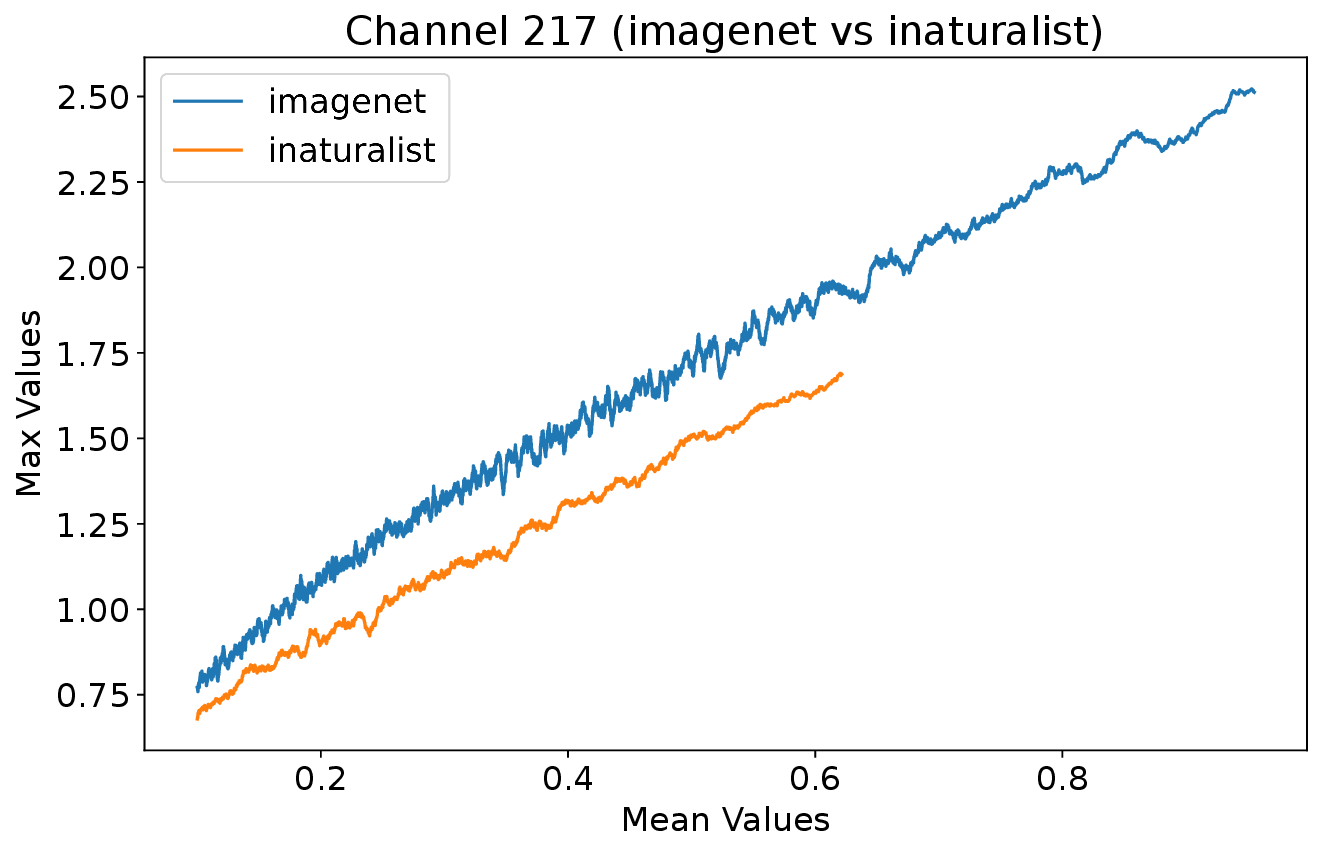}
    \label{fig:resnet-r1}
  \end{subfigure}

  \begin{subfigure}{\linewidth}
    \includegraphics[width=0.24\linewidth]{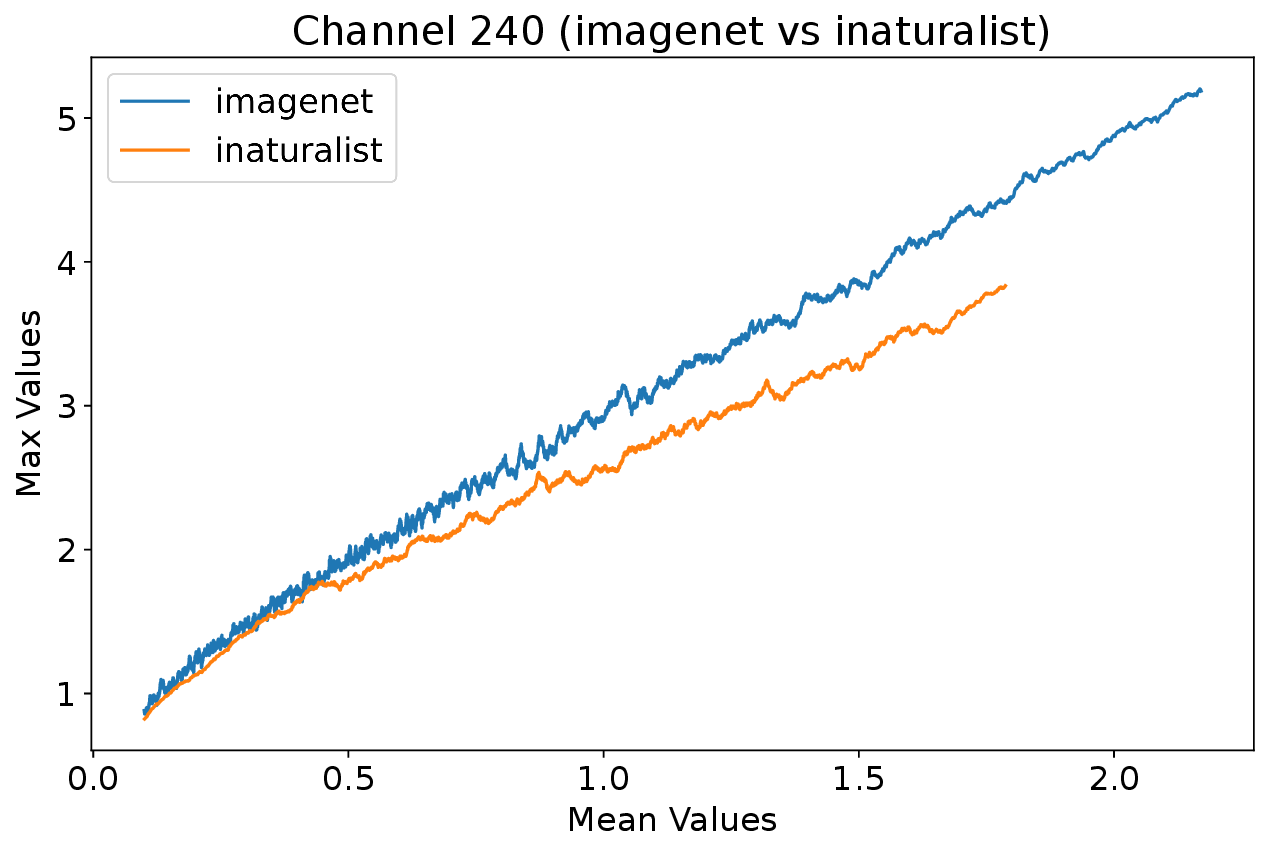}
    \includegraphics[width=0.24\linewidth]{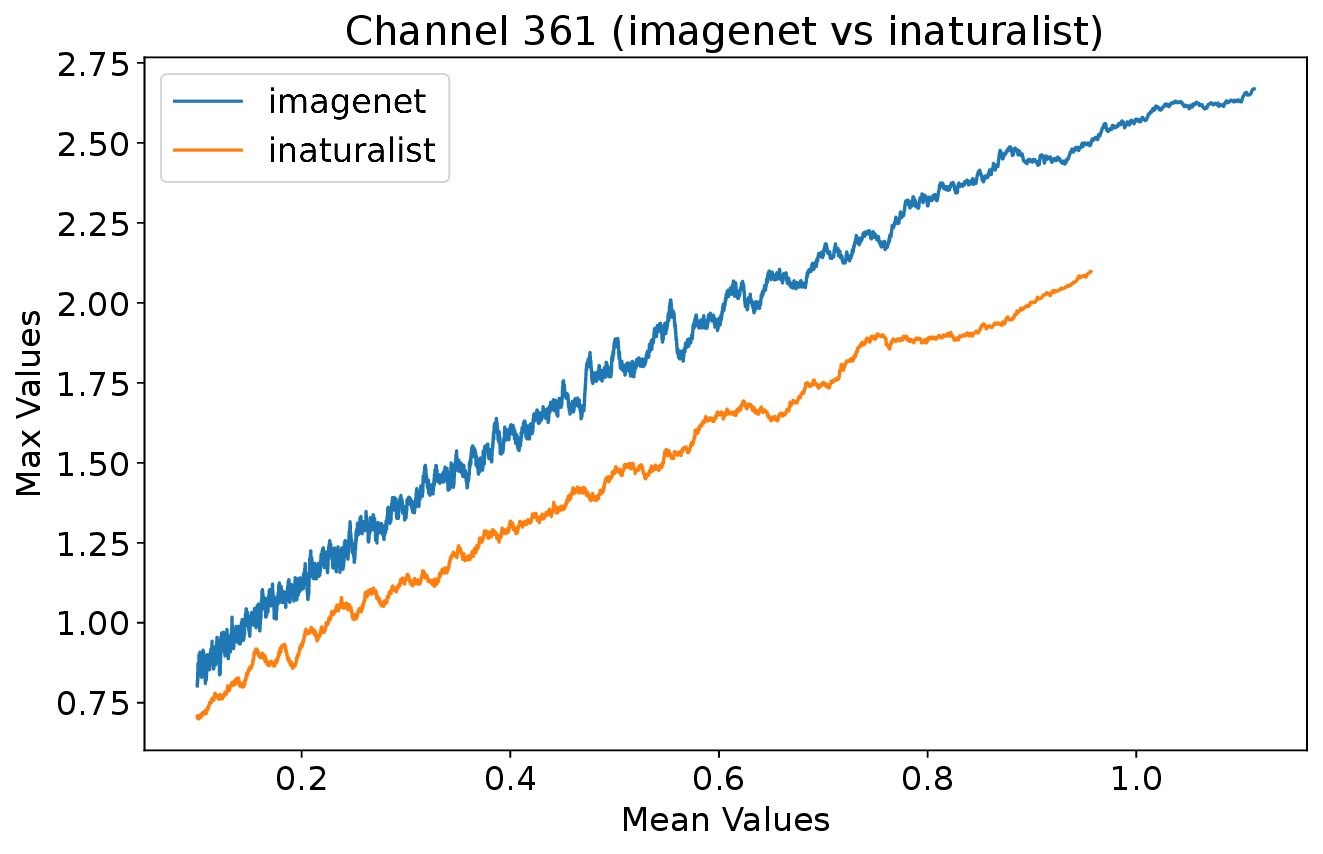}
    \includegraphics[width=0.24\linewidth]{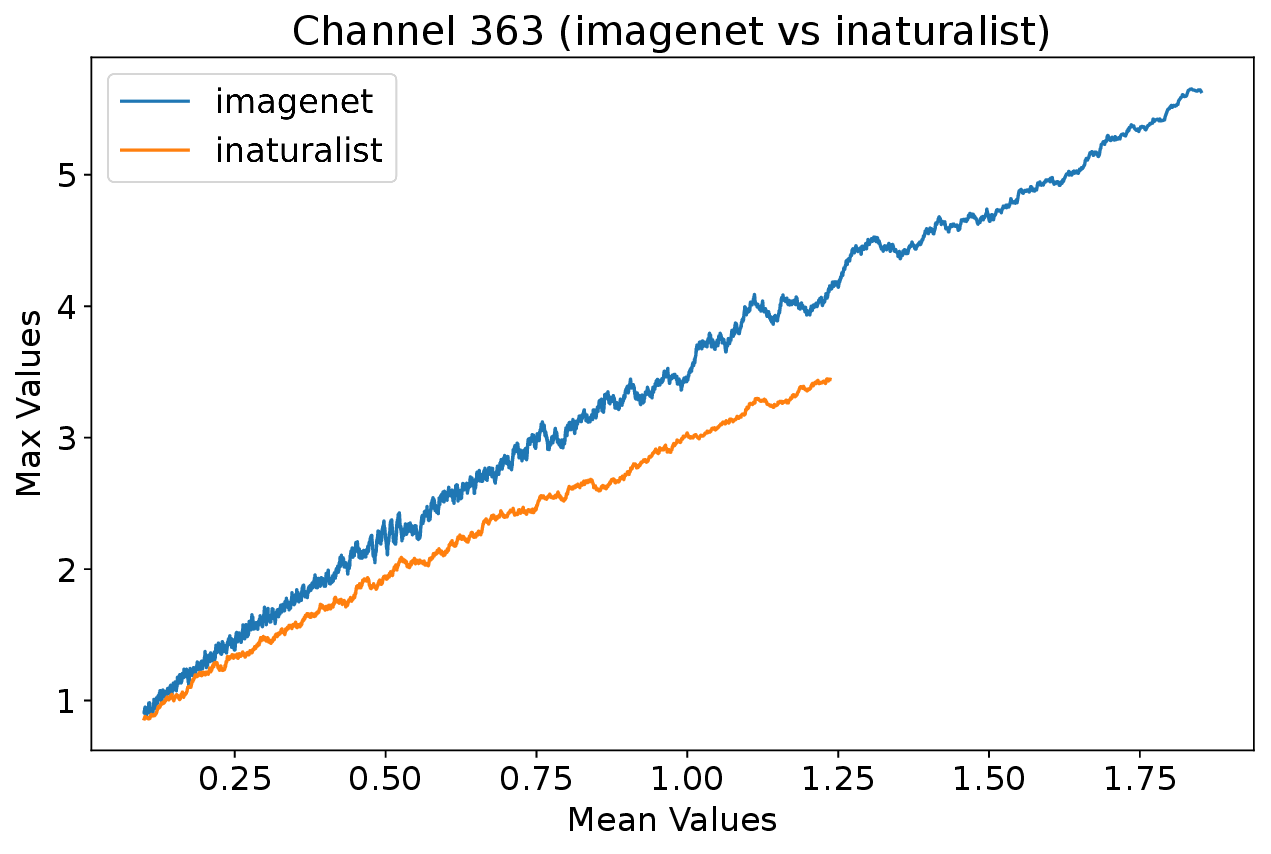}
    \includegraphics[width=0.24\linewidth]{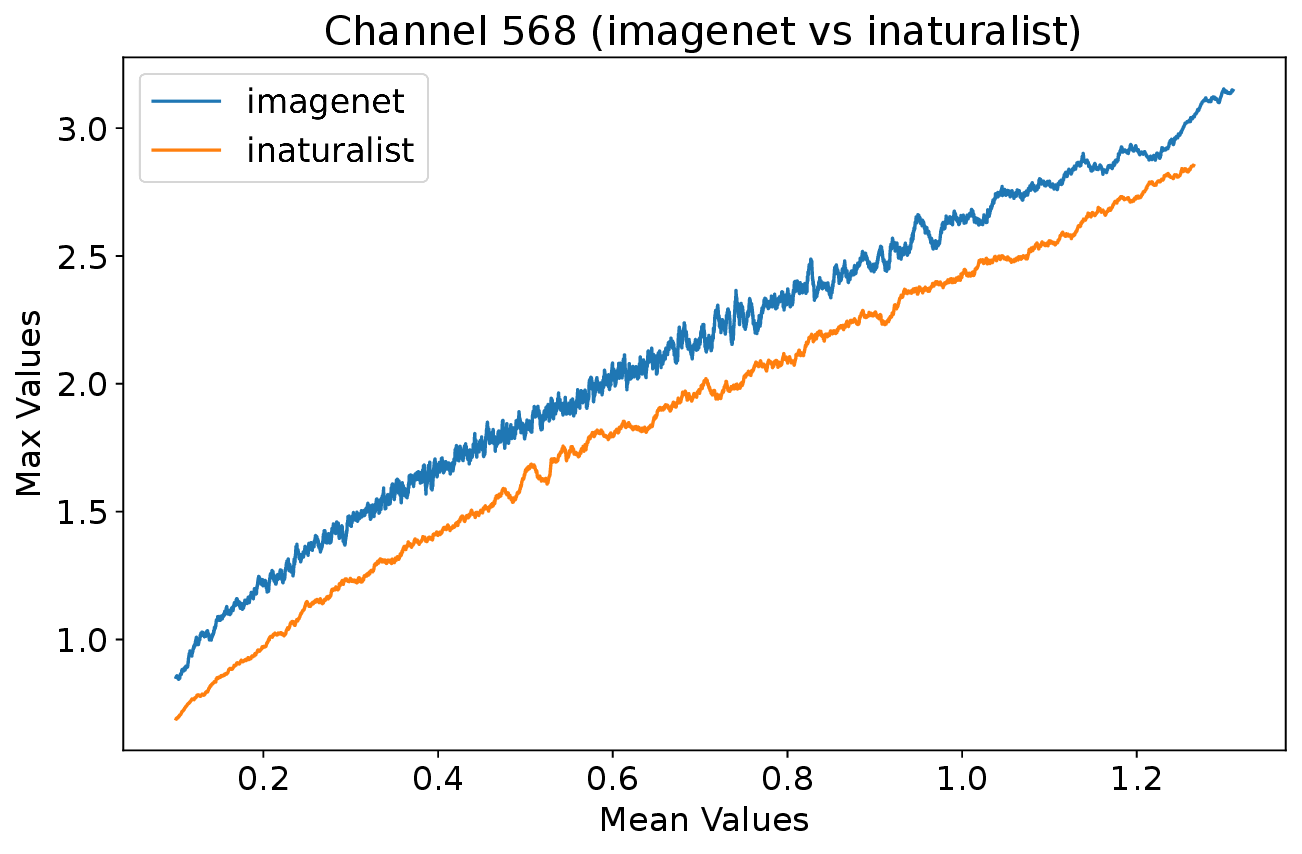}
    \label{fig:resnet-r2}
  \end{subfigure}

  \caption{\textbf{Activation distribution at penultimate layer before global pooling operation within the ResNet50~\cite{he2016deep} architecture applied to ImageNet-1k and iNaturalist datasets~\cite{van2018inaturalist}.} We only include data points with an average activation over $0.1$. The figures show that our Neural Activation Prior (NAP) method is also effective in ResNet50, proving that NAP can be applied to different architectures.}
  \label{fig:appendix-e-resnet}
\end{figure}

\begin{figure}[!h]
  \centering
  \begin{subfigure}{\linewidth}
    \includegraphics[width=0.24\linewidth]{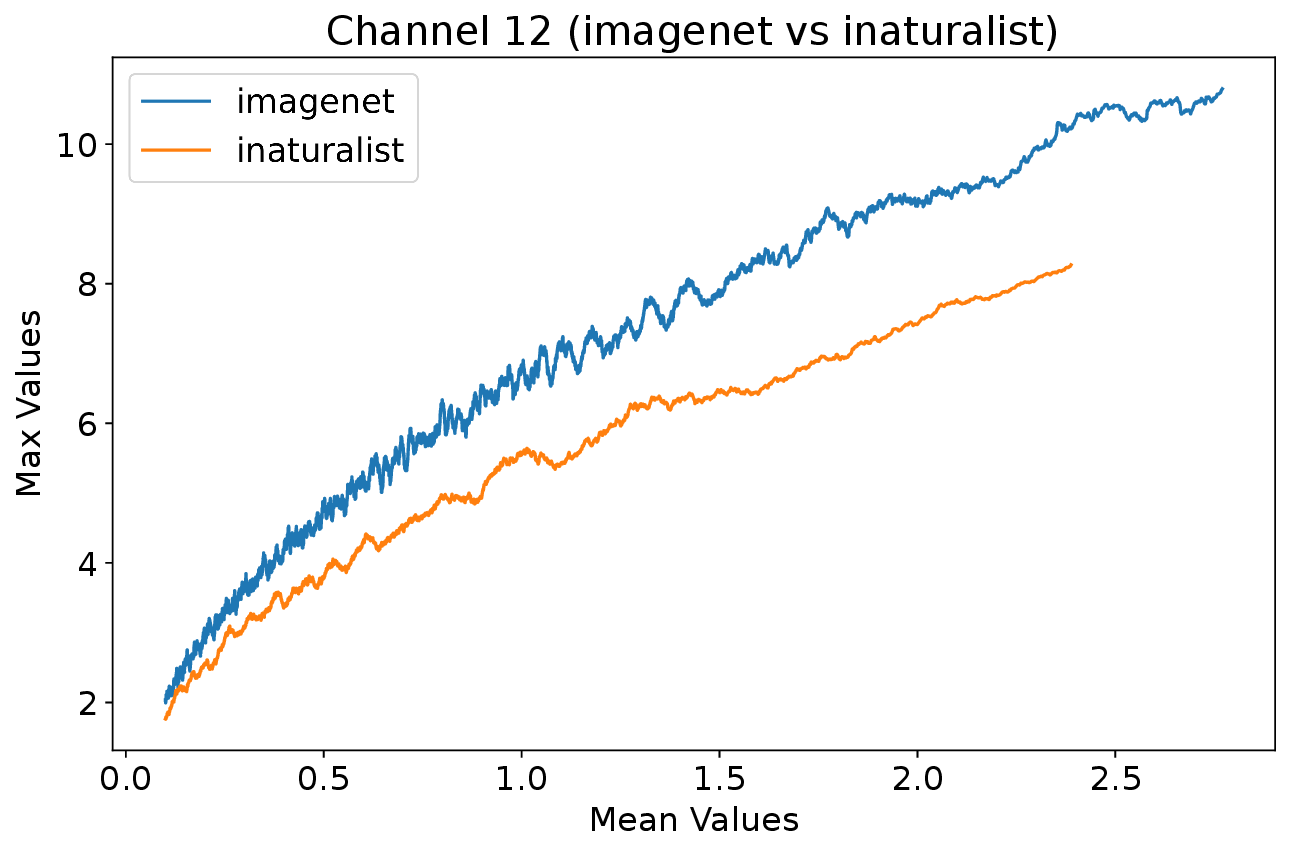}
    \includegraphics[width=0.24\linewidth]{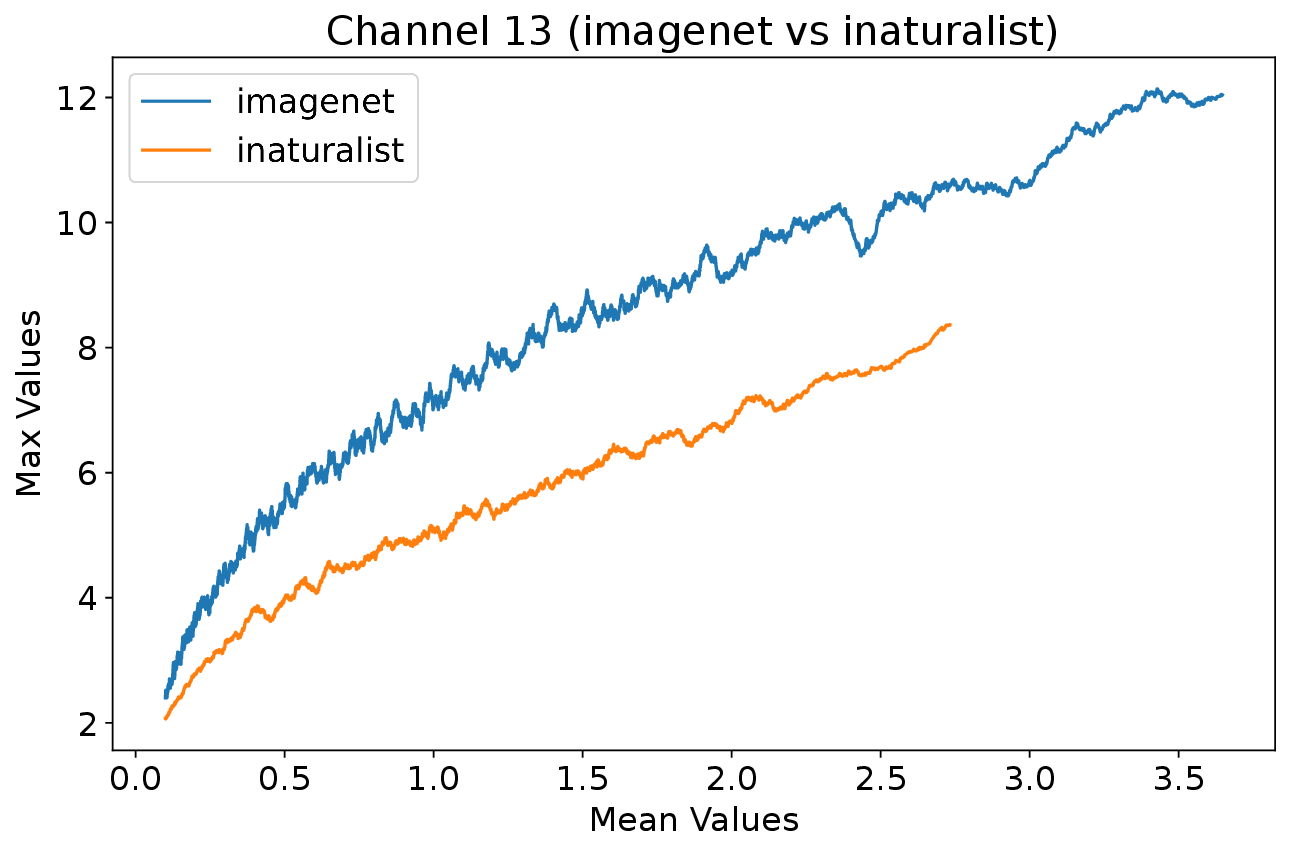}
    \includegraphics[width=0.24\linewidth]{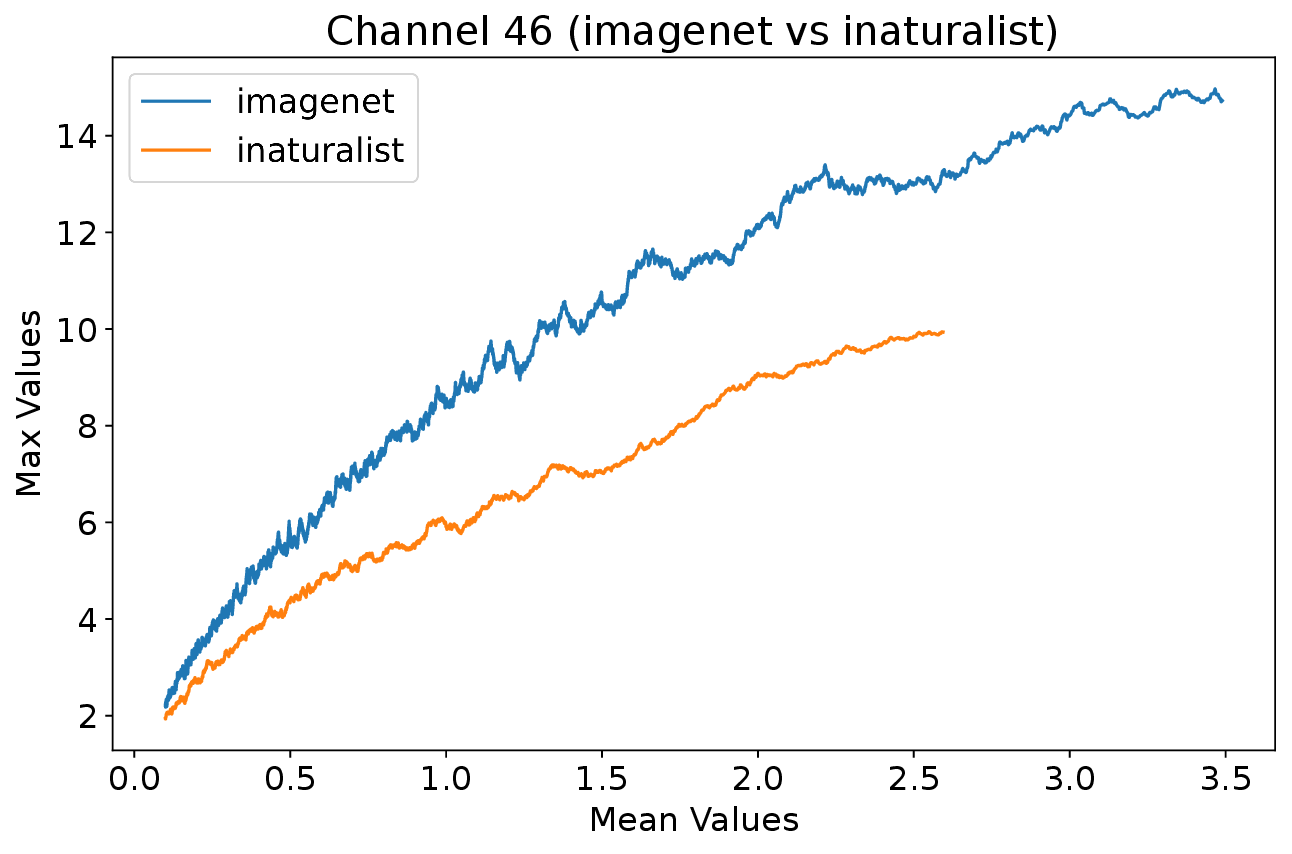}
    \includegraphics[width=0.24\linewidth]{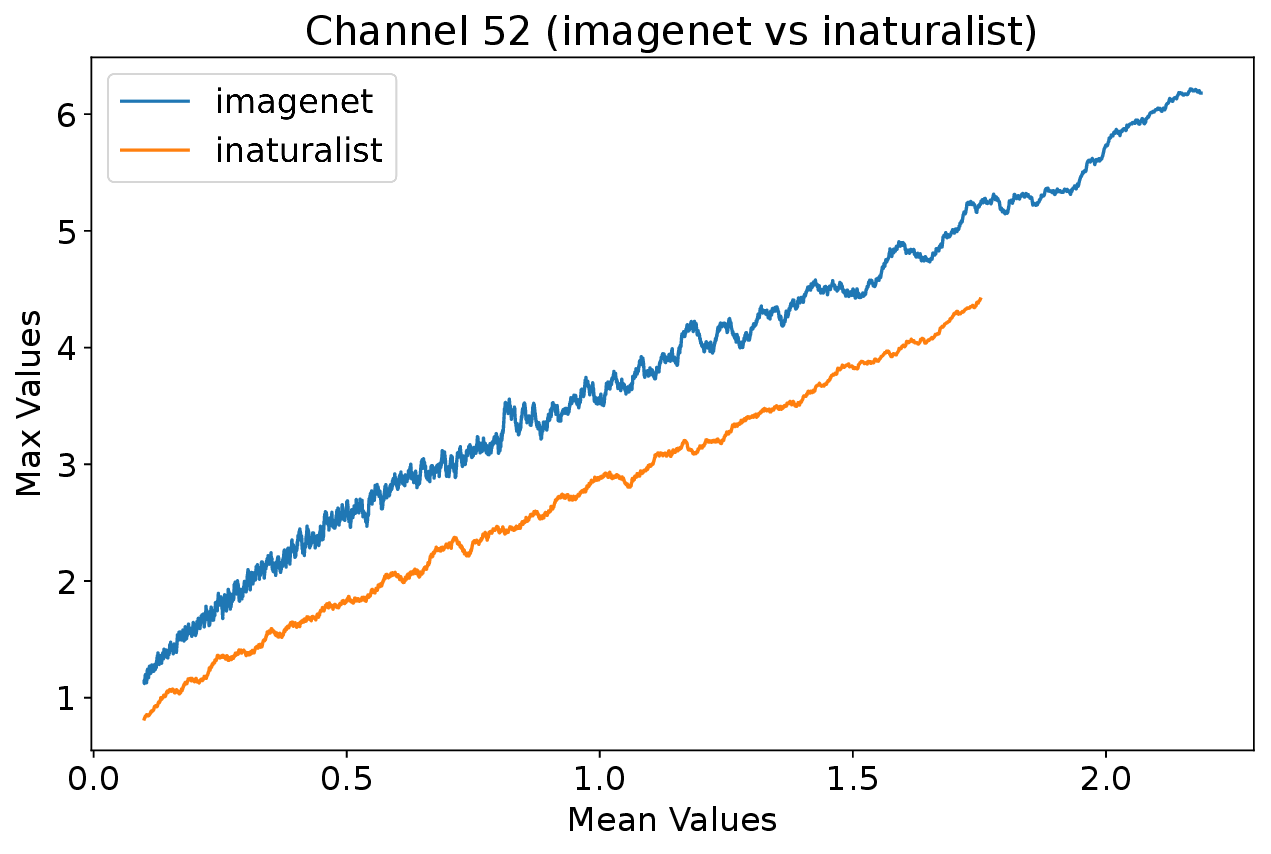}
    \label{fig:vgg-r1}
  \end{subfigure}

  \begin{subfigure}{\linewidth}
    \includegraphics[width=0.24\linewidth]{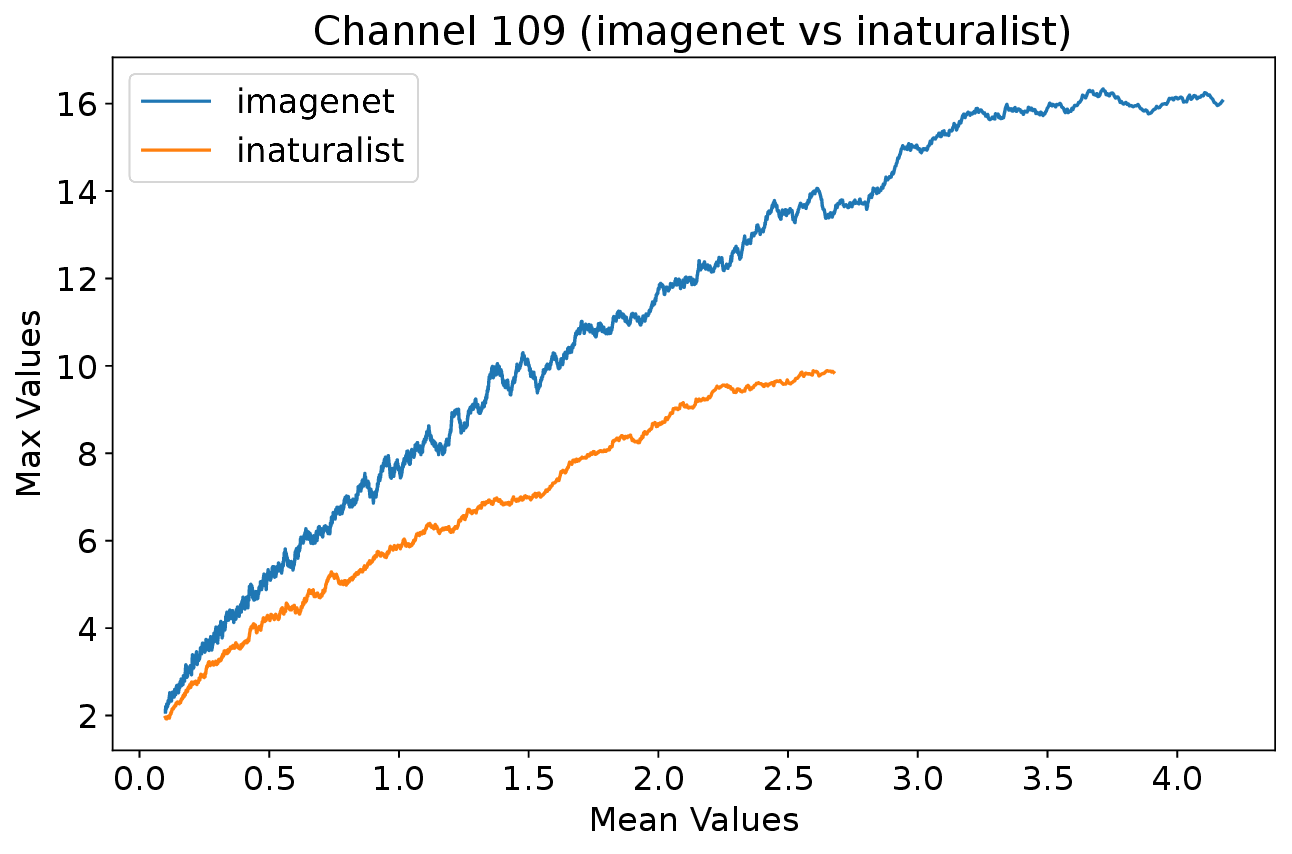}
    \includegraphics[width=0.24\linewidth]{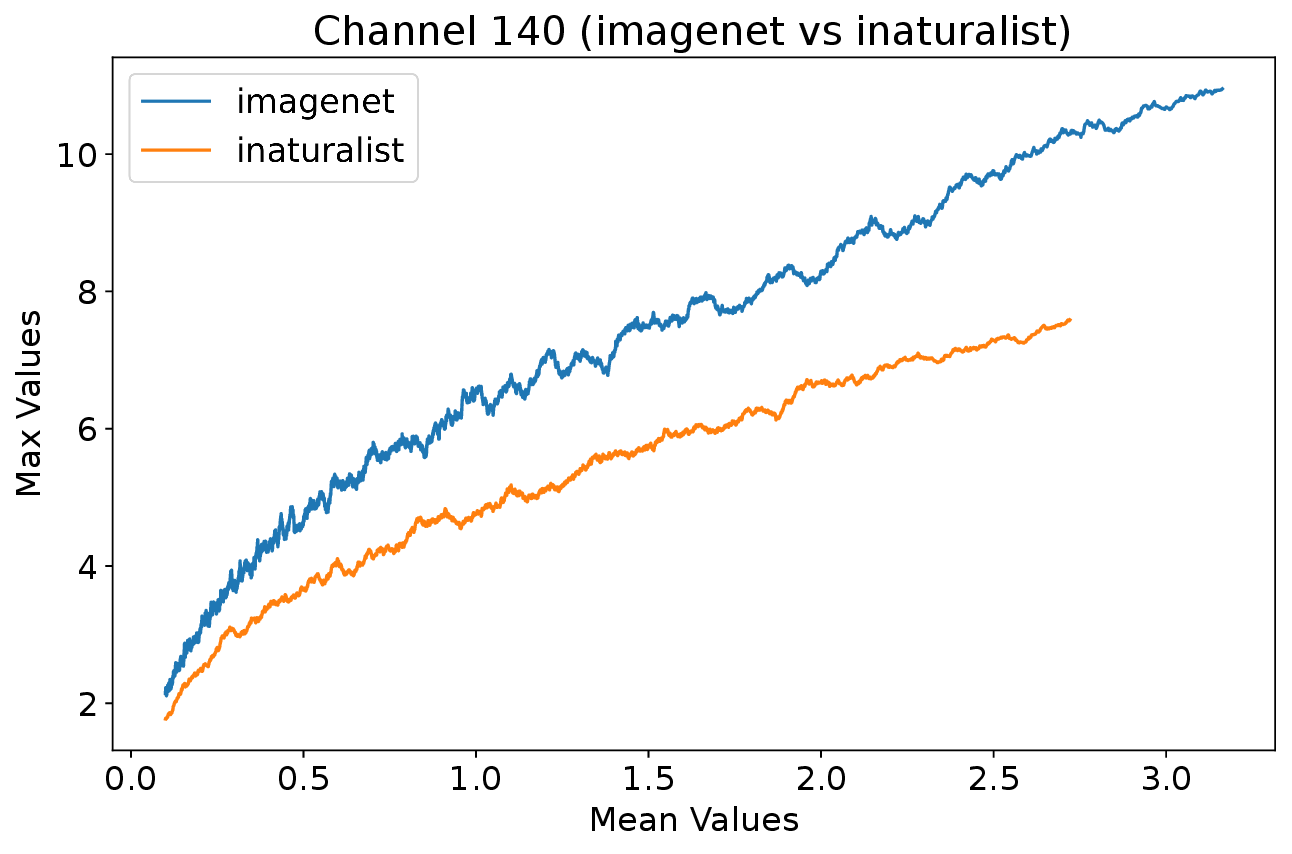}
    \includegraphics[width=0.24\linewidth]{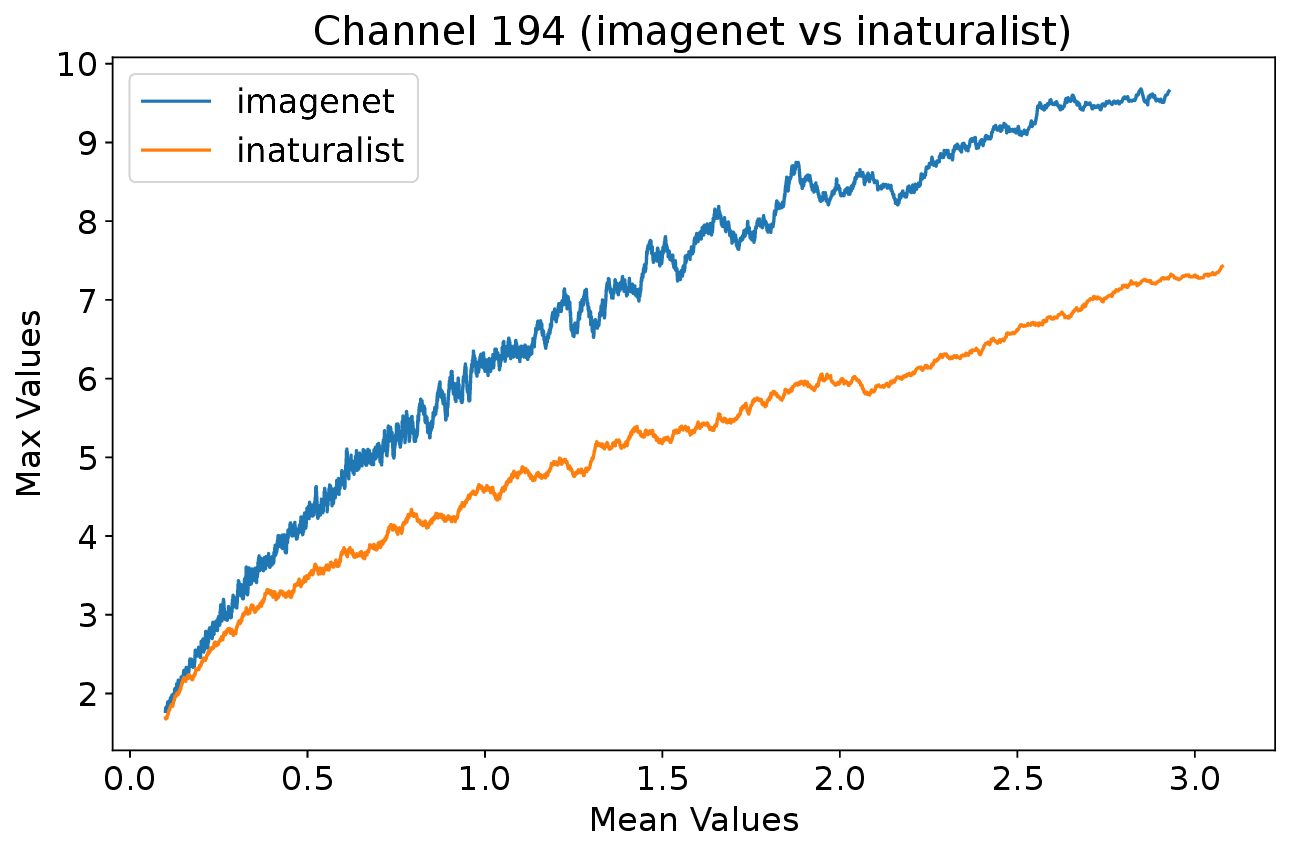}
    \includegraphics[width=0.24\linewidth]{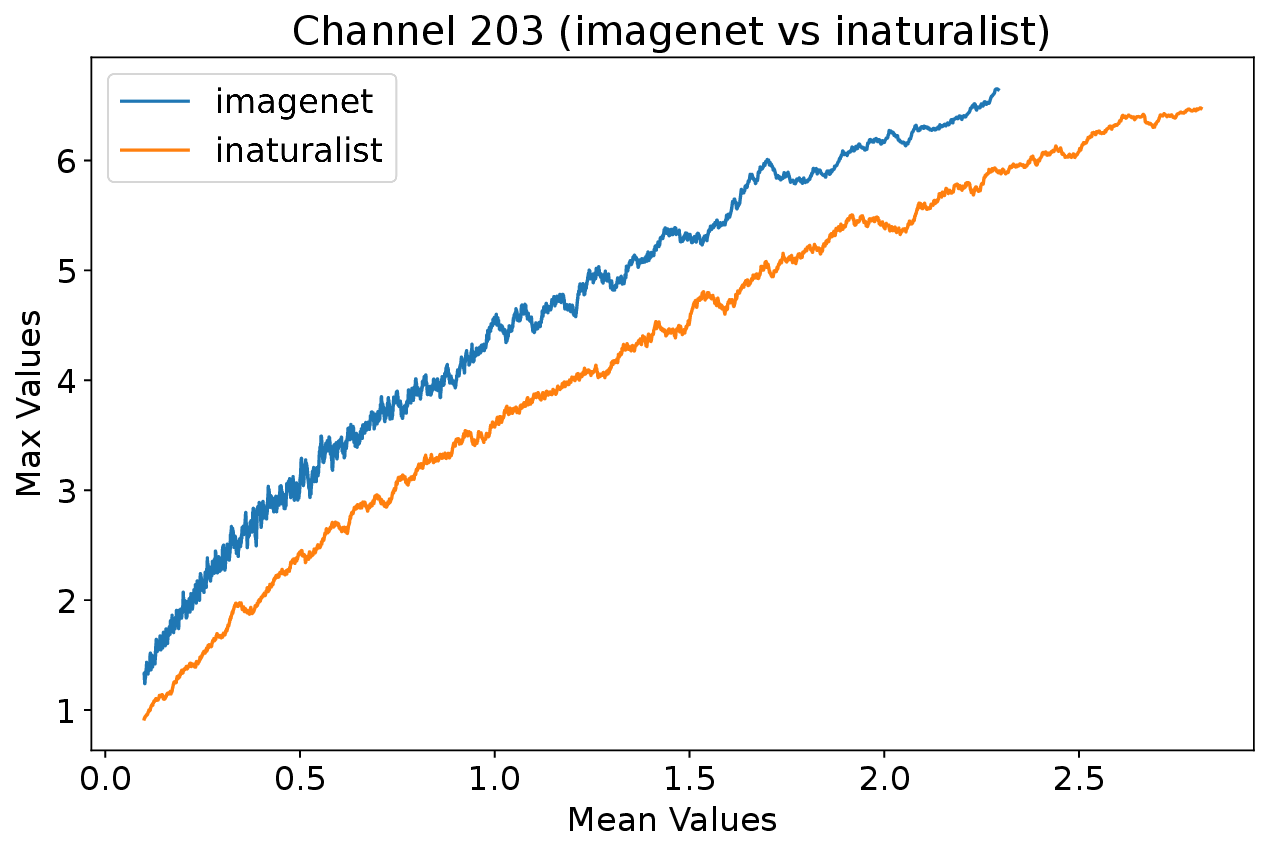}
    \label{fig:vgg-r2}
  \end{subfigure}

  \caption{\textbf{Activation distribution at penultimate layer before global pooling operation within the VGG16~\cite{simonyan2015very} architecture applied to ImageNet-1k and iNaturalist datasets~\cite{van2018inaturalist}.} We only include data points with an average activation over $0.1$. The figures show that our Neural Activation Prior (NAP) method is also effective in VGG16, proving that NAP can be applied to different architectures.}
  \label{fig:appendix-e-vgg}
\end{figure}

\section{Discussion}
\label{appendix:discuss}
Based on the ultra effectiveness of our prior, shown in the visualization results presented in Figure 1 in the main text and Figure~\ref{fig:appendix-e-mobilenet}, \ref{fig:appendix-e-resnet}, and \ref{fig:appendix-e-vgg} in this appendix, we are confident that more effective scoring functions exist. 
Due to space limitations, the primary focus and contribution of this paper is to introduce  this prior and and validate its effectiveness by proposing a new score function.
We leave the development of optimal scoring functions based on our prior for future work, aiming to further contribute to the community. We also hope this paper could encourage the researchers in  this community to build upon our work and advance the field.
\section{Licenses for existing assets}
\label{appendix:license}
We use several datasets and external code libraries in our research. To maintain anonymity, explicit references and detailed license information are not included in the submitted code. However, all creators and original owners of these assets are properly credited, and their licenses and terms of use are respected. The datasets used include CIFAR-10~\cite{krizhevsky2009cifar}, CIFAR-100~\cite{krizhevsky2009cifar}, ImageNet-1k~\cite{deng2009imagenet}, CIFAR-10-C~\cite{hendrycks2018corp}, CIFAR-100-C~\cite{hendrycks2018corp}, ImageNet-1k-C~\cite{hendrycks2018corp}, SVHN~\cite{netzer2011svhn}, iSUN~\cite{xu2015isun}, LSUN~\cite{yu2015lsun}, LSUN-crop~\cite{yu2015lsun}, iNaturalist~\cite{van2018inaturalist}, SUN~\cite{xiao2010sun}, Places~\cite{zhou2017places}, Textures~\cite{cimpoi2014texture}, and Places365~\cite{zhou2017places}. We also utilize DenseNet~\cite{huang2017densely}, ResNet~\cite{he2016deep}, VGG~\cite{simonyan2015very}, and ViT~\cite{dosovitskiy2020image} architectures in our experiments.
Our codebase for Neural Activation Prior (NAP) includes original code under the MIT License, with additional components from other sources. External code sources include ASH~\cite{djurisic2022extremely}, DICE~\cite{sun2022dice}, and~\cite{sun2022knn} under the MIT License.

\clearpage
\newpage
\section*{NeurIPS Paper Checklist}

\begin{enumerate}

\item {\bf Claims}
    \item[] Question: Do the main claims made in the abstract and introduction accurately reflect the paper's contributions and scope?
    \item[] Answer: \answerYes{} % Replace by \answerYes{}, \answerNo{}, or \answerNA{}.
    \item[] Justification: The abstract and introduction clearly state the main contributions and scope of the paper, which are supported by the theoretical and experimental results presented.

\item {\bf Limitations}
    \item[] Question: Does the paper discuss the limitations of the work performed by the authors?
    \item[] Answer: \answerYes{} % Replace by \answerYes{}, \answerNo{}, or \answerNA{}.
    \item[] Justification: The limitations of this work are discussed in detail in Appendix~\ref{appendix:limit}

\item {\bf Theory Assumptions and Proofs}
    \item[] Question: For each theoretical result, does the paper provide the full set of assumptions and a complete (and correct) proof?
    \item[] Answer: \answerNA{} % Replace by \answerYes{}, \answerNo{}, or \answerNA{}.
    \item[] Justification: Our paper does not include theoretical results. 

    \item {\bf Experimental Result Reproducibility}
    \item[] Question: Does the paper fully disclose all the information needed to reproduce the main experimental results of the paper to the extent that it affects the main claims and/or conclusions of the paper (regardless of whether the code and data are provided or not)?
    \item[] Answer: \answerYes{} % Replace by \answerYes{}, \answerNo{}, or \answerNA{}.
    \item[] Justification: The experimental setup, datasets used, and evaluation metrics are thoroughly described, ensuring that the main experimental results can be reproduced.

\item {\bf Open access to data and code}
    \item[] Question: Does the paper provide open access to the data and code, with sufficient instructions to faithfully reproduce the main experimental results, as described in supplemental material?
    \item[] Answer: \answerYes{} % Replace by \answerYes{}, \answerNo{}, or \answerNA{}.
    \item[] Justification: The paper provides detailed descriptions of the experimental setup in Section~\ref{sec:exp} and in Appendx~\ref{appendix:w}. Additionally, the code necessary to reproduce the experiments has been submitted in the supplemental material, along with comprehensive instructions.

\item {\bf Experimental Setting/Details}
    \item[] Question: Does the paper specify all the training and test details (e.g., data splits, hyperparameters, how they were chosen, type of optimizer, etc.) necessary to understand the results?
    \item[] Answer: \answerYes{} % Replace by \answerYes{}, \answerNo{}, or \answerNA{}.
    \item[] Justification: The paper provides detailed descriptions of the experimental setup in Section~\ref{sec:exp} and in Appendx~\ref{appendix:w}.

\item {\bf Experiment Statistical Significance}
    \item[] Question: Does the paper report error bars suitably and correctly defined or other appropriate information about the statistical significance of the experiments?
    \item[] Answer: \answerNo{} % Replace by \answerYes{}, \answerNo{}, or \answerNA{}.
    \item[] Justification: The proposed method is deterministic and does not involve any stochastic components or random initialization that would require error bars or statistical significance testing. Therefore, error bars are not applicable in this context.

\item {\bf Experiments Compute Resources}
    \item[] Question: For each experiment, does the paper provide sufficient information on the computer resources (type of compute workers, memory, time of execution) needed to reproduce the experiments?
    \item[] Answer: \answerYes{} % Replace by \answerYes{}, \answerNo{}, or \answerNA{}.
    \item[] Justification: Detailed information about the compute resources used for the experiments is provided in Section~\ref{sec:exp} of the paper.
    
\item {\bf Code Of Ethics}
    \item[] Question: Does the research conducted in the paper conform, in every respect, with the NeurIPS Code of Ethics \url{https://neurips.cc/public/EthicsGuidelines}?
    \item[] Answer: \answerYes{} % Replace by \answerYes{}, \answerNo{}, or \answerNA{}.
    \item[] Justification: The research presented in this paper adheres to the NeurIPS Code of Ethics. All experiments were conducted with integrity and transparency, ensuring fairness, accountability, and inclusivity.

\item {\bf Broader Impacts}
    \item[] Question: Does the paper discuss both potential positive societal impacts and negative societal impacts of the work performed?
    \item[] Answer: \answerYes{} % Replace by \answerYes{}, \answerNo{}, or \answerNA{}.
    \item[] Justification: The broader impacts of this work are discussed in Appendix~\ref{appendix:impacts}.
    
\item {\bf Safeguards}
    \item[] Question: Does the paper describe safeguards that have been put in place for responsible release of data or models that have a high risk for misuse (e.g., pretrained language models, image generators, or scraped datasets)?
    \item[] Answer: \answerNA{} % Replace by \answerYes{}, \answerNo{}, or \answerNA{}.
    \item[] Justification: This paper poses no such risks.

\item {\bf Licenses for existing assets}
    \item[] Question: Are the creators or original owners of assets (e.g., code, data, models), used in the paper, properly credited and are the license and terms of use explicitly mentioned and properly respected?
    \item[] Answer: \answerYes{} % Replace by \answerYes{}, \answerNo{}, or \answerNA{}.
    \item[] Justification: The paper properly credits all creators and original owners of assets used, including datasets and code in Appendix~\ref{appendix:license}. Additionally, the licenses and terms of use for these datasets are explicitly mentioned and respected. For code and models, appropriate references and acknowledgments are included, and the licenses under which they are released are adhered to, ensuring compliance with their terms of use.

\item {\bf New Assets}
    \item[] Question: Are new assets introduced in the paper well documented and is the documentation provided alongside the assets?
    \item[] Answer: \answerNA{} % Replace by \answerYes{}, \answerNo{}, or \answerNA{}.
    \item[] Justification: This paper does not release new assets.

\item {\bf Crowdsourcing and Research with Human Subjects}
    \item[] Question: For crowdsourcing experiments and research with human subjects, does the paper include the full text of instructions given to participants and screenshots, if applicable, as well as details about compensation (if any)? 
    \item[] Answer: \answerNA{} % Replace by \answerYes{}, \answerNo{}, or \answerNA{}.
    \item[] Justification: This paper does not involve crowdsourcing nor research with human subjects.

\item {\bf Institutional Review Board (IRB) Approvals or Equivalent for Research with Human Subjects}
    \item[] Question: Does the paper describe potential risks incurred by study participants, whether such risks were disclosed to the subjects, and whether Institutional Review Board (IRB) approvals (or an equivalent approval/review based on the requirements of your country or institution) were obtained?
    \item[] Answer: \answerNA{} % Replace by \answerYes{}, \answerNo{}, or \answerNA{}.
    \item[] Justification: This paper does not involve crowdsourcing nor research with human subjects.

\end{enumerate}

\end{document}